\documentclass{shthesis}

\usepackage[colorinlistoftodos]{todonotes}
\usepackage{subfig}
\usepackage{url}
\usepackage{longtable}
\usepackage{booktabs}
\usepackage{bibentry}
\nobibliography*
\usepackage[utf8]{inputenc}
\usepackage[framemethod=tikz]{mdframed}
\usepackage[nolist]{acronym}
\usepackage{CJKutf8}
\usepackage{natbib}
\usepackage{tabularx}
\usepackage{dcolumn}
\usepackage{longtable}
\usepackage{supertabular}
\usepackage{xspace}
\usepackage{listings}
\usepackage{fancyvrb}
\usepackage{comment}
\usepackage{wrapfig}  
\usepackage{caption}
\usepackage{longtable}
\usepackage{multirow}
\usepackage{colortbl}
\usepackage[T1]{fontenc}
\usepackage{tikz}
\usepackage{footnote}
\usepackage[ruled,vlined,linesnumbered]{algorithm2e}
\usepackage{url}
\usepackage{mathtools}
\usepackage{verbatim}
\usepackage{lipsum}
\usepackage{listingsutf8}
\usepackage{lscape}
\usepackage{pifont}
\usepackage{numprint}
\usepackage{adjustbox}
\usepackage{multirow}
\usepackage{multicol}
\usepackage{float}

\usepackage{hyperref}
\usepackage{graphicx}
\usepackage{numprint}
\usepackage{xcolor}
\usepackage{booktabs} 
\usepackage{microtype}

\usepackage{todonotes}
\usepackage{latexsym}
\usepackage{dcolumn}
\usepackage{graphics}
\usepackage{url}
\usepackage{longtable}
\usepackage{supertabular}
\usepackage{pdflscape}
\usepackage{pifont}

\usepackage{rotating}
\usepackage{tabularx}
\usepackage{lscape}
\usepackage{multirow}
\usepackage{multicol}
\usepackage{float}
\usepackage{footnote}
\usepackage{verbatim}
\usepackage{url}
\usepackage{mathtools}
\usepackage{xspace}
\usepackage{listingsutf8}
\usepackage{color}
\usepackage{tikz}
\usepackage{lipsum}
\usepackage[ruled,vlined]{algorithm2e}
\usepackage{mdframed}
\usepackage{wrapfig}
\usepackage[nolist]{acronym}
\usepackage{adjustbox}
\usepackage{array}
\usepackage{subfiles}
\newcommand{\furl}[1]{\footnote{\scriptsize \url{#1}}}

\newcolumntype{R}[2]{%
    >{\adjustbox{angle=#1,lap=\width-(#2)}\bgroup}%
    l%
    <{\egroup}%
}

\newcommand{\tablefont}[1]{\fontsize{3mm}{3.2mm}\selectfont}  

\usepackage{lineno,hyperref}
\modulolinenumbers[5]

\usepackage{times}%
\usepackage{dcolumn}
\usepackage{graphics}
\usepackage{graphicx}
\usepackage{url}
\usepackage{longtable}
\usepackage{supertabular}
\usepackage{pdflscape}
\usepackage{pifont}
\usepackage{rotating}
\usepackage{array}
\usepackage{tabularx}
\usepackage{lscape}
\usepackage{multirow}
\usepackage{multicol}
\usepackage{pdflscape}
\usepackage{float}
\usepackage{mathtools}
\usepackage{footnote}
\usepackage{verbatim}
\usepackage{supertabular}
\usepackage{hyperref}
\usepackage{url}
\usepackage{mathtools}
\usepackage{xspace}
\usepackage{listingsutf8}
\usepackage{lipsum}
\usepackage{tabulary}
\usepackage{svg}
\usepackage{threeparttable, tablefootnote}

\newtheorem{definition}{Definition}[chapter]

\definecolor{gray}{rgb}{0.4,0.4,0.4}
\definecolor{darkblue}{rgb}{0.0,0.0,0.6}
\definecolor{cyan}{rgb}{0.0,0.6,0.6}
\definecolor{cffffff}{RGB}{255,255,255}

\lstdefinelanguage{XML}
{
  morestring=[b]",
  morestring=[s]{>}{<},
  morecomment=[s]{<?}{?>},
  stringstyle=\color{black},
  identifierstyle=\color{darkblue},
  keywordstyle=\color{cyan},
  morekeywords={xmlns,version,type}
}

\definecolor{codegreen}{rgb}{0,0.6,0}
\definecolor{codegray}{rgb}{0.5,0.5,0.5}
\definecolor{codeblue}{rgb}{0,0,255}
\definecolor{backcolour}{rgb}{0.95,0.95,0.92}

\lstdefinestyle{rdf}{numberblanklines=true, morekeywords={},
backgroundcolor=\color{backcolour},   
    commentstyle=\color{codegreen},
    keywordstyle=\color{magenta},
    numberstyle=\tiny\color{codegray},
    stringstyle=\color{codeblue},
    basicstyle=\footnotesize,
    breakatwhitespace=false,         
    breaklines=true,                 
    captionpos=b,                    
    keepspaces=true,                 
    numbers=left,                    
    numbersep=5pt,                  
    showspaces=false,                
    showstringspaces=false,
    showtabs=false,                  
    tabsize=2
}

\lstdefinestyle{sparql}{numberblanklines=true, morekeywords={SERVICE,SELECT,DISTINCT,SAMPLE,FROM,WHERE,FILTER,ORDER,GROUP,BY,IN,AS,LIMIT}}
 
\newcommand{\rdfLangString}{\texttt{rdf:langString}}

\definecolor{dkgreen}{rgb}{0,0.6,0}
\definecolor{gray}{rgb}{0.5,0.5,0.5}
\definecolor{light-gray}{rgb}{0.97,0.97,0.97}

\definecolor{olivegreen}{rgb}{0.2,0.8,0.5}
\definecolor{grey}{rgb}{0.5,0.5,0.5}
\lstdefinelanguage{ttl}{
  basicstyle=\small\ttfamily,
  sensitive=true,
  morecomment=[l][\color{grey}]{@},
  morecomment=[l][\color{olivegreen}]{\#},
  morestring=[b][\color{blue}]\",
}
\lstdefinelanguage{manchester}
{morekeywords={owl,xml,dc,rdf,skos,description,PlainLiteral,int,float,
        some,only,value,min,exactly,max,and,or,not,SOME,ONLY,VALUE,MIN,EXACTLY,MAX,AND,OR,NOT,
        Prefix,Ontology,Import,Individual,Facts,Types,Class,
        DataProperty,ObjectProperty,AnnotationProperty,Annotations,
DifferentIndividuals,SubClassOf,EquivalentTo,DisjointWith,DisjointUnionOf,SubPropertyOf,DisjointClasses,DisjointProperties,
Symmetric,Asymmetric,Reflexive,Irreflexive,Transitive,Functional,InverseFunctional,
        Characteristics,Range,Domain,Datatype},
     basicstyle=\small\ttfamily,
     keywordstyle=\bfseries,
     commentstyle=\color{gray},
     stringstyle=\color{blue},
     numbers=left,
     numberstyle=\tiny\color{gray},
     stepnumber=1,
     numbersep=10pt,
     tabsize=2,
     showspaces=false,
     showstringspaces=false,
     breaklines=true,                           
     sensitive=true,                            
     morecomment=[l][commentstyle]{\#},         
     morestring=[b]",                           
     numbers=none
}
\lstdefinelanguage{sparql}{%
   morekeywords=[1]{CONSTRUCT,WHERE,SELECT},
   morekeywords=[2]{AND,FILTER,UNION,OPT,OPTIONAL,MINUS,ORDER,GROUP,BY,DESC,OFFSET,LIMIT},%
   morekeywords=[3]{sameTerm,isBLANK,isLITERAL,isIRI,BOUND,DISTINCT},
   morekeywords=[4]{rdf,rdfs,owl,dbo,res,xsd},
   morekeywords=[5]{>},
   morestring=[b]",%
   alsodigit={-},%
}[keywords,strings]

\usepackage{colortbl}
\usepackage{amsmath}
\usepackage{xcolor}
\usepackage{graphicx}
\usepackage{array}
\usepackage{makecell}

\colorlet{tableheadcolor}{gray!25} 
\colorlet{tablerowcolor}{gray!10} 
\lstdefinestyle{rdf}{numberblanklines=false, morekeywords={},
backgroundcolor=\color{backcolour},   
    commentstyle=\color{codegreen},
    keywordstyle=\color{magenta},
    numberstyle=\tiny\color{codegray},
    stringstyle=\color{codeblue},
    basicstyle=\footnotesize,
    breakatwhitespace=false,         
    breaklines=true,                 
    captionpos=b,                    
    keepspaces=true,                 
    numbers=left,                    
    numbersep=5pt,                  
    showspaces=false,                
    showstringspaces=false,
    showtabs=false,                  
    tabsize=2
}

\lstdefinestyle{sparql}{numberblanklines=false, morekeywords={SERVICE,SELECT,DISTINCT,SAMPLE,FROM,WHERE,FILTER,ORDER,GROUP,BY,IN,AS,LIMIT}}

\definecolor{mycolor}{rgb}{0.122, 0.435, 0.698}
\definecolor{mycolorch}{rgb}{0.8, 0.0, 0.0}
\definecolor{mycolorrq}{rgb}{1.0, 0.75, 0.0}

\newmdenv[innerlinewidth=0.5pt, roundcorner=4pt,linecolor=mycolor,innerleftmargin=6pt,
innerrightmargin=6pt,innertopmargin=6pt,innerbottommargin=6pt]{mybox}

\newmdenv[innerlinewidth=0.5pt, roundcorner=4pt,linecolor=mycolorch,innerleftmargin=6pt,
innerrightmargin=6pt,innertopmargin=6pt,innerbottommargin=6pt]{myboxch}

\newmdenv[innerlinewidth=0.5pt, roundcorner=4pt,linecolor=mycolorrq,innerleftmargin=6pt,
innerrightmargin=6pt,innertopmargin=6pt,innerbottommargin=6pt]{mycolorrq}

\newcommand*\rot{\multicolumn{1}{R{80}{1em}}}

\ifLuaTeX
  \RequirePackage{lua-visual-debug}
\fi

\usepackage{xspace}
\usepackage{pifont} 

\usepackage{pdfpages}

\usepackage{pgfplots}
\graphicspath{{./figures/}}

\usepackage[ngerman,american]{babel}  

\title{Knowledge Graphs for Multilingual Language
Translation and Generation}

\author{Diego Campos Moussallem}

\degree{Dr.\ rer.\ nat.}
\degreeyear{2020}
\degreemonth{March}

\department{Faculty for Computer Science,\\ Electrical Engineering and Mathematics}
\university{Paderborn University}
\universitycity{Paderborn}
\universitycountry{Germany}

\usepackage{snapshot}

\begin{document}
\startcontents
  
\frontmatter

\begin{titlepage}%
  \begin{center}%
    \hspace{-1.25cm}\includegraphics[width=0.5\textwidth]{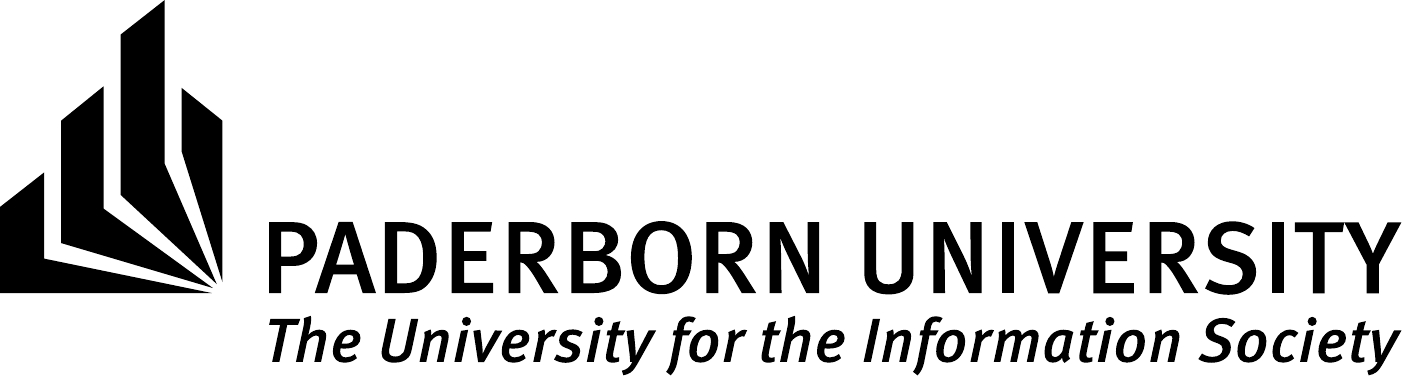}\par%
    \vspace{1cm}%
    {\Large \textsc{Doctoral Dissertation}\par}%
    {%
      \Huge%
      \rule{.9\linewidth}{.6pt}\par%
      \textcolor{shtitlecolor}{\shtitlebreak}\par%
      \rule[1.3ex]{.9\linewidth}{.6pt}\par
    }%
    \vspace{3em}%
    \normalsize%
    \textrm{%
      A dissertation presented\\%
      by\\%
      \shauthor\\%
      to the\\%
      \shdepartment\\%
      of\\%
      Paderborn University\\%
      \vspace*{24pt}%
      in partial fulfillment of the requirements\\%
      for the degree of\\%
      \shdegree\\%
      \vspace{\fill}%
      Paderborn, Germany\\
      \shdegreemonth\ \shdegreeyear%
    }\par%
  \end{center}\par%
\end{titlepage}%


\thispagestyle{empty}
\vspace*{\fill}
\subsubsection{Dissertation}

Knowledge Graphs for Multilingual Language Translation and Generation

\noindent
Diego Campos Moussallem, Paderborn University

\noindent
Paderborn, Germany, 2020

\subsubsection{Reviewers}

\noindent
Prof.\ Dr.\ Axel-Cyrille Ngonga Ngomo, Paderborn University

\noindent
Prof.\ Dr.\ S{ö}ren Auer, Leibniz Universit{ät} Hannover

\noindent
Prof.\ Dr.\ Jens Lehmann , Universität Bonn

\subsubsection{Doctoral Committee}

\noindent
Prof.\ Dr.\ Axel-Cyrille Ngonga Ngomo, Paderborn University

\noindent
Prof.\ Dr.\ S{ö}ren Auer, Leibniz Universit{ä}t Hannover

\noindent
Prof.\ Dr.\ Jens Lehmann, Universität Bonn

\noindent
Prof.\ Dr.\ Heike Wehrheim, Paderborn University

\noindent
Prof.\ Dr.\ Gregor Engels, Paderborn University


\chapter*{\centering Abstract}
\section*{\centering \shtitle}
The \ac{NLP} community has recently seen outstanding progress, catalysed by the release of different \ac{NN} architectures. Neural-based approaches have proven effective by significantly increasing the output quality of a large number of automated solutions for \ac{NLP} tasks~\citep{belinkov-glass-2019-analysis}. Despite these notable advancements, dealing with entities still poses a difficult challenge as they are rarely seen in training data. Entities can be classified into two groups, i.e., proper nouns and common nouns. Proper nouns are also known as Named Entities (NE) and correspond to the name of people, organizations or locations, e.g., \textit{John}, \textit{WHO} or \textit{Canada}.  Common nouns describe classes of objects, e.g., \textit{spoon} or \textit{cancer}.  Both types of entities can be found in a \ac{KG}. Recent work has successfully exploited the contribution of \ac{KG}s in \ac{NLP} tasks, such as \ac{NLI}~\citep{annervaz2018} and \ac{QA}~\citep{Sorokincoling2018}.
Only a few works had exploited the benefits of \ac{KG}s in \ac{NMT} when the work presented herein began.
Additionally, few works had 
studied the contribution of \ac{KG}s to \ac{NLG} tasks. Moreover, the multilinguality also remained 
an open research area in these respective tasks~\citep{young2018recent}.


In this thesis, we focus on the use of \ac{KG}s for machine translation and the generation of texts to deal with the problems caused by entities 
and consequently enhance the quality of automatically generated texts. 
Before handling entities in translation or generation, the first research challenge of this thesis lies in the disambiguation of entities. Some entities are highly ambiguous, e.g., \textit{Kiwi} can be a fruit or bird. However, once they are disambiguated, their translations are found in multilingual \ac{KG}s. We addressed this challenge by devising MAG, a multilingual knowledge graph-based entity linking approach for 40 languages. MAG achieves an average of 0.63 F-measure across all languages and places first out of 13 annotation systems. 

Our second research challenge is how to cope with entities while generating natural language sentences in different languages from \ac{RDF} \acp{KG}. The underlying rationale is that generating entities from \ac{KG}s shares similar \ac{NMT} problems as 
translating them between languages in texts. We noticed that previous work has predominantly focused on English, and only a few works provided 
solutions for other languages. 
We dealt with this challenge by creating a Portuguese \ac{RDF} verbalizer, named RDF2PT, which was further extended to Spanish and English. 
RDF2PT generates sentences and small summaries in Portuguese, which show fluency almost equivalent to humans, scoring 4 (exact mean) 
on a 5-Likert scale. 
Further, we examined the \ac{REG} task that aims to 
choose the referential form of entities while generating texts. We then created the first neural-based \ac{REG} model, named NeuralREG, which clearly outperforms the state of the art
, scoring 5.26 (exact mean) on a 7-Likert scale.   

Our third research challenge involves the translation of entities in text. With this aim, we applied \ac{KG}s into \ac{NMT} models. We thus created the first \ac{KG}-augmented \ac{NMT} model, named KG-NMT, by combining \ac{EL} and \ac{KGE}. KG-NMT achieves consistent translation improvements up to +3 BLEU, METEOR, and chrF3 on open domain datasets, and on domain-specific data and ontologies. Later, we discerned that applying \ac{KG}s into \ac{NLP} tasks requires rich language-based \ac{KG}s. We therefore devised our fourth research challenge which pertains to the low resource language problem in \ac{KG}s.
To that end, we developed the first neural-based approach, named THOTH, for translating and enriching \ac{KG}s 
across languages. THOTH achieves a translation accuracy of 86\%, and its artificially enriched \ac{KG}s improve the \ac{EL} task by +19\% F-measure. Overall, our findings show that the application of \ac{KG}s is an effective way of handling entities and addressing its related data sparsity issues in multilingual text translation and generation.


\chapter*{\centering Zusammenfassung}
\section*{\centering \shtitle}
\begin{otherlanguage}{ngerman}
Die Natural Language Processing (NLP)-Gemeinschaft hat in letzter Zeit herausragende Fortschritte erzielt, die durch die Veröffentlichung verschiedener Architekturen künstlicher neuronaler Netze (NN) katalysiert wurden. NN-basierte Ansätze haben sich als effektiv erwiesen, da sie die Qualität der  automatisiert erstellten Lösungen für eine große Zahl von NLP-Aufgaben~\citep{belinkov-glass-2019-analysis} deutlich erhöht haben. Trotz dieser bemerkenswerten Fortschritte stellt der Umgang mit Entitäten immer noch eine schwierige Herausforderung dar, da sie in den Trainingsdaten nur selten zu vorkommen. Entitäten lassen sich in zwei Gruppen einteilen: Eigennamen und Gattungsnamen. Eigennamen werden auch als Named Entities (NE) bezeichnet und entsprechen den Namen von Personen, Organisationen oder Orten, z. B. \textit{John}, \textit{WHO} oder \textit{Kanada}. Gattungsnamen beschreiben Klassen von Objekten, z. B. \textit{Löffel} oder \textit{Krebs}. Beide Typen von Entitäten können in einem Wissensgraphen (KG) gefunden werden. In jüngster Zeit wurden KGs erfolgreich bei der Lösung von NLP-Aufgaben genutzt, wie z. B. Natural Language Inference~\citep{annervaz2018} und Question Answering~\citep{Sorokincoling2018}. Dagegen haben sich nur wenige Arbeiten mit der Anwendung von KGs für die maschinelle, neuronale Übersetzung (NMT) oder der Generierung von natürlicher Sprache (NLG) beschäftigt, als mit dieser Arbeit begonnen wurde. Darüber hinaus ist die Mehrsprachigkeit bei diesen beiden Problemen weiterhin ein offenes Forschungsgebiet~\citep{young2018recent}.

In dieser Arbeit konzentrieren wir uns auf die Verwendung von KGs für die maschinelle Übersetzung und die Generierung von Texten, um die durch Entitäten verursachten Probleme zu behandeln und folglich die Qualität automatisch generierter Texte zu verbessern. Zuvor wird in dieser Arbeit die Disambiguierung von Entitäten behandelt. Einige Entitäten sind hochgradig mehrdeutig, z. B. kann es sich bei dem Begriff  \textit{Kiwi} um eine Frucht oder einen Vogel handeln. Die Disambiguierung ermöglicht letztendlich das Auffinden von Übersetzungen in mehrsprachigen KGs. Zur Auflösung von Ambiguitäten wurde das Framework MAG entwickelt, welches auf mehrsprachigen Wissensgraphen basiert und die Verknüpfung von Entitäten in über 40 Sprachen ermöglicht. MAG erreicht ein durchschnittliches F-Measure von 0,63 über alle 40 Sprachen und steht damit an erster Stelle von 13 Annotationssystemen.

Im zweiten Teil wird die Frage behandelt, wie mit Entitäten bei der Generierung von Sätzen basierend auf dem Resource Description Framework (RDF) in verschiedenen natürlichen Sprachen umzugehen ist. Die zugrundeliegende Überlegung ist, dass die Erzeugung von Entitäten aus KGs ähnliche NMT-Probleme aufweist wie die Übersetzung zwischen Sprachen in Texten. Da sich vorherige Ansätze hauptsächlich auf die englische Sprache fokussieren und es nur wenige Ansätze für weitere Sprachen gibt, wurde ein RDF-Verbalizer entwickelt, welcher sowohl portugiesische, spanische als auch englische Texte generieren kann. RDF2PT erzeugt Sätze und kleine Zusammenfassungen auf Portugiesisch, die eine fast menschenähnliche Sprachkompetenz zeigen und auf einer 5-Likert-Skala im Mittel mit 4 bewertet wird. Ferner wurde das Referring Expression Generation (REG) Problem behandelt, welches sich mit der Auswahl der referentielle Form von Entitäten beschäftigt. Anschließend entwickelten wir das erste NN-basierte REG-Modell, genannt NeuralREG, das den Stand der Technik deutlich übertrifft und auf einer 7-Likert-Skala mit 5,26 (exakter Mittelwert) bewertet wurde.

Unsere dritte Forschungsaufgabe betrifft die Übersetzung von Entitäten in Textex. Ziel ist es KGs in NMT-Modelle zu integrieren, indem Entity Linking (EL) und Knowledge Graph Embeddings (KGE) zu einem KG-NMT Modell kombiniert werden. KG-NMT erzielt konsistente Übersetzungsverbesserungen von bis zu +3 BLEU, METEOR und chrF3 bei offenen Domänendatensätzen und bei domänenspezifischen Daten und Ontologien. Zudem wurde festgestellt, dass die Anwendung von KGs in NLP-Aufgaben umfangreiches sprach basiertes KGs erfordert. Folglich beschäftigt sich der vierte Teil dieser Arbeit mit der Behandlung von Sprachen, für die nur wenig sprach basiertes Wissen verfügbar ist. Zu diesem Zweck entwickelten wir den ersten neuronal-basierten Ansatz namens THOTH, um KGs sprachübergreifend zu übersetzen und anzureichern. THOTH erreicht eine Übersetzungsgenauigkeit von 86\%, und seine künstlich angereicherten KGs verbessern die EL-Aufgabe um +19\% F-Measure. Insgesamt zeigen unsere Ergebnisse, dass die Anwendung von KGs eine effektive Methode ist, um mit Entitäten umzugehen und die damit verbundenen Probleme der Datensparsamkeit bei der Übersetzung und Erstellung mehrsprachiger Texte zu lösen.
\end{otherlanguage}

\chapter*{\centering Acknowledgments}
First of all, I would like to thank my wife, Carol. This thesis has only been possible due to her unconditional support. It is uncountable how often she supported me during all the challenging and stressful moments for completing this thesis. The thesis was written within the Data Science research group (DICE), led by Prof. Dr. Axel-Cyrille Ngonga Ngomo, who I wholeheartedly thank for being my advisor and granting me the freedom to develop and pursue my research ideas that have lead to this thesis work. 
I want to thank Dr. Sebastian Hellmann for inviting me to become a member of the Agile Knowledge Engineering and Semantic Web (AKSW) group at the University of Leipzig, where DICE began to take shape. I am delighted to be part of AKSW and DICE and have had the opportunity to work with so many talented people. Additionally, I want to thank Dr. Ricardo Usbeck for always supporting me within and out of the Ph.D. environment. I extend my gratitude to Dr. Paul Buitelaar and Dr. Mihael Arcan for the invitation to collaborate and do an internship in their research group, INSIGHT, at the National University of Ireland Galway. I am also grateful to have had the opportunity to meet Dr. Thiago Castro Ferreira, with whom I worked closely in various research projects within the framework of natural language generation. 
Special thanks go to Dr. Diego Esteves, who introduced me to the ASKW group and encouraged me to pursue a Ph.D. abroad. He became more than a true friend, a brother, and I will always be grateful to him. Moreover, I would like to thank my friends and colleagues from Leipzig and Paderborn, especially Tommaso Soru, André Valdestilhas, Edgard Marx, Kleanthi Georgala, Kunal Jha, and Amrapali Zaveri (in memoriam). It was an honor to meet all of you. Furthermore, I would like to extend my deepest gratitude to my family and to my parents, Vilma and Makhoul, especially my mother, who sacrificed her life to raise me to who I am now. I would also like to thank my mother-in-law, Isabel, for her great support. Next, I would like to thank my English teacher, Marcos Moret, for accompanying me all this time and for becoming a true friend. Finally, I want to thank the National Council for Scientific and Technological Development (CNPq) for the scholarship, which supported a significant part of this research. Likewise, I would like to thank DAAD for funding my German course and travel expenses.

\cleardoublepage 
\pdfbookmark{\contentsname}{Contents}
\begin{spacing}{\shnormalspacing}
  \hypersetup{linkcolor=}
  \printcontents{}{1}{\chapter*{\contentsname}}
\end{spacing}

\begin{acronym}[UML]
	\acro{AOS}{Agricultural Ontology Services}
	\acro{AGRIS}{Agricultural Science and Technology}
	\acro{API}{Application Programming Interface}
	\acro{AOS}{Agricultural Ontology Services}
	\acro{AGRIS}{Agricultural Science and Technology}
	\acro{API}{Application Programming Interface}
	\acro{A2KB}{Annotation to Knowledge Base}
	\acro{AI}{Artificial Intelligence}
	\acro{BPSO}{Binary Particle-Swarm Optimization}
	\acro{BPMLOD}{Best Practices for Multilingual Linked Open Data}
	\acro{BPSO}{Binary Particle-Swarm Optimization}
	\acro{BPMLOD}{Best Practices for Multilingual Linked Open Data}
	\acro{BFS}{Breadth-First-Search}
	\acro{BPE}{Byte Pair Encoding}
	\acro{BoW}{Bag-of-Words}
	\acro{CBD}{Concise Bounded Description}
	\acro{COG}{Content Oriented Guidelines}
	\acro{CSV}{Comma-Separated Values}
	\acro{CBMT}{Corpus-Based Machine Translation}
	\acro{CLIR}{Cross-Language Information Retrieval}
	\acro{DPSO}{Deterministic Particle-Swarm Optimization}
	\acro{DALY}{Disability Adjusted Life Year}
	\acro{DBMS}{Relational Database Management System}

	\acro{ER}{Entity Resolution}
	\acro{EM}{Expectation Maximization}
	\acro{EBMT}{Example-Based Machine Translation}
	\acro{EBNF}{Extended Backus--Naur Form}
	\acro{EL}{Entity Linking}
	\acro{FAO}{Food and Agriculture Organization of the United Nations}
	\acro{GIS}{Geographic Information Systems}
	\acro{GHO}{Global Health Observatory}
	\acro{GRU}{Gated recurrent unit}
	\acro{HDI}{Human Development Index}
	\acro{ICT}{Information and communication technologies}
	\acro{IFRS}{International Financial Reporting Standards}
	\acro{ICD}{International Classification of Diseases}
	\acro{IT}{Information Technology}
    \acro{KB}{Knowledge Base}
    \acro{KG}{Knowledge Graph}
    \acrodefplural{KG}{Knowledge Graphs}
    \acro{KGE}{Knowledge Graph Embeddings}
    \acro{KBSE}{Knowledge Base Semantic Embedding}
	\acro{LR}  {Language Resource}
	\acro{LD}  {Linked Data}
	\acro{LLOD}  {Linguistic Linked Open Data}
	\acro{LIMES}{LInk discovery framework for MEtric Spaces}
	\acro{LS}  {Link Specifications}
	\acro{LDIF}{Linked Data Integration Framework}
	\acro{LGD} {LinkedGeoData}
	\acro{LOD} {Linked Open Data}
	\acro{LOV} {Linked Open Vocabularies}
	\acro{LSTM}{Long Short-Term Memories}
	\acro{MSE}{Mean Squared Error}
	\acro{MWE}{Multiword Expressions}
	\acro{MT}{Machine Translation}
	\acro{ML}{Machine Learning}
	\acro{MR}{Machine Reading}
	\acro{NIF}{Natural Language Processing Interchange Format}
	\acro{NIF4OGGD}{NLP Interchange Format for Open German Governmental Data}
	\acro{NLP}{Natural Language Processing}
	\acro{NER}{Named Entity Recognition}
	\acro{NMT}{Neural Machine Translation}
	\acro{NN}{Neural Network}
	\acro{NLG}{Natural Language Generation}
	\acro{NED}{Named Entity Disambiguation}
	\acro{NERD}{Named Entity Recognition and Disambiguation}
	\acro{NL}{Natural Language}
	\acro{NIF}{NLP Interchange Format}
	\acro{NIF4OGGD}{NLP Interchange Format for Open German Governmental Data}
	\acro{NLP}{Natural Language Processing}
	\acro{NER}{Named Entity Recognition}
	\acro{NEL}{Named Entity Linking}
	\acro{NE}{Named Entity}
	\acro{NN}{Neural Network}
	\acro{NLI}{Natural Language Inference}
	\acro{OSM}{OpenStreetMap}
	\acro{OWL}{Web Ontology Language}
	\acro{OOV}{out-of-vocabulary}
	\acro{PFM}{Pseudo-F-Measures}
	\acro{PSO}{Particle-Swarm Optimization}
	\acro{PBSMT}{Phrase-Based Statistical Machine Translation}
	
	\acro{QA}{Question Answering}
	\acro{RDF}{Resource Description Framework}
	\acro{RBMT}{Rule-Based Machine Translation}
	\acro{RNN}{Recurrent Neural Network}
	\acro{ReLU}{rectified linear unit}
	\acro{RDFS}{RDF Schema}
	\acro{SKOS}{Simple Knowledge Organization System}
	\acro{SPARQL}{SPARQL Protocol and RDF Query Language}
	\acro{SRL}{Statistical Relational Learning}
	\acro{SWT}{Semantic Web Technologies}
	\acro{SW}{Semantic Web}
	\acro{SMT}{Statistical Machine Translation}
	\acro{SWMT}{Semantic Web Machine Translation}
	\acro{SKOS}{Simple Knowledge Organization System}
	\acro{SPARQL}{SPARQL Protocol and RDF Query Language}
	\acro{SRL}{Statistical Relational Learning}
	\acro{SF}{surface forms}

    \acro{TBMT} {Transfer-Based Machine Translation}
	\acro{UML}{Unified Modeling Language}
	\acro{USL}{Ukrainian Sign Language}
	\acro{URI}{Uniform Resource Identifier}
	\acro{WHO}{World Health Organization}
	\acro{WKT}{Well-Known Text}
	\acro{W3C}{World Wide Web Consortium}
	\acro{WSD}{Word Sense Disambiguation}
	\acro{WMT}{Workshop on Machine Translation}
    \acro{XML}{Extensible Markup Language}
	\acro{YPLL}{Years of Potential Life Lost}

	\acro{AOS}{Agricultural Ontology Services}
	\acro{AGRIS}{Agricultural Science and Technology}
	\acro{API}{Application Programming Interface}
	\acro{AOS}{Agricultural Ontology Services}
	\acro{AGRIS}{Agricultural Science and Technology}
	\acro{API}{Application Programming Interface}
	\acro{A2KB}{Annotation to Knowledge Base}
	\acro{BPSO}{Binary Particle-Swarm Optimization}
	\acro{BPMLOD}{Best Practices for Multilingual Linked Open Data}
	\acro{BPSO}{Binary Particle-Swarm Optimization}
	\acro{BPMLOD}{Best Practices for Multilingual Linked Open Data}
	\acro{BFS}{Breadth-First-Search}
	\acro{BPE}{Byte Pair Encoding}
	\acro{BoW}{Bag-of-Words}
	\acro{CBD}{Concise Bounded Description}
	\acro{COG}{Content Oriented Guidelines}
	\acro{CSV}{Comma-Separated Values}
	\acro{CBMT}{Corpus-Based Machine Translation}
	\acro{CLIR}{Cross-Language Information Retrieval}
	\acro{DPSO}{Deterministic Particle-Swarm Optimization}
	\acro{DALY}{Disability Adjusted Life Year}

	\acro{ER}{Entity Resolution}
	\acro{EM}{Expectation Maximization}
	\acro{EBMT}{Example-Based Machine Translation}
	\acro{EBNF}{Extended Backus--Naur Form}
	\acro{EL}{Entity Linking}
	\acro{FAO}{Food and Agriculture Organization of the United Nations}
	\acro{GIS}{Geographic Information Systems}
	\acro{GHO}{Global Health Observatory}
	\acro{GRU}{Gated recurrent unit}
	\acro{HDI}{Human Development Index}
	\acro{ICT}{Information and communication technologies}
	\acro{IFRS}{International Financial Reporting Standards}
	\acro{ICD}{International Classification of Diseases}
	\acro{IT}{Information Technology}
	\acro{IRI}{International Resource Identifier}
    \acro{KB}{Knowledge Base}
    \acro{KG}{Knowledge Graph}
    \acro{KGE}{Knowledge Graph Embeddings}
    \acro{KBSE}{Knowledge Base Semantic Embedding}
	\acro{LR}  {Language Resource}
	\acro{LD}  {Linked Data}
	\acro{LLOD}  {Linguistic Linked Open Data}
	\acro{LIMES}{LInk discovery framework for MEtric Spaces}
	\acro{LS}  {Link Specifications}
	\acro{LDIF}{Linked Data Integration Framework}
	\acro{LGD} {LinkedGeoData}
	\acro{LOD} {Linked Open Data}
	\acro{LSTM}{Long Short-Term Memories}
	\acro{MSE}{Mean Squared Error}
	\acro{MWE}{Multiword Expressions}
	\acro{MT}{Machine Translation}
	\acro{ML}{Machine Learning}
	\acro{MR}{Machine Reading}
	\acro{MOS}{Manchester OWL Syntax}
	\acro{NIF}{Natural Language Processing Interchange Format}
	\acro{NIF4OGGD}{NLP Interchange Format for Open German Governmental Data}
	\acro{NLP}{Natural Language Processing}
	\acro{NER}{Named Entity Recognition}
	\acro{NMT}{Neural Machine Translation}
	\acro{NN}{Neural Network}
	\acro{NLG}{Natural Language Generation}
	\acro{NED}{Named Entity Disambiguation}
	\acro{NERD}{Named Entity Recognition and Disambiguation}
	\acro{NL}{Natural Language}
	\acro{NIF}{NLP Interchange Format}
	\acro{NIF4OGGD}{NLP Interchange Format for Open German Governmental Data}
	\acro{NLP}{Natural Language Processing}
	\acro{NER}{Named Entity Recognition}
	\acro{NEL}{Named Entity Linking}
	\acro{NE}{Named Entity}
	\acro{NN}{Neural Network}
	\acro{NLI}{Natural Language Inference}
	\acro{OSM}{OpenStreetMap}
	\acro{OWL}{Web Ontology Language}
	\acro{OOV}{out-of-vocabulary}
	\acro{PFM}{Pseudo-F-Measures}
	\acro{PSO}{Particle-Swarm Optimization}
	\acro{PBSMT}{Phrase-Based Statistical Machine Translation}
	
	\acro{QA}{Question Answering}
	\acro{RDF}{Resource Description Framework}
	\acro{RBMT}{Rule-Based Machine Translation}
	\acro{RNN}{Recurrent Neural Network}
	\acro{ReLU}{rectified linear unit}
	\acro{SKOS}{Simple Knowledge Organization System}
	\acro{SPARQL}{SPARQL Protocol and RDF Query Language}
	\acro{SRL}{Statistical Relational Learning}
	\acro{SWT}{Semantic Web Technologies}
	\acro{SW}{Semantic Web}
	\acro{SMT}{Statistical Machine Translation}
	\acro{SWMT}{Semantic Web Machine Translation}
	\acro{SKOS}{Simple Knowledge Organization System}
	\acro{SPARQL}{SPARQL Protocol and RDF Query Language}
	\acro{SRL}{Statistical Relational Learning}
	\acro{SF}{surface forms}
	\acro{SVM}{Support Vector Machines}

    \acro{TBMT} {Transfer-Based Machine Translation}
	\acro{UML}{Unified Modeling Language}
	\acro{USL}{Ukrainian Sign Language}
	\acro{URI}{Uniform Resource Identifier}
	\acro{WHO}{World Health Organization}
	\acro{WKT}{Well-Known Text}
	\acro{W3C}{World Wide Web Consortium}
	\acro{WSD}{Word Sense Disambiguation}
	\acro{WWW}{World Wide Web}
    \acro{XML}{Extensible Markup Language}
	\acro{YPLL}{Years of Potential Life Lost}

	\acro{AOS}{Agricultural Ontology Services}
	\acro{AGRIS}{Agricultural Science and Technology}
	\acro{API}{Application Programming Interface}
	\acro{BPSO}{Binary Particle-Swarm Optimization}
	\acro{BPMLOD}{Best Practices for Multilingual Linked Open Data}
	\acro{CBD}{Concise Bounded Description}
	\acro{COG}{Content Oriented Guidelines}
	\acro{CSV}{Comma-Separated Values}
	\acro{CBMT}{Corpus-Based Machine Translation}
	\acro{CLIR}{Cross-Language Information Retrieval}
	\acro{DPSO}{Deterministic Particle-Swarm Optimization}
	\acro{DALY}{Disability Adjusted Life Year}

	\acro{ER}{Entity Resolution}
	\acro{EM}{Expectation Maximization}
	\acro{EBMT}{Example-Based Machine Translation}
	\acro{EBNF}{Extended Backus--Naur Form}
	\acro{EL}{Entity Linking}
	\acro{FAO}{Food and Agriculture Organization of the United Nations}
	\acro{GIS}{Geographic Information Systems}
	\acro{GHO}{Global Health Observatory}
	\acro{HDI}{Human Development Index}
	\acro{ICT}{Information and communication technologies}
    \acro{KB}{Knowledge Base}
    \acro{KBSE}{Knowledge Base Semantic Embedding}
	\acro{LR}  {Language Resource}
	\acro{LD}  {Linked Data}
	\acro{LLOD}  {Linguistic Linked Open Data}
	\acro{LIMES}{LInk discovery framework for MEtric Spaces}
	\acro{LS}  {Link Specifications}
	\acro{LDIF}{Linked Data Integration Framework}
	\acro{LGD} {LinkedGeoData}
	\acro{LOD} {Linked Open Data}
	\acro{MSE}{Mean Squared Error}
	\acro{MWE}{Multiword Expressions}
	\acro{MT}{Machine Translation}
	\acro{ML}{Machine Learning}
	\acro{NIF}{Natural Language Processing Interchange Format}
	\acro{NIF4OGGD}{NLP Interchange Format for Open German Governmental Data}
	\acro{NLP}{Natural Language Processing}
	\acro{NER}{Named Entity Recognition}
	\acro{NMT}{Neural Machine Translation}
	\acro{NN}{Neural Network}
	\acro{NLG}{Natural Language Generation}
	\acro{NED}{Named Entity Disambiguation}
	\acro{NERD}{Named Entity Recognition and Disambiguation}
	\acro{NL}{Natural Language}
	\acro{OSM}{OpenStreetMap}
	\acro{OWL}{Web Ontology Language}
	\acro{OOV}{out-of-vocabulary}
	\acro{PFM}{Pseudo-F-Measures}
	\acro{PSO}{Particle-Swarm Optimization}
	\acro{QA}{Question Answering}
	\acro{RDF}{Resource Description Framework}
	\acro{RBMT}{Ru\-le-Ba\-sed Ma\-chi\-ne Trans\-la\-tion}
	\acro{REG}{Referring Expression Generation}
	\acro{SKOS}{Simple Knowledge Organization System}
	\acro{SPARQL}{SPARQL Protocol and RDF Query Language}
	\acro{SRL}{Statistical Relational Learning}
	\acro{SWT}{Semantic Web Technologies}
	\acro{SW}{Semantic Web}
	\acro{SMT}{Statistical Machine Translation}
	\acro{SWMT}{Semantic Web Machine Translation}

    \acro{TBMT} {Transfer-Based Machine Translation}
	\acro{UML}{Unified Modeling Language}
	\acro{USL}{Ukrainian Sign Language}
	\acro{URL}{Uniform Resource Locator}
	\acro{WHO}{World Health Organization}
	\acro{WKT}{Well-Known Text}
	\acro{W3C}{World Wide Web Consortium}
	\acro{WSD}{Word Sense Disambiguation}
    \acro{XML}{Extensible Markup Language}
	\acro{YPLL}{Years of Potential Life Lost}

	\acro{AOS}{Agricultural Ontology Services}
	\acro{AGRIS}{Agricultural Science and Technology}
	\acro{API}{Application Programming Interface}
	\acro{AOS}{Agricultural Ontology Services}
	\acro{AGRIS}{Agricultural Science and Technology}
	\acro{API}{Application Programming Interface}
	\acro{A2KB}{Annotation to Knowledge Base}
	\acro{BPSO}{Binary Particle-Swarm Optimization}
	\acro{BPMLOD}{Best Practices for Multilingual Linked Open Data}
	\acro{BPSO}{Binary Particle-Swarm Optimization}
	\acro{BPMLOD}{Best Practices for Multilingual Linked Open Data}
	\acro{BFS}{Breadth-First-Search}
	\acro{BPE}{Byte Pair Encoding}
	\acro{BoW}{Bag-of-Words}
	\acro{CBD}{Concise Bounded Description}
	\acro{COG}{Content Oriented Guidelines}
	\acro{CSV}{Comma-Separated Values}
	\acro{CBMT}{Corpus-Based Machine Translation}
	\acro{CLIR}{Cross-Language Information Retrieval}
	\acro{DPSO}{Deterministic Particle-Swarm Optimization}
	\acro{DALY}{Disability Adjusted Life Year}

	\acro{ER}{Entity Resolution}
	\acro{EM}{Expectation Maximization}
	\acro{EBMT}{Example-Based Machine Translation}
	\acro{EBNF}{Extended Backus--Naur Form}
	\acro{EL}{Entity Linking}
	\acro{FAO}{Food and Agriculture Organization of the United Nations}
	\acro{GIS}{Geographic Information Systems}
	\acro{GHO}{Global Health Observatory}
	\acro{GRU}{Gated recurrent unit}
	\acro{HDI}{Human Development Index}
	\acro{ICT}{Information and communication technologies}
	\acro{IFRS}{International Financial Reporting Standards}
	\acro{ICD}{International Classification of Diseases}
	\acro{IT}{Information Technology}
    \acro{KB}{Knowledge Base}
    \acro{KG}{Knowledge Graph}
    \acro{KGE}{Knowledge Graph Embeddings}
    \acro{KBSE}{Knowledge Base Semantic Embedding}
	\acro{LR}  {Language Resource}
	\acro{LD}  {Linked Data}
	\acro{LLOD}  {Linguistic Linked Open Data}
	\acro{LIMES}{LInk discovery framework for MEtric Spaces}
	\acro{LS}  {Link Specifications}
	\acro{LDIF}{Linked Data Integration Framework}
	\acro{LGD} {LinkedGeoData}
	\acro{LOD} {Linked Open Data}
	\acro{LSTM}{Long Short-Term Memories}
	\acro{MSE}{Mean Squared Error}
	\acro{MWE}{Multiword Expressions}
	\acro{MT}{Machine Translation}
	\acro{ML}{Machine Learning}
	\acro{MR}{Machine Reading}
	\acro{NIF}{Natural Language Processing Interchange Format}
	\acro{NIF4OGGD}{NLP Interchange Format for Open German Governmental Data}
	\acro{NLP}{Natural Language Processing}
	\acro{NER}{Named Entity Recognition}
	\acro{NMT}{Neural Machine Translation}
	\acro{NN}{Neural Network}
	\acro{NLG}{Natural Language Generation}
	\acro{NED}{Named Entity Disambiguation}
	\acro{NERD}{Named Entity Recognition and Disambiguation}
	\acro{NL}{Natural Language}
	\acro{NIF}{NLP Interchange Format}
	\acro{NIF4OGGD}{NLP Interchange Format for Open German Governmental Data}
	\acro{NLP}{Natural Language Processing}
	\acro{NER}{Named Entity Recognition}
	\acro{NEL}{Named Entity Linking}
	\acro{NE}{Named Entity}
	\acro{NN}{Neural Network}
	\acro{NLI}{Natural Language Inference}
	\acro{OSM}{OpenStreetMap}
	\acro{OWL}{Web Ontology Language}
	\acro{OOV}{out-of-vocabulary}
	\acro{PFM}{Pseudo-F-Measures}
	\acro{PSO}{Particle-Swarm Optimization}
	\acro{PBSMT}{Phrase-Based Statistical Machine Translation}
	
	\acro{QA}{Question Answering}
	\acro{RDF}{Resource Description Framework}
	\acro{RBMT}{Rule-Based Machine Translation}
	\acro{RNN}{Recurrent Neural Network}
	\acro{ReLU}{rectified linear unit}
	\acro{SKOS}{Simple Knowledge Organization System}
	\acro{SPARQL}{SPARQL Protocol and RDF Query Language}
	\acro{SRL}{Statistical Relational Learning}
	\acro{SWT}{Semantic Web Technologies}
	\acro{SW}{Semantic Web}
	\acro{SMT}{Statistical Machine Translation}
	\acro{SWMT}{Semantic Web Machine Translation}
	\acro{SKOS}{Simple Knowledge Organization System}
	\acro{SPARQL}{SPARQL Protocol and RDF Query Language}
	\acro{SRL}{Statistical Relational Learning}
	\acro{SF}{surface forms}
	\acro{SVM}{Support Vector Machines}

    \acro{TBMT} {Transfer-Based Machine Translation}
	\acro{UML}{Unified Modeling Language}
	\acro{USL}{Ukrainian Sign Language}
	\acro{URI}{Uniform Resource Identifier}
	\acro{WHO}{World Health Organization}
	\acro{WKT}{Well-Known Text}
	\acro{W3C}{World Wide Web Consortium}
	\acro{WSD}{Word Sense Disambiguation}
    \acro{XML}{Extensible Markup Language}
	\acro{YPLL}{Years of Potential Life Lost}
\end{acronym}  
\mainmatter
\chapter{Introduction}
\label{ch:introduction}


The technological progress of recent decades has made both the distribution of and access to content in different languages simpler. Still, the Web has approximately 48\% of the pages unavailable in English.\footnote{\url{https://www.internetworldstats.com/stats7.htm}} Translation aims to support users who need to access content in a language in which they are not fluent~\citep{slocum1985survey, Koehn2010}.

However, translation is a difficult task due to the complexity and diversity of the natural language families~\citep{Jurafsky2000}. In addition, manual translation does not scale to the magnitude of the Web. One remedy for this problem is \ac{MT}. The main goal of \ac{MT} is to enable people to assess content in languages other than the languages in which they are fluent~\citep{Bar-Hillel1960}. From a formal point of view, this means that the goal of \ac{MT} is to transfer semantics from a piece of text in an input language to a piece of text in an output language~\citep{hutchins1992introduction}. At the time of writing, large information portals such as Google\footnote{http://translate.google.com.br/about/} or Bing\footnote{http://www.bing.com/translator/help/} already offer \ac{MT} services even though they are not entirely open-source.

\ac{MT} systems are now popular on the Web, but they still generate a large number of incorrect translations. The two most common types of errors are responsible for roughly 70\% of the translation errors: 40\% of the translation errors are the result of reordering errors, where an \ac{MT} system outputs sentences in a target language with incorrect word sequence. Another 30\% are due to lexical and syntactical ambiguity, i.e., when a single sentence or a word can have more than one meaning~\citep{moussallem2018machine}. Thus, addressing these barriers is a key challenge for modern translation systems.  


Recently, a novel \ac{SMT} paradigm has emerged called \ac{NMT}. \ac{NMT} relies on \ac{NN} algorithms. \ac{NMT} has been achieving significant improvements and is now the state of the art in \ac{MT} approaches. Since \ac{NMT} has shown impressive results on reordering \citep{stahlberg2019neural}, an important challenge in \ac{NMT} lies in the disambiguation process, both at the syntactic and semantic levels.
Additionally, \ac{NMT} approaches struggle with \ac{OOV} words (rare words) since they operate with a fixed vocabulary size. Although the community has been combining efforts to address this problem by proposing character-based~\citep{luong2016achieving,chung2016character} or \ac{BPE} models~\citep{sennrich2015neural}, \ac{OOV} words are still an open problem as they are highly co-related to the disambiguation of entities~\citep{koehn2017six}. Entities can be classified into two groups, i.e., proper nouns and common nouns. Proper nouns are also known as Named Entities (NE) and correspond to the name of people, organizations or locations, e.g., \textit{John}, \textit{WHO} or \textit{Canada}.  Common nouns describe classes of objects, e.g., \textit{spoon} or \textit{cancer}.

One possible solution to address the remaining issues of \ac{MT} regarding semantic ambiguity and \ac{OOV} words lies in the use of \acp{KG}, which have emerged over recent decades as a paradigm to make the semantics of data explicit so that it can be used by machines~\citep{Berners-Lee2001}. \ac{KG}s are a family of flexible knowledge representation paradigm intended to facilitate the processing of knowledge for both humans and machines. \acp{KG} (especially \acp{KG} in the \ac{RDF} format) commonly stores knowledge in triples. Each triple consists of 
\begin{enumerate}
    \item a subject which is often an entity.
    \item a relation which is often called a property.
    \item an object which is an entity or a literal.\footnote{a string or a value with a unit}
\end{enumerate}
For example, the following triple expresses that Albert Einstein was born in Ulm:
\begin{lstlisting}[language=ttl]
:Albert_Einstein :birthPlace :Ulm .
\end{lstlisting}
The explicit semantic knowledge in \ac{KG}s can enable \ac{MT} systems to supply translations with significantly better quality while maintaining the translation process scalable~\citep{heuss2013lessons}. In addition, the disambiguated knowledge about real-world entities, their properties, and relationships can potentially be used to infer the right meaning of ambiguous sentences or words as well as improve the performance of MT systems on the reordering task. 

Recent work has successfully exploited the apparent opportunity of using \ac{KG}s for the improvements of other \ac{NLP} tasks such as \ac{NLI}~\citep{annervaz2018}, \ac{QA}~\citep{Sorokincoling2018}, and \ac{MR}~\citep{yang2017leveraging}. According to \cite{moussallem2018machine}, the distinct opportunity of using \ac{KG}s for \ac{MT} has already been studied by several approaches. However, none had defacto implemented and used the benefits of \ac{KG}s in the training phase of \ac{NMT} before this work. 

\ac{NLG} is the task of automatically converting non-linguistic data into coherent natural language text \citep{reiter2000,gatt2017}. Recently, a new line of research has emerged, which relies on \ac{KG}s as input data. It has a task named RDF-to-Text, which generates texts from \ac{RDF} \acp{KG}~\citep{colin2016webnlg}. This task is an extension of \ac{MT} as understood classically given that it translates from a non-natural to a natural language.
Therefore, we envisage that it will help enhance the fluency in language translation.

In this thesis, we devise novel approaches that rely on \ac{KG}s to improve the disambiguation, translation, and generation of entities in texts. Section \ref{challenges} specifies the problems and identifies motivation and research challenges. Section~\ref{intro:overview} summarizes the contributions of the thesis. 


\section{Problem Specification and Challenges}
\label{challenges}

A large number of \ac{MT} approaches have been developed over the last two decades. For instance, translators began by using methodologies based on linguistics, which led to the family of \ac{RBMT}\citep{arnold1994machine}. However, \ac{RBMT} systems have a critical drawback in their reliance on manually crafted rules, thus making the development of new translation modules for different languages even more difficult as each language has its own syntax~\citep{costa2012study,thurmair2004comparing}. \ac{SMT} and \ac{EBMT} were developed to deal with the scalability issue in \ac{RBMT}~\citep{brown1990statistical}, a necessary characteristic of \ac{MT} systems that handle data at Web scale. Presently, these approaches have begun to address the drawbacks of rule-ba\-sed approaches. However, certain problems that had already been solved for \ac{RBMT} methods reappeared. The majority of these problems are connected to the issue of ambiguity, including syntactic and semantic variations~\citep{Koehn2010}. Subsequently, \ac{RBMT} and \ac{SMT} have been combined to resolve the drawbacks of these two fa\-mi\-lies of approa\-ches. This combination of methods is called hybrid \ac{MT}. Although hybrid approaches have been achieving good results, they still suffer from some of the limitations of \ac{RBMT}~\citep{costa2015latest,costa2015much,Thumair}. For example, the creation of manually crafted rules to handle syntax divergences. 


Below, we detail some key \ac{MT} challenges, which were unresolved when we began our work and still experienced by the \ac{MT} approaches aforementioned~\citep{moussallem2018machine}:

\begin{enumerate}

\item  \emph{Complex semantic ambiguity}: This challenge is mostly caused by the existence of homonyms, polysemous words, and named entities. Homonyms are different words that mean different things but share the same orthographic and phonological forms. For example, ``\texttt{bank}" can mean ``the land alongside or sloping down to a river or lake" or ``financial organization". Polysemous words are considered as the same word but with different, still related senses. For instance, ``\texttt{wood}" can refer to a piece of a tree or a collection of many trees. \ac{MT} systems commonly struggle to translate these words correctly, even if the models are built upon n-grams with large $n$ (e.g., 7-grams). Therefore, a significant amount of parallel data is usually necessary to translate such words and expressions adequately~\citep{moussallemlrec2018}. However, data is not only the main aspect to consider while learning translations. For example, some homonyms such as ``\texttt{kiwi}" can also refer to a named entity, and therefore it requires more specific learning features than a vast amount of parallel data to determine its correct meaning.




\item  \emph{Structural divergence}: By definition, structural reordering is reorganizing the order of the syntactic constituents of a language according to its original structure~\citep{bisazza2016survey}.
It, in turn, is a critical issue because fluency is one of the key aspects in the translation process. Every language has its own syntax. Thus an \ac{MT} system, which aims to translate a given language pair, needs to have an adequate model for the syntax of the involved languages. For instance, reordering a sentence from Japanese to English is one of the most challenging techniques because of the SVO (sub\-ject-verb-ob\-je\-ct) and SOV (sub\-ject-object-verb) word-order difference. One English word often groups multiple meanings of Japanese characters. For example, Kanji (Japanese) characters make subtle distinctions between homonyms that would not be clear in a phonetic language such as English. The following words,\begin{CJK}{UTF8}{min} 史 (history), 師 (teacher), 市 (a market or city), 矢 (arrow), 士 (a warrior or gentleman)\end{CJK} are pronounced as (shi), the same as ``she" (English feminine pronoun).

\item  \emph{Linguistic properties/features}: A large number of languages display a complex tense system. When con\-fron\-ted with sentences from such languages, it can be hard for \ac{MT} systems to recognize the current input tense and to translate the input sentence into the right tense in the target language. For instance, some irregular verbs in English like ``set'' and ``put'' cannot be determined to be in the present or past tense without previous knowledge or pre-processing techniques when translated to morphologically rich languages, e.g., Portuguese, German or Slavic languages. Additionally, the grammatical gender of words in such morphologically rich languages contributes to the problem of tense generation where a certain \ac{MT} system has to decide which inflection to use for a given word. This challenge is a direct consequence of the structural reordering issue and remains a significant problem for modern translator systems.
\end{enumerate}

Additionally, recent literature suggests 5 different challenges, which are described more generically below~\citep{lopez2013beyond}.

\begin{enumerate}
    \item Recent work focuses excessively on English and European languages as one of the involved languages in \ac{MT} approaches. In addition, there is a lack of research on low-resource language pairs such as African and/or South American languages.
    \item Previous work shows limitations of \ac{SMT} approaches for translating across domains. Most \ac{MT} systems exhibit good performance on legislative domains due to a large amount of data provided by the European Union. In contrast, translations performed on sports and life-hacks commonly fail because of the lack of training data.
    \item Few \ac{MT} approaches are able to translate non-standard speech texts from social networks (e.g., tweets). This kind of text poses several challenges for \ac{MT} systems, such as syntactic variations.
    \item There is a shortage of \ac{MT} approaches for translating morphologically rich languages. This challenge shares the same problem with the first one, namely the excessive focus on English as one of the involved languages. Therefore, \ac{MT} systems that translate content between, for instance, Arabic and Spanish, are rare.
    \item For the speech translation task, the bilingual parallel data, which are used for training the \ac{MT} models, differs widely from real user speech.
\end{enumerate}

The challenges above are clearly not independent, which means that addressing one of them can have an impact on the others. We focus on the portions of these problems related to entities. Entities are found in a \ac{KG}, where they are described within triples~\citep{auer2007dbpedia,vrandevcic2014wikidata}. It is already clear that the real benefit of \ac{KG}s comes from their capacity to provide unseen knowledge about emergent data, which appears every day. Thus, our central research question can be stated as follows:
{\begin{mycolorrq}
\begin{itemize}
\item[RQ.] Can \acp{KG} alleviate the ambiguity problem and be used to improve the quality of automatic text translation and generation? 
\end{itemize}
\end{mycolorrq}}
In the following subsections, we present the challenges that need to be tackled to answer our central research question.

\subsection{Challenge 1: Multilingual Entity Disambiguation}
\label{challengeNED}

Understanding the \ac{EL} task in a multilingual environment is the first step to discern how to deal with entities in text translation and generation. One of the most important \ac{MT} tasks is \ac{EL}, also known as \ac{NED}. The goal of \ac{EL} is the disambiguation of entities and common words (concepts and terminologies) in texts. Disambiguation refers to the process of removing the ambiguity of words by identifying their single semantic meaning for a particular context, in our case entities. Formally, the goal of \ac{EL} algorithm is as follows: given a piece of text, a reference knowledge base $K$, and a set of entity mentions in that text, map each entity mention to the corresponding resource in $K$. Several challenges have to be addressed when implementing an \ac{EL} system. First, an entity can have a large number of \ac{SF} (also known as labels) due to synonymy, acronyms, and typos. For example, \texttt{New York City}, \texttt{NY} and \texttt{Big Apple} are labels for the same entity. Moreover, multiple entities can share the same name due to homonymy and ambiguity. For example, both the state and the city of New York are called \texttt{New York}. 

Despite the complexity of the task, \ac{EL} approaches have recently been achieving increasingly better results by relying on trained machine learning models~\citep{roder2017gerbil}. A portion of these approaches claim to be multilingual, and most of them rely on models that are trained on English corpora with cross-lingual dictionaries. However, these underlying models being trained on English corpora make them prone to errors when migrated to a different language. Additionally, such approaches rarely make their models or data available on more than three languages due to the lack of training data~\citep{roder2017gerbil}.

A large number of multilingual approaches have been developed over recent years\-~\citep{PBOH}. However, to the best of our knowledge, no work has investigated the real disambiguation capability of \ac{KG}s in a broader multilingual and deterministic context. Thus, our first goal is to investigate the disambiguation task based on \ac{KG}s and analyze whether they can contribute to the translation of entities. Hence, we derive the following research questions:
{\begin{mycolorrq}
\begin{itemize}
\item[RQ1.] Can a \ac{KG}-based \ac{EL} approach achieve a similar F-score performance across languages? 
\item[RQ2.] Does a language-based \ac{KG} influence the disambiguation quality of entities in multilingual sentences?
\end{itemize}
\end{mycolorrq}}

\subsection{Challenge 2: Text Generation with Entities}
\label{challengeNLG}

The input data in RDF-to-Text consists of entities and the relations between them, therefore generating references for these entities is a core task in many NLG systems \citep{krahmer2012}. \ac{REG}, the task responsible for generating these references, is typically presented as a two-step procedure. First, the referential form needs to be chosen, asking whether a reference at a given point in the text should assume the form of, for example, a proper noun (``Stephen Hawking''), a pronoun (``he/him/his'') or description (``the physicist''). Second, the REG model must account for the different ways in which a particular referential form can be realized. For example, both ``Stephen'' and ``Hawking'' are name variants of Stephan Hawking that may occur in a text. He can also alternatively be described as, say, ``the brilliant scientist''.

A generic \ac{NLG} pipeline is composed of three tasks - \emph{document planing}, \emph{micro planning} and \emph{realization}. Before generating the respective referring expressions for the entities, several steps have to be taken into account. For example, \autoref{lst:example} shows a fragment of Stephen Hawking sub-\ac{KG}\footnote{\url{http://dbpedia.org/resource/Stephen_Hawking}} which represents the following information: \emph{``Stephen Hawking was a scientist who worked in physics. He was born in Oxford and died in Cambridge."}. 

\begin{lstlisting}[label=lst:example,language=ttl,caption=An excerpt of RDF triples.]
:Stephen_Hawking :type :Scientist
:Stephen_Hawking :deathPlace :Cambridge 	
:Stephen_Hawking :field :Physics
:Stephen_Hawking :birthPlace :Oxford 	
\end{lstlisting}

Even though the generation of natural language from \ac{KG}s has gained substantial attention~\citep{colin2016webnlg}, English is the only language that has been widely targeted. Only a few authors (e.g., ~\cite{keet2017toward} for IsiZhulu) have exploited the generation of other languages. Consequently, there is a lack of multilingual approaches for the generation of texts from \ac{RDF} \acp{KG}. Additionally, most of the earlier \ac{REG} approaches focus either on selecting referential forms \citep{orita2015,ferreira2016b}, or on selecting referential content, typically zooming in on one specific kind of reference such as pronouns~\citep{henschel2000,callaway2002}, definite descriptions \citep{dale1991}, or proper noun generations \citep{siddharthan2011,deemter2016,ferreira2017}. Therefore, no previous work has addressed the full REG task, which given a number of entities in a text, produces corresponding referring expressions by simultaneously selecting both form and content. Moreover, in previous models, notions such as \textit{salience} play a central role, where it is assumed that entities, which are salient in the discourse, are more likely to be referred to using shorter referring expressions (like a pronoun) than less salient entities, which are typically referred to using longer expressions (like full proper nouns).

Although some basic linguistics mistakes have been solved by Neural Network-based approaches, the lack of complex models for linguistic rules still causes ambiguity problems in text generation (e.g., errors on re\-la\-ti\-ve pro\-nouns)\-~\citep{bisazza2016survey}. The issues mentioned above leads to the following research question:
{\begin{mycolorrq}
\begin{itemize}
\item[RQ3:] Can \ac{KG}s as input support the generation of multilingual text?
\item[RQ4:] Can \ac{KG}s be used for accomplishing the full \ac{REG} task?
\end{itemize}
\end{mycolorrq}}
\subsection{Challenge 3: Entity Translation in Texts}
\label{challengeNMT}

Entities are a common and arduous problem across different \ac{NLP} tasks. Regarding \ac{MT}, \ac{NE}'s primary issue is caused by common words from a source language that are used as proper nouns in a target language. For instance, the word ``Kiwi" is a family name in New Zealand which comes from the M\=aori culture, but it also can be a fruit, a bird, or a computer program. Most words have multiple interpretations depending on the context in which they are mentioned. In the \ac{MT} field, \ac{WSD} techniques involve finding the respective meaning and correct translation to these ambiguous words in target languages. This ambiguity problem was identified early in \ac{MT} development. In 1960, Bar-Hillel~\citeyearpar{Bar-Hillel1960} stated that an \ac{MT} system is not able to find the right meaning without specific knowledge. Although the ambiguity problem has been lessened significantly since the contribution of Carpuat and subsequent works~\citep{carpuat2007improving,navigli2009word,costa2014statistical}, this problem remains a challenge. 

According to \cite{moussallem2018machine}, \ac{KG}s were applied mainly to the output translation of \ac{PBSMT} approaches in the target language as a post-editing technique. Although applying this technique has increased the quality of a translation, it is tedious to implement when  common words have to be translated instead of named entities, then be applied several times to achieve a successful translation. In \ac{MT} systems, dealing with entities is directly related to the ambiguity problem. Therefore, we argue that the entity problem has to be resolved in that broader context.

Recently, \ac{NMT} models have shown significant improvements in translation and have been widely adopted given their sustained improvements over the previous state-of-the-art \ac{PBSMT} approaches~\citep{koehn2007moses}. A number of \ac{NN} architectures have therefore been proposed in the recent years, ranging from recurrent~\citep{bahdanau2014neural,sutskever2014sequence} to self-attentional networks~\citep{vaswani2017attention}. A given \ac{NMT} model is basically trained to maximize the likelihood of each token in the target sentence, by taking into account the source sentence and the previous target tokens as input.
However, a major drawback of \ac{NMT} models is that they need large amounts of training data to return adequate results and have a limited vocabulary size due to their computational complexity~\citep{luong2016achieving}. The data sparsity problem in \ac{MT}, which is mostly caused by a lack of training data, manifests itself particularly in the poor translation of rare and \ac{OOV} words, e.g., entities or terminological expressions rarely or never seen in the training phase. 

Previous work has attempted to deal with entities and the data scarcity by introducing character-based models~\citep{luong2016achieving} or \ac{BPE} algorithms~\citep{sennrich2015neural}. Additionally, different strategies were developed for overcoming the lack of training data, such as back-translation~\citep{sennrich2016improving}, which relies on the use of monolingual data being translated by a different \ac{NMT} model and added as additional synthetic training data. Moreover, the benefits of incorporating type information on entities---e.g., \ac{NE}-tags such as \texttt{PERSON}, \texttt{LOCATION} or \texttt{ORGANIZATION}---into \ac{NMT} by relying on \ac{NER} systems have been shown in previous works~\citep{ugawa2018neural,li2018named}. Despite the significant advancement of previous work in \ac{NMT}, translating entities and terminological expressions remains a challenge~\citep{koehn2017six} and none of the above mentioned approaches have exploited the application of \ac{KG}s in \ac{NMT} systems. Hence it leads to the following research question:
{\begin{mycolorrq}
\begin{itemize}
\item[RQ5:] Can an \ac{NMT} model enhanced with a bilingual \ac{KG} improve translation quality?
\end{itemize}
\end{mycolorrq}}
\subsection{Challenge 4: Low-resource Knowledge Graphs}
\label{challengeKG}

Considerable amounts of partly human effort have been invested in making \ac{KG}s available across languages. However, even popular \ac{KG}s like DBpedia and Wikidata are most abundant in their English version~\citep{lakshen2018challenges}. Additionally, region-specific facts are often limited to the \ac{KG} specific to the region from which they emanate or to the \ac{KG} in the language spoken in said region~\citep{aprosio2013towards}. This lack of multilingual knowledge availability limits the porting of \ac{NLP} tasks such as \ac{EL}, \ac{NLG}, and \ac{NMT} to different languages. 

Previous works have tried to address the translation of \ac{KG}s by carrying out a localization task that relies on \ac{SMT} systems for translating the labels of \ac{KG}s into target languages. This kind of approach ignores an essential part of a \ac{KG}, namely its graph structure. For example, considering a highly ambiguous label in DBpedia \ac{KG} such as \textit{Kiwi}, an \ac{MT} system has to predict in which sub-\ac{KG} domain this word has to be translated in the target language. Otherwise, \emph{Kiwi} can be erroneously translated to the common term for inhabitants of New Zealand,\footnote{\url{http://dbpedia.org/resource/Kiwi_(people)}}  or a bird,\footnote{\url{http://dbpedia.org/resource/Kiwi}} thus affecting the structure and alignment quality of the translated \ac{KG}. These domains can be derived in \ac{KG}s through predicates such as type predicates (i.e., \texttt{rdf:type} in \ac{RDF}). Taking the graph structure of \ac{KG} into account can support an \ac{MT} system when spotting the correct translation for ambiguous labels. Few works have designed approaches to tackle this problem. Hence we investigate the following research questions:
{\begin{mycolorrq}
\begin{itemize}
\item[RQ6:]  Can \ac{NMT} support a full (\ac{URI}s and labels) translation of \ac{KG}s?
\item[RQ7:]  Can an artificially enriched \ac{KG} improve the performance of a system on \ac{NLP} tasks?
\end{itemize}
\end{mycolorrq}}
\section{Thesis Overview}
\label{intro:overview}

\subsection{Contributions}

In the following, our contributions are summarized according to each of the challenges aforementioned.


{\begin{myboxch}
\textbf{Challenge 1:} Multilingual Entity Disambiguation
\end{myboxch}}
\textbf{Contribution 1:} This drawback is addressed by presenting a novel multilingual, knowledge-base agnostic and deterministic approach to entity linking, dubbed MAG. MAG is based on a combination of context-based retrieval on structured knowledge bases and graph algorithms. We evaluate MAG on 23 datasets and in 7 languages. Our results show that MAG achieves state-of-the-art performance on English datasets and outperforms all other approaches on non-English languages \citep{moussallem2017mag}. Further, we extend MAG to 40 languages and deploy two versions as demos - one using DBpedia, another Wikidata \citep{moussallem2018entity}. The demos answer on average more than 170,000 requests per year.
{\begin{myboxch}
\textbf{Challenge 2:} Text Generation with Entities
\end{myboxch}}
\textbf{Contribution 2:} We address this research gap by presenting RDF2PT, an approach that verbalizes \ac{RDF} data to Brazilian Portuguese. We evaluate RDF2PT in an open questionnaire, with 44 native speakers divided into experts and non-experts. Our results suggest that RDF2PT is able to generate text similar to that generated by humans and can hence be easily understood \citep{rdf2pt_lrec_2018}. Afterward, we extend RDF2PT to Spanish \citep{bengal} and English~\citep{Ngonga2019}. 

\textbf{Contribution 3:} Traditionally, \ac{REG} models first decide on the form and then on the content of references to discourse entities in text and rely thereby on features such as salience and grammatical function. No previous work has investigated either how to tackle both sub-tasks at once or use \ac{RDF} \ac{KG} as input to this task. We handle this problem by presenting the first approach relying on deep neural networks, which makes decisions about form and content in one go without explicit feature extraction. Using \ac{RDF} \ac{KG}~\citep{bonatti2019knowledge}, the neural model substantially improves over two strong baselines~\citep{moussallem2018neuralreg}. We also extend our training data to the German language, making it able to generate referring expressions in German~\citep{moussallem2018enriching}.
{\begin{myboxch}
\textbf{Challenge 3:} Entity Translation in Texts
\end{myboxch}}
\textbf{Contribution 4:} While neural networks have led to substantial progress in machine translation, their success depends heavily on large amounts of training data. However, parallel training corpora are not always readily available. Out-of-vocabulary words, mostly entities and terminological expressions, pose a difficult challenge to \ac{NMT} systems.
We alleviate this problem by implementing the first \ac{KG}-augmented \ac{NMT} model, named KG-NMT. We use knowledge graph embeddings to enhance the semantic feature extraction of neural models. Thus, this approach optimize the translation of entities and terminological expressions in texts, consequently leading to better translation quality. Our knowledge-graph-augmented neural translation model, dubbed \textit{KG-NMT}, achieve significant and consistent improvements of +3 {\sc BLEU}, {\sc METEOR} and {\sc chrF3} on average on the \textit{newstest} datasets between 2015 and 2018 for the WMT English-German translation task~\citep{moussallem2019augmenting}.
{\begin{myboxch}
\textbf{Challenge 4:} Low-resource Knowledge Graphs
\end{myboxch}}
\textbf{Contribution 5:} We address the current limitations of knowledge graphs w.r.t. multilinguality by proposing THOTH, the first full neural-based approach for translating and enriching knowledge graphs across languages. THOTH extracts bilingual alignments between a source and target knowledge graph and learns how to translate from one to the other by relying on two different recurrent neural network models along with knowledge graph embeddings. We evaluate THOTH extrinsically by comparing the German DBpedia with the German translation of the English DBpedia on two tasks: fact checking and entity linking. In addition, we run a manual intrinsic evaluation of the translation. Our results show that THOTH is a promising approach since it achieves a translation accuracy of 88.56\%. Moreover, its enrichment improves the quality of the German DBpedia significantly, as we report +18.4\% accuracy for fact validation and +19\% F$_1$ for entity linking~\citep{moussallem2019thoth}.
\newline
\newline
The main contributions of this thesis can be summarized as follows:
{\begin{mybox}
\begin{enumerate}

\item A multilingual, knowledge-base agnostic and deterministic entity-linking approach for 40 languages ~\citep{moussallem2017mag,moussallem2018entity}

\item A multilingual \ac{RDF}-to-text approach that works on Brazilian Portuguese, Spanish and English and is virtually extensible for German, French and Italian~\citep{rdf2pt_lrec_2018} 

\item The first \ac{NN} model for tackling the full \ac{REG} task by using \ac{KG}s~\citep{moussallem2018neuralreg}

\item The first \ac{NMT} model augmented with \ac{KG}s~\citep{moussallem2019augmenting}

\item The first full \ac{KG} translation and enrichment approach based on \ac{NN} models~\citep{moussallem2019thoth}

\end{enumerate}
\end{mybox}}

\subsection{Structure}

Having motivated our work in this chapter, Chapter \ref{ch:contributions} summarizes the main contributions of the thesis and presents them in a coherent way. Chapter \ref{ch:conclusion} concludes the thesis with a summary and outlook on future research directions. The publications underlying this cumulative thesis can be found in Appendix A along with a detailed breakdown of the contributions of individual authors in Appendix B.

\chapter{Contributions}
\label{ch:contributions}

This chapter describes the main contributions of this thesis: in Section \ref{sec:mag}, we present a multilingual \ac{EL} approach; in Section~\ref{sec:rdf2pt}, we unveil a Brazilian Portuguese \ac{RDF}-\ac{KG}-based \ac{NLG} approach; in Section~\ref{sec:neuralREG}, we discuss a neural-based \ac{REG} model; in Section~\ref{sec:KG-NMT}, we give some insights into a \ac{KG}-augmented \ac{NMT} model; in Section~\ref{sec:THOTH}, we develop a neural-based approach for translating and enriching \ac{KG}s. 

\section{MAG: A Multilingual, Knowledge-base Agnostic and Deterministic Entity Linking Approach}
\label{sec:mag}
\begin{figure*}
\centering
\includegraphics[width=0.9\textwidth]{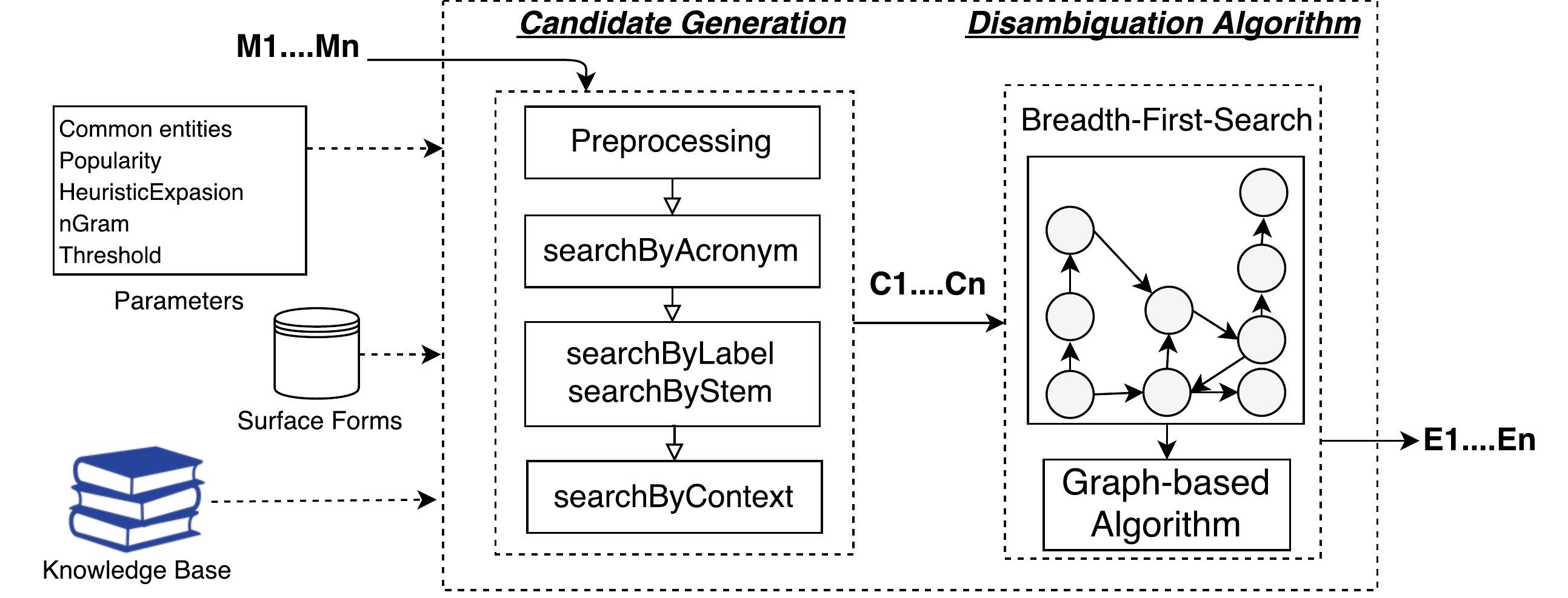}
\caption{MAG architecture overview.}
\label{fig:architecture}
\end{figure*}

{\small\textit{FOR ALLEVIATING THE LACK OF MULTILINGUAL \ac{EL} APPROACHES}~\ref{challengeNED}} , we devised a multilingual, knowledge-base agnostic and deterministic approach, named MAG. The \ac{EL} process implemented by MAG consists of two phases. Several indexes are generated during the offline phase. The entity linking per se is carried out during the online phase and consists of two steps: 1) candidate generation and 2) disambiguation. An overview can be found in~\autoref{fig:architecture}.

\subsection{Offline Index Creation}
\label{sec:index}
MAG relies on the following five indexes: surface forms, person names, rare references, acronyms and context.

\textbf{Surface forms.}
MAG relies exclusively on structured data to generate surface forms for entities so as to remain KB-agnostic. For each entity in the reference \ac{KB}, our approach harvests all labels of the said entity as well as its type and indexes them.
Additional \ac{SF}s can be collected from different sources~\citep{AGDISTIS_ISWC,bryl2015gathering}.

\textbf{Person names} - This index accounts for the variations in names for referencing persons~\citep{krahmer2012computational} across languages and domains. Persons are referred to by different portions of their names. For example, the artist \texttt{Beyonc\'e Giselle Knowles-Carter} is often referred to as \texttt{Beyonc\'e} or \texttt{Beyonc\'e Know\-les}. 
Moreover, languages such as Chinese and Japanese put the family name in front of the given name (in contrast to English, where names are written in the reverse order). Our technique handles the problem of labelling persons by generating all possible permutations of the words within the known labels of persons and adding them to the index of names.

\textbf{Rare references} - This index is created if textual descriptions are available for the resources of interest (e.g., if resources have a \texttt{rdfs:comment} property). A large number of textual entity descriptions provide type information pertaining to the resource at hand, as in the example ``Michael Joseph Jackson was an American singer ..."\footnote{See \texttt{rdfs:comment} of \url{http://dbpedia.org/resource/Michael_Jackson}.}. Hence, we use different language versions of the Stanford POS tagger~\citep{toutanova2000enriching} and TreeTagger\footnote{\url{https://www.cis.uni-muenchen.de/~schmid/tools/TreeTagger/}} on the first line of a resource's description and collect any noun phrase that contains an adjective. For example, we can extract the supplementary \ac{SF} \texttt{American singer} for our example. 

\textbf{Acronyms} - Acronyms are used across a large number of domains, e.g., in news (see AIDA and MSNBC datasets). We thus reuse a handcrafted index from \-STANDS4.\footnote{See \url{http://www.abbreviations.com/}} 

\textbf{Context} - Our context index relies on the \ac{CBD}\footnote{\url{https://www.w3.org/Submission/CBD/}} of resources. The literals found in the \ac{CBD} of each resource are first freed of stop words. Then, each preprocessed string is added as an entry that maps to the said resource.

\subsection{Candidate Generation}
\label{sec:candidates}

The candidate generation and the disambiguation steps occur online, i.e., when MAG is given a document and a set of mentions to disambiguate. The goal of the candidate generation step is to retrieve a tractable number of candidates for each of the mentions. These candidates are later inserted into the disambiguation graph, which is used to determine the mapping between entities and mentions (see Section \ref{sec:disambiguation}).  
    
First, we \textbf{preprocess mentions} to improve the retrieval quality using well-known pre-processing \ac{NLP} techniques such as regular expressions, lemmatization, stemming and true casing. 


The second step of the candidate generation, the \textbf{candidate search}, is divided into three parts:

\textbf{By Acronym} - If a mention is considered an acronym by our preprocessing, we expand the mention with the list of possible names from the acronym index mentioned above. For example, ``PSG" is replaced by ``Paris Saint-Germain".

\textbf{By Label} -  
First, MAG retrieves candidates for a mention using exact matches to their respective principal reference. For example, the mention ``Barack Obama" and the principal reference of the former president of the USA, which is also ``Barack Obama", match exactly. In cases it finds a string similarity match with the main reference of 1.0, the remaining steps are skipped. If this search does not return any candidates, MAG starts a new search using a trigram similarity threshold $\sigma$ over the \ac{SF} index. In cases where the set of candidates is still empty, MAG stems the mention and repeats the search.
For example, MAG stems ``Northern India" to ``North India" to account for linguistic variability. 

\begin{figure}
\centering
\includegraphics[width=0.65\textwidth]{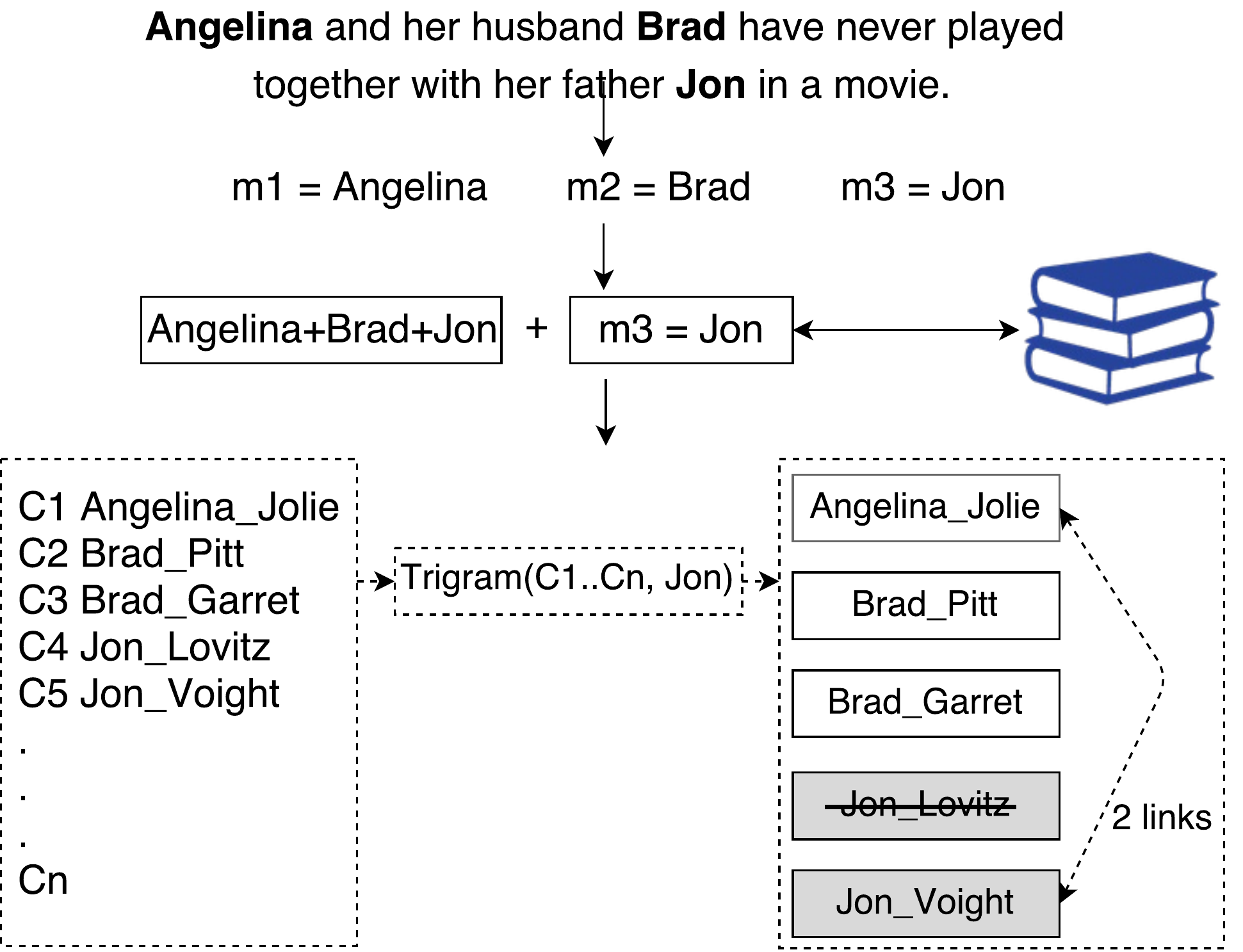}
\caption{Search using the context index. White boxes on the right side depict candidates discarded by the trigram filter.}
\label{fig:context}
\end{figure}

\textbf{By Context} - Here, two \textbf{post-search filters} are applied to find possible candidates from the context index. 
Before applying both filters, MAG extracts all entities contained in the input document.
These entities are used as an addition while searching a mention in the context index.
This search relies on TF-IDF~\citep{ramos2003using} which reflects the importance of a word or string in a document corpus relative to the relevance in its index.
Afterwards, MAG first filters unlikely candidates by applying trigram similarity. Second, MAG retrieves all direct links among the remaining candidates in the \ac{KB}.
Our approach uses the number of connections to find highly related entity sets for a specific mention. 
This is similar to finding a dense subgraph~\citep{AIDA}.
\autoref{fig:context} illustrates an example which contains three ambiguous entities, namely ``Angelina", ``Brad" and ``Jon". Regarding the mention ``Jon", MAG searches the context index using ``[(Angelina + Brad + Jon) + Jon]" as a query. MAG keeps only ``Jon\_Lovitz" and ``Jon\_Voight" after trigram filtering. Only ``Jon\_Voight", the father of ``Angelina\_Jolie", has direct connections with the other candidates and is thus chosen.

To improve the quality of candidates, ranking entities according to their popularity is an appropriate factor. If MAG makes use of this ranking configuration, the number of candidates retrieved from the index is increased and then the result is sorted. Afterward, MAG returns the top 100 candidates. The popularity is calculated using Page Rank~\citep{page1999pagerank} over the underlying \ac{KB}.In case it is unable to leverage Page Rank on certain KB, it falls back to a heuristic of inlinks and outlinks. 

\subsection{Entity Disambiguation Algorithm}
\label{sec:disambiguation}

After the candidate generation step, the computation of the optimal candidate to mention assignment starts by constructing a disambiguation graph $G_d$ with depth $d$ similar to the approach of AGDISTIS.
\begin{definition}
Knowledge Base: We define \ac{KB} $K$ as a directed graph $G_K = (V, E)$ where the nodes $V$ are resources of $K$, the edges $E$ are properties of $K$ and $x,y\in V, (x,y) \in E \Leftrightarrow \exists p : (x, p, y) \mbox{ is a triple in }K$.
\end{definition}
Given the set of candidates $C$, we begin by building an initial graph $G_0 = (V_0, E_0)$ where $V_0$ is the set of all resources in $C$ and $E_0=\emptyset$. Starting with $G_0$ the algorithm expands the graph using \ac{BFS} technique in order to find hidden paths among candidates. The extension of a graph is $G_i = (V_i, E_i)$ to a graph $\rho(G_i) = G_{i+1} = (V_{i+1}, E_{i+1})$ with $i=0, \ldots, d$.
The $\rho$ (\ac{BFS}) operator iterates $d$ times on the input graph $G_0$ to compute the initial disambiguation graph $G_d$. 
After constructing $G_d$, it needs to identify the correct candidate node for a given mention. Here, we rely on HITS~\citep{HITS} or Page Rank~\citep{page1999pagerank} as disambiguation graph algorithms. This choice comes from a comparative study of the differences between both~\citep{devi2014comparative}.

\textbf{HITS} uses hub and authority scores to define a recursive relationship between nodes. An authority node is a node that many hubs link to and a hub is a node that links to many authorities. The authority values are equal to the sum of the hub scores of each node that points to it. The hub values are equal to the sum of the authority scores of each node that it points to. According to previous work~\citep{AGDISTIS_ISWC}, we chose 20 iterations for HITS which suffice to achieve convergence in general.

\textbf{Page Rank} has a wide range of implementations. We implemented the general version in accordance with~\citep{page1999pagerank}. Thus, we defined the possibility of jumping from any node to any other node in the graph during the random walk with a probability $\alpha = (1-w) = 0.15$. We empirically chose 50 Page Rank iterations which has shown to be a reasonable number for \ac{EL}~\citep{doser}. We assigned a standard weight $w = 0.85$ for each node. Finally, the sum is calculated by spreading the current weight divided by outgoing edges.

Independent of the chosen graph algorithm, the highest candidate score among the set of candidates $C$ is chosen as correct disambiguation for a given mention $m_i$.
Note, MAG also considers {emergent entities}~\citep{Hoffart:2014:DEE:2566486.2568003} and assigns a new URI to them.\footnote{\url{https://www.w3.org/TR/cooluris/}}

\subsection{Evaluation}
\label{sec:evaluation}

We measured the performance of MAG on 17 datasets and compared it to the state of the art for \ac{EL} in English. Second, we evaluated MAG's portability to other languages. To this end, we compared MAG and the multilingual state of the art using 6 datasets from different languages. For both evaluations we use HITS and Page Rank. Third, we carried out a fine-grained evaluation providing a deep analysis of MAG using the method proposed in~\citep{waitelonis2016don}. Throughout our experiments, we used DBpedia as reference \ac{KB}. For our overall evaluation, we relied on the GERBIL platform \citep{gerbil} and integrated all datasets into it for the sake of comparability.

\subsection{Results}
\label{sec:results_mag}

\textit{\textbf{On English datasets.}} The English results are shown in the first part of \autoref{engfmeasure}. An analysis of our results shows that although the acronym index is an interesting addition for potential improvements, its contribution amounts only to 0.05{\%} F-measure on average over all datasets. Also, the {popularity} feature improves the results in almost every data set. It can be explained by the analysis of \citep{waitelonis2016don}, which demonstrates that most datasets were created using more popular entities as mentions.
Thus, this bias eases their retrieval\footnote{see the results without popularity using HITS~\url{http://gerbil.aksw.org/gerbil/experiment?id=201701220014}}.
HITS has shown better results on average than Page Rank.\footnote{\url{http://gerbil.aksw.org/gerbil/experiment?id=201701240030}} However, Page Rank did show promising results in some datasets (e.g., Spotlight corpus, AQUAINT, and N3-RSS-500). MAG using HITS outperformed the other approaches on 4 of the 17 datasets while achieving comparable results on others, e.g., ACE2004, MSNBC, and OKE datasets.
 
\begin{table*}[htb]
\fontsize{9.5pt}{9pt}\selectfont
\setlength\tabcolsep{2pt} 
\centering
\caption{Micro F-measure across approaches. Red entries are the top scores while blue represents the second best scores.}
\label{engfmeasure}
\begin{tabular}{@{}l|lclllllllllll|l|l@{}}
\rot{\textbf{Language}} & \textbf{Tools/datasets}  & \rot{\textbf{AGDISTS}} & \rot{\textbf{AIDA}} & \rot{\textbf{Babelfy}} & \rot{\textbf{DBpedia}} & \rot{\textbf{DoSer}} & \rot{\textbf{entityclassifier.eu}} & \rot{\textbf{FRED}} & \rot{\textbf{Kea}} & \rot{\textbf{NERD-ML}} & \rot{\textbf{PBOH}} & \rot{\textbf{WAT}} & \rot{\textbf{xLisa}} & \rot{\textbf{MAG + HITS}} & \rot{\textbf{MAG + PR}} \\ \toprule
\multirow{17}{*}{\rotatebox{90}{English}}

& ACE2004 & 0.65 & 0.70 & 0.53 & 0.48 & {\color[HTML]{FE0000} \textbf{0.75}} & 0.50 & 0.00 & 0.66 & 0.58 & {\color[HTML]{2033FF} \textbf{0.72}} & 0.66 & 0.70 & 0.69 & 0.60 \\ 

&AIDA/CoNLL-Complete & 0.55 & 0.68 & 0.66 & 0.50 & 0.69 & 0.50 & 0.00 & 0.61 & 0.20 & {\color[HTML]{FE0000} \textbf{0.75}} & {\color[HTML]{2033FF} \textbf{0.71}} & 0.48 & 0.59 & 0.54\\ 

&AIDA/CoNLL-Test A & 0.54 & 0.67 & 0.65 & 0.48 & 0.69 & 0.48 & 0.00 & 0.61 & 0.00 & {\color[HTML]{FE0000} \textbf{0.75}} & {\color[HTML]{2033FF} \textbf{0.7}} & 0.45 & 0.59 & 0.54\\ 

&AIDA/CoNLL-Test B & 0.52 & 0.69 & 0.68 & 0.52 & 0.69 & 0.48 & 0.00 & 0.61 & 0.00 & {\color[HTML]{FE0000} \textbf{0.75}} & {\color[HTML]{2033FF} \textbf{0.72}} & 0.47 & 0.57 & 0.52\\ 

&{AIDA/CoNLL-Training} & 0.55 & 0.69 & 0.66 & 0.50 & 0.69 & 0.52 & 0.00 & 0.61 & 0.28 & {\color[HTML]{FE0000} \textbf{0.75}} & {\color[HTML]{2033FF} \textbf{0.71}} & 0.48 & 0.60 & 0.55\\ 

&{AQUAINT} & 0.52 & 0.55 & 0.68 & 0.53 & {\color[HTML]{FE0000} \textbf{0.82}} & 0.41 & 0.00 & 0.78 & 0.60 & {\color[HTML]{2033FF} \textbf{0.81}} & 0.73 & 0.76 & 0.67 & 0.68\\ 

&{Spotlight} & 0.27 & 0.25 & 0.52 & 0.71 & {\color[HTML]{FE0000} \textbf{0.81}} & 0.25 & 0.04 & 0.74 & 0.56 & {\color[HTML]{2033FF} \textbf{0.79}} & 0.67 & 0.71 & 0.65 & 0.66\\ 

&{IITB} & 0.47 & 0.18 & 0.37 & 0.30 & 0.43 & 0.14 & 0.00 & {\color[HTML]{2033FF} \textbf{0.48}} & 0.43 & 0.38 & 0.41 & 0.27 & {\color[HTML]{FE0000} \textbf{0.52}} & 0.43\\ 

&{KORE50} & 0.27 & {\color[HTML]{2033FF} \textbf{0.70}} & {\color[HTML]{FE0000} \textbf{0.74}} & 0.46 & 0.52 & 0.30 & 0.06 & 0.60 & 0.31 & 0.63 & 0.62 & 0.51 & 0.24 & 0.24\\ 

&{MSNBC} & 0.73 & 0.69 & 0.71 & 0.42 & {\color[HTML]{FE0000} \textbf{0.83}} & 0.51 & 0.00 & 0.78 & 0.62 & {\color[HTML]{2033FF} \textbf{0.82}} & 0.73 & 0.5 & 0.79 & 0.75\\ 

&{Microposts2014-Test} & 0.33 & 0.42 & 0.48 & 0.50 & {\color[HTML]{FE0000} \textbf{0.76}} & 0.41 & 0.05 & 0.64 & 0.52 & {\color[HTML]{2033FF} \textbf{0.73}} & 0.60 & 0.55 & 0.45 & 0.44\\ 

&{Microposts2014-Train} & 0.42 & 0.51 & 0.51 & 0.48 & {\color[HTML]{FE0000} \textbf{0.77}} & 0.00 & 0.31 & 0.65 & 0.52 & {\color[HTML]{2033FF} \textbf{0.71}} & 0.63 & 0.59 & 0.49 & 0.44\\ 

&{N3-RSS-500} & 0.66 & 0.45 & 0.44 & 0.20 & 0.48 & 0.00 & 0.00 & 0.44 & 0.38 & 0.53 & 0.44 & 0.45 & {\color[HTML]{FE0000} \textbf{0.69}} & {\color[HTML]{2033FF} \textbf{0.67}}\\ 

&{N3-Reuters-128} & 0.61 & 0.47 & 0.45 & 0.33 & {\color[HTML]{FE0000} \textbf{0.69}} & 0.00 & 0.41 & 0.51 & 0.41 & {\color[HTML]{2033FF} \textbf{0.65}} & 0.52 & 0.39 & {\color[HTML]{FE0000} \textbf{0.69}} & 0.64\\ 

&{OKE 2015 Task 1 evaluation} & 0.59 & 0.56 & 0.59 & 0.31 & 0.59 & 0.00 & 0.46 & {\color[HTML]{FE0000} \textbf{0.63}} & 0.61 & {\color[HTML]{FE0000} \textbf{0.63}} & 0.57 & {\color[HTML]{2033FF} \textbf{0.62}} & 0.58 & 0.55\\ 

&{OKE 2015 Task 1 example} & 0.50 & {\color[HTML]{2033FF} \textbf{0.60}} & 0.40 & 0.22 & 0.55 & 0.00 & {\color[HTML]{2033FF} \textbf{0.60}} & 0.55 & 0.00 & 0.50 & {\color[HTML]{2033FF} \textbf{0.60}} & 0.50 & {\color[HTML]{FE0000} \textbf{0.67}} & 0.50\\ 

&{OKE 2015 Task 1 training} & 0.62 & 0.67 & 0.71 & 0.25 & {\color[HTML]{FE0000} \textbf{0.78}} & 0.00 & 0.61 & {\color[HTML]{FE0000} \textbf{0.78}} & {\color[HTML]{2033FF} \textbf{0.77}} & 0.76 & 0.72 & 0.75 & 0.72 & 0.70\\ 

\midrule
\multirow{6}{*}{\rotatebox{90}{Multilingual}}
&{N$^3$ news.de} & 0.61 & 0.52 & 0.50 & 0.48 & 0.56 & 0.28 & 0.00 & 0.61 & 0.33 & 0.30 & 0.59 & 0.36 & {\color[HTML]{FE0000} \textbf{0.76}} & {\color[HTML]{2033FF} \textbf{0.63}}\\ 

&{Italian Abstracts} & 0.22 & 0.28 & {\color[HTML]{2033FF} \textbf{0.33}} & 0.00 & 0.00 & 0.00 & 0.00 & 0.00 & 0.00 & 0.20 & 0.00 & 0.00 & {\color[HTML]{FE0000} \textbf{0.80}} & {\color[HTML]{FE0000} \textbf{0.80}}\\ 

&{Spanish Abstracts} & 0.25 & 0.33 & 0.26 & 0.00 & 0.24 & 0.27 & 0.00 & 0.47 & 0.00 & 0.31 & 0.33 & 0.31 & {\color[HTML]{FE0000} \textbf{0.75}} &  {\color[HTML]{2033FF} \textbf{0.68}}\\ 

&{Japanese Abstracts} & 0.15 & 0.00 & 0.00 & 0.00 & 0.00 & 0.00 & 0.00 & 0.00 & 0.00 & {\color[HTML]{2033FF} \textbf{0.38}} & 0.00 & 0.00 & {\color[HTML]{FE0000} \textbf{0.54}} & {\color[HTML]{FE0000} \textbf{0.54}} \\ 

&{Dutch Abstracts} & 0.33 & 0.36 & 0.36 & 0.28 & 0.36 & 0.22 & 0.00 & 0.40 & 0.00 & 0.5 & 0.40 & 0.25 & {\color[HTML]{2033FF} \textbf{0.66}} & {\color[HTML]{FE0000} \textbf{0.67}}\\

&{French Abstracts} & 0.00 & 0.00 & {\color[HTML]{2033FF} \textbf{0.28}} & 0.22 & 0.00 & 0.25 & 0.00 & 0.00 & 0.00 & 0.20 & {\color[HTML]{2033FF} \textbf{0.28}} & {\color[HTML]{2033FF} \textbf{0.28}} & {\color[HTML]{FE0000} \textbf{0.80}} & {\color[HTML]{FE0000} \textbf{0.80}}\\ 
\midrule

&\textbf{Average} & 0.45 & 0.48 & 0.50 & 0.36 & 0.55 & 0.24 & 0.11 & 0.53 & 0.31 & {\color[HTML]{2033FF} \textbf{0.59}} & 0.54 & 0.45 & {\color[HTML]{FE0000} \textbf{0.63}} & {\color[HTML]{2033FF} \textbf{0.59}}\\ \midrule

&\textbf{Standard Deviation} & 0.19 & 0.22 & 0.18 & 0.19 & 0.27 & 0.21 & 0.21 & 0.23 & 0.26 & 0.20 & 0.22 & 0.20 & 0.13 & 0.13 \\
\bottomrule
\end{tabular}
\end{table*}

\textit{\textbf{On Multilingual datasets.}} Here, we show the easy portability and high quality of MAG for many different languages. Next to German, Italian, Spanish, French and Dutch, we chose Japanese to show the promising potential of MAG across different language systems. MAG's preprocessing \ac{NLP} techniques are multilingual, thus there is no additional implementation for handling the mentions with different characters. We used the same set of parameters as in the English evaluation but excluded the acronyms as they were only collected for English. Moreover, we performed the Page Rank algorithm over each \ac{KB} in each respective language to collect popularity values of their entities. 
The results displayed in the second part of ~\autoref{engfmeasure} show that MAG, using HITS, outperform all publicly available state-of-the-art approaches. Additionally, for Dutch, Page Rank outperforms HITS score. The improved performance of MAG is due to its knowledge-base agnostic algorithms and indexing models. For instance, although the mention ``Obama" has a high popularity in English, it may have less popularity in Italian or Spanish \ac{KB}s. Studies about the generation of proper names support this observation~\citep{dale1995computational,ferreira2017generating}.

\textit{\textbf{Fine-Grained Evaluation.}} In this analysis, we measured the quality of a given \ac{EL} for linking different types of entities. This extension also considers that a corpus tends to focus strongly on prominent or popular entities, which may cause evaluation problems. Hence, the extension evaluates the capability of a given \ac{EL} system to find entities with different levels of popularity, thus revealing its degree of bias towards popular entities. The fine-grained analysis shows that MAG is better at linking persons than other types of entities, and it can be explained by the indexes created by MAG in the offline phase. They collect last names and rare surfaces for entities. In addition, the results show that MAG is not biased towards linking only popular entities, as can be seen in \autoref{tab:finegrained}.

\begin{table*}[htb!]
\setlength\tabcolsep{2pt}
\footnotesize
\centering
\caption{Fine-grained micro F1 evaluation.}
\label{tab:finegrained}
\begin{tabular}{@{} lcccccc @{}}
\toprule
\textbf{Filter }& \textbf{IITB} & \textbf{N3-RSS-500} & \textbf{MSNBC} & \textbf{Spotlight} & \textbf{N3-Reuters-128} & \textbf{OKE 2015} \\
\toprule
Persons & 0.95 & 0.83 & 0.94 & 0.84 & 0.80 & 0.92 \\
Page Rank 10\% & 0.73 & 0.67 & 0.83 & 0.74 & 0.79 & 0.76 \\
Page Rank 10\%-55\% & 0.72 & 0.72 & 0.70 & 0.69 & 0.73 & 0.79 \\
Page Rank 55\%-100\% & 0.73 & 0.71 & 0.73 & 0.75 & 0.76 & 0.82 \\
Hitsscore 10\% & 0.77 & 0.74 & 0.77 & 0.69 & 0.73 & 0.76 \\
Hitsscore 10\%-55\% & 0.69 & 0.66 & 0.64 & 0.69 & 0.79 & 0.78 \\
Hitsscore 55\%-100\% & 0.71 & 0.66 & 0.84 & 0.74 & 0.77 & 0.80
\\
 \bottomrule
\end{tabular}
\end{table*}

\subsection{Demonstration}
\label{sec:demo}

This demonstration extends MAG to support EL in 40 different languages, including especially low-resources languages such as Ukrainian, Greek, Hungarian, Croatian, Portuguese, Japanese and Korean. Our demo relies on online web services which allow for an easy access to our entity linking approaches and can disambiguate against DBpedia and Wikidata. Additionally, MAG supports POST requests as well as it has a user-friendly web interface.

\begin{figure}[htb]
\centering
\includegraphics[scale=0.30]{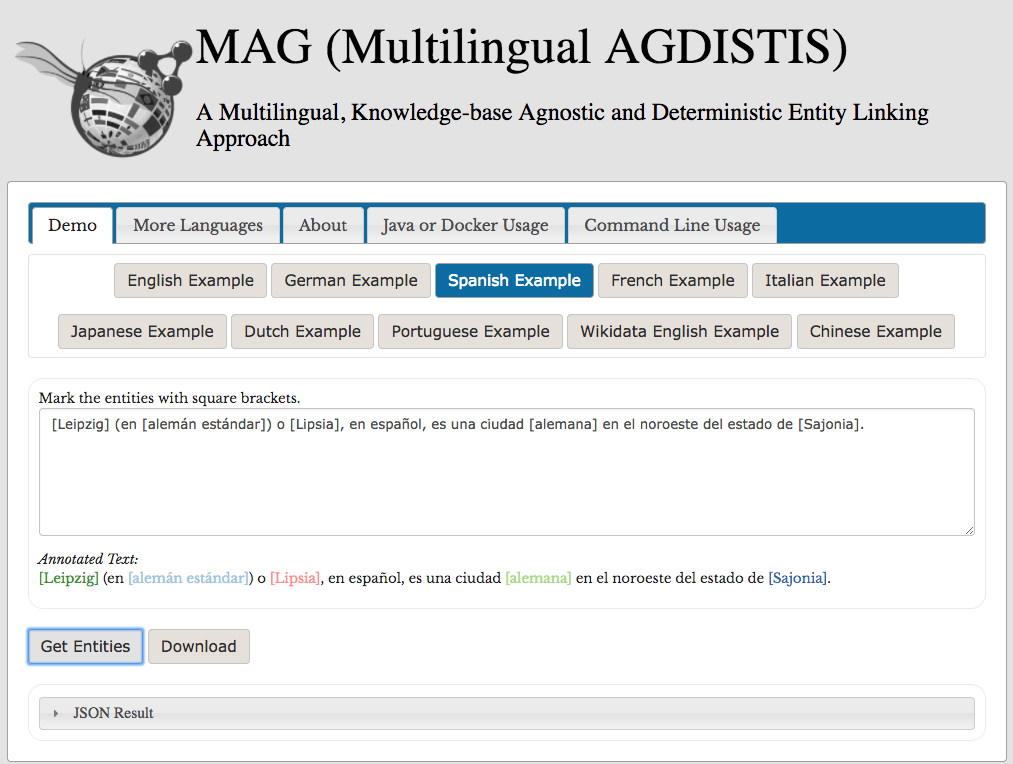}
\caption{A screenshot of MAG's web-based demo working on Spanish. }
\label{fig:spanish}
\end{figure}



\subsection{Reproducibility}

MAG was implemented within the AGDISTIS framework.\footnote{\url{https://github.com/dice-group/AGDISTIS}}
In addition, all experimental data, code and, results are publicly available.\footnote{\url{https://hobbitdata.informatik.uni-leipzig.de/agdistis/}}

\subsection{Summary}

The main contributions of this paper can be summarized as follows:
\begin{itemize}
    \item We present a novel multilingual and deterministic approach for \ac{EL} that combines lightweight and easily extensible graph-based algorithms with a new context-based retrieval method.
    \item MAG features an innovative candidate generation method that relies on various filter methods and search types for a better candidate selection.
    \item We provide a thorough evaluation of our overall system on 23 datasets using the GERBIL platform~\citep{gerbil}. Our results outperform all state-of-the-art approaches on 6 non-English datasets while achieving state-of-the-art performance on English.
\end{itemize}



\section{RDF2PT: Generating Brazilian Portuguese Texts from RDF \-Da\-ta}
\label{sec:rdf2pt}
{\small\textit{FOR MITIGATING THE LACK OF MULTILINGUAL APPROACHES IN TEXT GENERATION} \ref{challengeNLG}}, we developed RDF2PT, an approach that verbalizes \ac{RDF} data to Brazilian Portuguese language. 
A generic \ac{NLG} pipeline is composed by three tasks which are \emph{Document Planing}, \emph{Micro Planning} and \emph{Realization}. RDF2PT operates mostly at the level of the first two and to the \emph{Realization} task, RDF2PT uses an adaption of SimpleNLG to Brazilian Portuguese~\citep{de2014adapting}.

\subsection{Document Planning}
\label{subsection:document_planning}

This initial phase is divided into two sub-tasks. First, \emph{Content determination}, which decides what information a certain \ac{NLG} system should include in the generated text. Second, \emph{Discourse planning}, which determines the order of the information in paragraphs and its rhetorical relation.

\textit{\textbf{Content determination.}} RDF2PT assumes the description of a resource to be the set of \ac{RDF} statements of which this resource is the subject. Hence, given a resource, RDF2PT first performs a SPARQL query to get its most specific class through the predicate \texttt{rdf:type}. Afterward, RDF2PT gets all resources that belong to this specific class and ranks their predicates by using Page Rank~\citep{page1999pagerank} over the \ac{KB}. 
Once the predicates are ranked, RDF2PT considers only the top seven most popular predicates of the class to describe the input resource.
 

\textit{\textbf{Discourse planning.}} In this step, RDF2PT clusters and orders the triples. The subjects are ordered with respect to the number of their occurrences, thus assigning them to those input triples that mention them. RDF2PT processes the input in descending order with respect to the frequency of the variables they contain, starting with the projection variables and only after that, turning to other variables. 




\subsection{Micro Planning}
\label{subsection:content determination}

This step is concerned with the planning of a sentence. It comprises three sub-tasks. Firstly, \emph{Sentence aggregation} decides whether information will be presented individually or separately. Second, \emph{Lexicalization} chooses the right words and phrases in natural language for expressing the semantics about the data. Third, \emph{Referring Expression}) is the task responsible for generating syntagms (references) to discourse entities. 

\textit{\textbf{Sentence aggregation.}} This task is divided into two phases, \emph{subject grouping} and \emph{object grouping}. \emph{Subject grouping} collapses the predicates and objects of two triples if their subjects are the same. \emph{Object grouping} collapses the subjects of two triples if the predicates and objects of the triples are the same. The common elements are usually subject noun phrases and verb phrases (verbs together with object noun phrases). To maximize the grouping effects, we additionally collapse common prefixes and suffixes of triples, irrespective of whether they are full subject noun phrases or complete verb phrases.



\textit{\textbf{Lexicalization.}} This step comprises the main contribution of RDF2PT for verbalizing the triples in Brazilian Portuguese. In contrast to English, Brazilian Portuguese is a morphologically rich language which contains the grammatical gender of words. Grammatical gender plays a key role because it affects the generation of determiners and pronouns. It also influences the inflection of nouns and verbs. For instance, the passive expression of the verb \texttt{nascer} (en: ``be born") is \texttt{nascida} if the subject is feminine or \texttt{nascido} if masculine. Thus, the gender of words is essential for comprehending the semantics of a given Portuguese text. Also, Brazilian Portuguese has different possibilities in the expression of subject possessives. Hence, RDF2PT has to deal with the following phenomena while lexicalizing: 

\begin{itemize}

\item \textbf{Grammatical gender} - In Portuguese, the gender varies between masculine and feminine. This variation leads to supplementary challenges when lexicalizing words automatically. For example, a gender may be represented by articles ``um" and ``o" (masculine) or ``uma" and ``a" (feminine). However, the gender also affects the inflection of words. For instance, for the word ``cantor" (en: ``singer"), if the subject is feminine, the word becomes ``cantora". However, there are words that do not inflect, e.g., the word ``gerente" (en: ``manager"). If the subject is a woman, we only refer to it by using the article ``a'', i.e., ``a gerente". Therefore, there are some challenges to tackle for recognizing the gender and assigning it correctly. A tricky example to solve automatically is ``O Rio de Janeiro \'{e} uma cidade" (en: Rio de Janeiro is a city). In this case, the subject is masculine but its complement is feminine. Developing handcrafted rules to handle these phenomena can become a hard task. To deal with this challenge, we use a Part-Of-Speech tagger (TreeTagger in our case) as it retrieves the gender along with the parts of speech. All the obtained genders are attached along with the lexicalizations for supporting the realization step. 

\item  \textbf{Classes and resources} - The lexicalization of classes and resources is gathered by using a SPARQL query to get their Portuguese labels through the \texttt{rdfs:label} predicate\footnote{Note that it could be any property which returns a natural language representation of the given URI, see \citep{ell2011}.}. In case such a label does not exist, we use either the fragment of their URI (the string after the \verb|#| character) if it exists, or the string after the last occurrence of ``\verb|/|". Finally, this natural language representation is lexicalized as a noun phrase. Afterwards, RDF2PT recognizes the gender. In case the resource is recognized as a person, RDF2PT applies a string similarity measure (0.8 threshold) between the lexicalized word and a list of names provided by LD2NL. This list is divided by masculine and feminine which in turn results in the gender. On the other hand, if the resource is not a person, we use Tree-tagger.

\item \textbf{Properties} - The lexicalization of properties relies on one of the results of\-\citet{ngonga2013sorry}, i.e., that most property labels are either nouns or verbs. To determine which lexicalization to use automatically, we rely on the insight that the first and last words of a property label in Portuguese are commonly the key for determining the type of property. We then use the Tree-Tagger to get the part of speech of predicates. Properties whose label begins with a verb are lexicalized as verbs. For example, the predicate \texttt{dbo:knownFor}, which Portuguese label is ``conhecido por", has the first word identified as an inflection of the verb ``conhecer" (en:know). Therefore, 
we devised a set of rules to capture this behavior.

\item \textbf{Literals} - In an RDF graph, literals usually consist of a \emph{lexical form} \texttt{LF} and a \emph{datatype IRI} \texttt{DT}. If the datatype is \rdfLangString, a non-empty \emph{language tag} is specified and the literal is denoted as a \emph{language-tagged string}.\footnote{In RDF 1.0 literals have been divided into ``plain' literals with no type and optional language tags, and typed literals.}
Accordingly, the lexicalization of strings with language tags is carried out by using simply the lexical form, while omitting the language tag. For example, 
\texttt{``Albert Einstein"@pt} is lexicalized as ``Albert Einstein" or 
\texttt{"Alemanha"@pt} (``Germany"@en) is lexicalized as ``Alemanha". 

\end{itemize}

\textit{\textbf{REG.}} In this step, RDF2PT relies on the number of subjects contained by the \ac{RDF} statements and only uses other expressions to refer to a given subject in case there is more than one mention of it. RDF2PT replaces the subject by possessive or personal pronouns with the corresponding gender depending on the predicates. For instance, given a triple \texttt{dbr: Albert\_Einstein dbo:birthPlace dbr:Ulm}, the predicate is a noun phrase then the subject is replaced by a possessive form which is ``seu" (en:``his"). However, Brazilian Portuguese has two different ways to express possession and this variation exists due to the necessity of handling complex syntaxes in some sentences and also because the gender of pronouns agrees with objects instead of subjects. For example, ``A professora proibiu que o aluno utilizasse \texttt{seu} dicion\'ario." (eng: ``The teacher forbade the student to use \texttt{his/her} dictionary''). The possessive pronoun \texttt{seu} in this sentence does not indicate explicitly to whom the dictionary belongs, if it belongs to the \texttt{professora} (eng:teacher) or \texttt{aluno} (eng:student). Thus, we have explicitly to define the possessive pronoun in order to decrease the ambiguity in texts and it is obviously important when generating text from data. If this sentence was translated into English, we would have indicated to whom the dictionary belonged, \texttt{her} or \texttt{his}. To this end, we handle the ambiguity of possessive pronouns by interspersing the alternative forms, e.g., \texttt{dele} (eng:his) or \texttt{dela} (eng: her)" that agrees with the subject. However, it is used just in case more than one subject exists in the same description.


	
\subsection{Linguistic realisation}
\label{subsec:realisation}

This last step is responsible for mapping the obtained descriptions of sentences from the aforementioned tasks and verbalizing them syntactically, morphologically and orthographically into a correct natural language text. To this end, we perform this step by relying on a Brazilian adaptation of SimpleNLG~\citep{de2014adapting} and \cite{ngonga2013sorry}. 

\subsection{Evaluation} 

We based our evaluation methodology on~\citet{gardent2017creating} and \citet{ferreira2016towards}. Our main goal was to evaluate how well RDF2PT represents the information obtained from the data. We hence divided our evaluation set into expert and non-expert users. Both sets were made up of native speakers of Brazilian Portuguese. We selected six DBpedia categories like~\citep{gardent2017creating} for selecting the topic of texts. The categories were Astronaut, Scientist, Building, WrittenWork, City, and University. 

\textbf{Experts} -  We aimed to evaluate the adequacy and fluency of the generated texts from 10 experts. All experts hold at least a master degree in the fields \ac{NLP} or \ac{SW}. In the questionnaire, we used the same two questions as \citep{gardent2017creating}: (1) Adequacy: Does the text contain only and all the information from the data? (2) Fluency: Does the text sound fluent and natural?

\textbf{Non-experts} - We evaluated the clarity and fluency of the generated texts. To this end, we created three types of texts, (1) baseline, (2) RDF2PT and (3) Human. 
The experiment was performed by 30 participants (10 per list). They were asked to rate each text considering the clarity and fluency based on two questions from~\citet{ferreira2016towards} on a scale from 1 (Very Bad) to 5 (Very Good). The questions were: (1) Fluency: Does the text present a consistent, logical flow? (2) Clarity: Is the text easy to understand?

In total, we created three versions of 18 texts (one text per resource) selected randomly from the aforementioned DBpedia categories (total: 54 texts). These texts were distributed over three lists, such that each list contained one variant of each text, and there was an equal number of texts from the three types (Baseline, RDF2PT, Human).




\begin{table*}[htb]
\setlength\tabcolsep{2pt}
\footnotesize
\centering
\begin{tabular}{@{} ll @{}}
\toprule
\textbf{Version} & \textbf{Text} \\
\toprule
Baseline & Albert Einstein \'e cientista, Albert Einstein campo \'e f\'isica, Albert Einstein lugar \\ 
& falecimento Princeton. Albert Einstein ex-institui\c{c}\~ao \'e Universidade Zurique, \\
& Albert Einstein \'e conhecido Equivalência massa-energia, \\
& Albert Einstein pr\^{e}mio \'e Medalha Max Planck, Albert Einstein estudante \\
& doutorado \'e Ernst Gabor Straus.\\
\midrule
RDF2PT & Albert Einstein foi um cientista, o campo dele foi a f\'isica e ele  \\
& no Princeton. Al\'em disso, sua ex-institui\c{c}\~ao foi a Universidade de Zurique, \\ 
& ele \'e conhecido pela Equival\^encia massa-energia, o pr\^emio dele \\ 
& foi a Medalha Max Planck e o estudante de doutorado dele foi o Ernst Gabor Straus.\\
\midrule
Humano & Albert Einstein era um cientista, que trabalhava na \'area de F\'isica. Era conhecido \\ 
& pela f\'ormula de equival\^encia entre massa e energia. Formou-se na Universidade de Zurique. \\ 
& Einstein ganhou a medalha Max Planck por seu trabalho. Em Princeton, onde morreu, \\ 
& teve sob sua orienta\c{c}\~ao Ernst Gabor Straus. \\
 \bottomrule
\end{tabular}
\caption{Example of text in the Baseline, RDF2PT approach and Human version.}
\label{tab:exampleText}
\end{table*}

\subsection{Results}
\label{sec:results}

\begin{figure}[htb]
\centering
\includegraphics[width=0.6\textwidth]{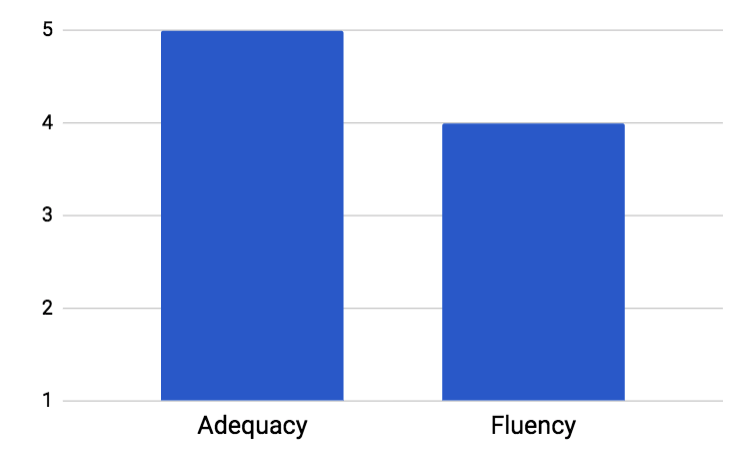}
\caption{RDF2PT results in experts survey}
\label{fig:experts}
\end{figure}

\textbf{Experts}~\autoref{fig:experts} displays the average fluency and clarity of the texts. The results suggest that RDF2PT is able to capture and represent the information from data adequately. Also, the generated texts are fluent enough to be understood by humans.

\begin{figure}[htb]
\centering
\includegraphics[width=0.6\textwidth]{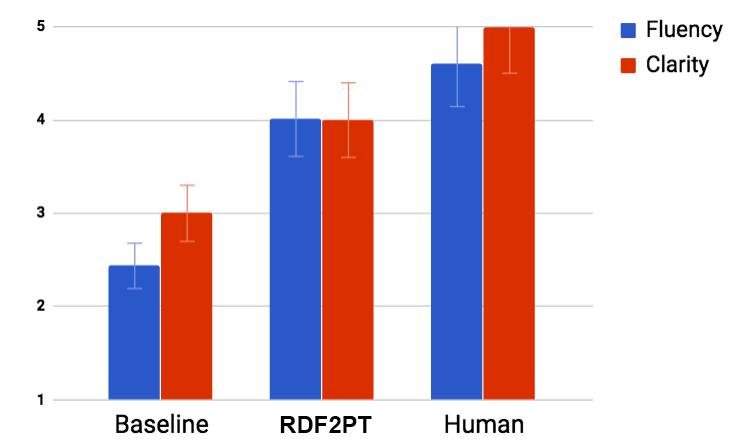}
\caption{Results in non-experts experiment}
\label{fig:non-experts}
\end{figure}

\textbf{Non-experts}~\autoref{fig:non-experts} depicts the average fluency and clarity of the texts where their topics are described by \emph{Baseline}, \emph{RDF2PT} and \emph{Human} approaches respectively. This figure clearly shows that \emph{Baseline} texts are rated lower than both the \emph{RDF2PT} and \emph{Human} texts, in fact, \emph{RDF2PT} is superior to \emph{Baseline} and close to \emph{Human}.

We performed a statistical analysis in order to measure the significance of the difference between the types (Baseline, RDF2PT, Human). First, we carried out a Friedman test~\citep{friedman1937use} which resulted in a significant difference in the fluency ($\textit{x}^2$ = 193.61, $\rho$  \textless 0.0001) and clarity ($\textit{x}^2$ = 180.9, $\rho$  \textless 0.0001) for the three kinds of texts. Afterward, we conducted a post-hoc analysis with Wilcoxon signed-rank test corrected for multiple comparisons using the Bonferroni method, resulting in a significance level set at $\rho$  \textless 0.017. Texts of the Baseline are hence significantly less statistically understandable (Z=525 and $\rho$  \textless 0.017.) and fluent (Z=275.5 and $\rho$  \textless 0.017.) than those generated by the RDF2PT approach. However, RDF2PT also generates texts less comprehensible (Z=1617.5 and $\rho$  \textless 0.017.) and fluent (Z=1640.0 and $\rho$  \textless 0.017.) than those generated by humans. Clearly, humans were superior to Baseline in terms of comprehensibility (Z=234.5 and $\rho$  \textless 0.017.) and fluency (Z=264.0.0 and $\rho$  \textless 0.017.), as we expected. 
Therefore, there is a significant difference among all models, being baseline~\textless~model~\textless~human.      

\subsection{Reproducibility}

All experimental data, code, and results are publicly available\footnote{\url{https://github.com/dice-group/RDF2PT}} as well as the extensions to more than one language. \footnote{\url{https://github.com/diegomoussallem/RDF2NL}} In addition, the experiment was run on CrowdFlower and is publicly available.\footnote{\url{https://ilk.uvt.nl/~tcastrof/semPT/evaluation/}}

\subsection{Summary}

The main contributions of this paper can be summarized as follows:
\begin{itemize}
    \item We present the first RDF-to-Text approach to Brazilian Portuguese.
    \item RDF2PT is extensible for Spanish and English as well as other languages such as Italian, French and German. 
    \item RDF2PT generates natural language sentences close to the human quality.
\end{itemize}

\section{NeuralREG: An End-to-End Approach to Referring Expression Generation}
\label{sec:neuralREG}
{\small\textit{FOR IMPROVING THE REFERRING FORM OF ENTITIES IN TEXT GENERATION} \ref{challengeNLG}}, we created the first approach, named NeuralREG, which relies on deep neural networks for making decisions about form and content in one go without explicit feature extraction from R\ac{RDF}-\ac{KG}.

NeuralREG accepts as input entities which are delexicalized to general tags (e.g., ENTITY-1, ENTITY-2) to decrease data sparsity. Based on the delexicalized input, the model generates outputs which may be likened to templates in which references to the discourse entities are not realized (as in ``The ground of ENTITY-1 is located in ENTITY-2.''). To this end, NeuralREG uses as training data a specific constructed set of 78,901 referring expressions to 1,501 entities in the context of the \ac{RDF}-\ac{KG}, derived from a (delexicalized) version of the WebNLG corpus \citep{claire2017,claire2017b}.

NeuralREG aims to generate a referring expression $y = \lbrace y_{1}, y_{2}, ... , y_{T} \rbrace$ with $T$ tokens to refer to a target entity token $x^{(wiki)}$ given a discourse pre-context $X^{(pre)} = \lbrace x^{(pre)}_{1}, x^{(pre)}_{2}, ..., x^{(pre)}_{m} \rbrace$ and pos-context $X^{(pos)} = \lbrace x^{(pos)}_{1}, x^{(pos)}_{2}, ..., x^{(pos)}_{l} \rbrace$ with $m$ and $l$ tokens, respectively. The model is implemented as a multi-encoder, attention-decoder network with bidirectional \citep{schuster1997bidirectional} \ac{LSTM} \citep{hochreiter1997long} sharing the same input word-embedding matrix $V$, we detail it in the next sections.

\subsection{Context encoders}

Our model starts by encoding the pre- and pos-contexts with two separate bidirectional \ac{LSTM} encoders \citep{schuster1997bidirectional,hochreiter1997long}. These modules learn feature representations of the text surrounding the target entity $x^{(wiki)}$, which are used for the referring expression generation. The pre-context $X^{(pre)} = \lbrace x^{(pre)}_{1}, x^{(pre)}_{2}, ..., x^{(pre)}_{m} \rbrace$ is represented by forward and backward  hidden-state vectors $(\overrightarrow{h}^{(pre)}_1, \cdots, \overrightarrow{h}^{(pre)}_m)$ and $(\overleftarrow{h}^{(pre)}_1, \cdots, \overleftarrow{h}^{(pre)}_m)$. The final annotation vector for each encoding timestep $t$ is obtained by the concatenation of the forward and backward representations  $h^{(pre)}_t = [\overrightarrow{h}^{(pre)}_t, \overleftarrow{h}^{(pre)}_t]$. The same process is repeated for the pos-context resulting in representations $(\overrightarrow{h}^{(pos)}_1, \cdots, \overrightarrow{h}^{(pos)}_l)$ and $(\overleftarrow{h}^{(pos)}_1, \cdots, \overleftarrow{h}^{(pos)}_l)$ and annotation vectors $h^{(pos)}_t = [\overrightarrow{h}^{(pos)}_t, \overleftarrow{h}^{(pos)}_t]$. Finally, the encoding of target entity $x^{(wiki)}$ is simply its entry in the shared input word-embedding matrix $V_{wiki}$. 

\subsection{Decoder}

The referring expression generation module is an \ac{LSTM} decoder implemented in three different versions: \texttt{Seq2Seq}, \texttt{CAtt} and \texttt{HierAtt}.  All decoders at each timestep $i$ of the generation process take as input features their previous state $s_{i-1}$, the target entity-embedding $V_{wiki}$, the embedding of the previous word of the referring expression $V_{y_{i-1}}$ and finally the summary vector of the pre- and pos-contexts $c_i$. The difference between the decoder variations is the method to compute $c_i$.

\textbf{\texttt{Seq2Seq}} models the context vector $c_i$ at each timestep $i$ concatenating the pre- and pos-context annotation vectors averaged over time:

\begin{equation}
\footnotesize{
\hat{h}^{(pre)} = \frac{1}{N}\sum^N_i h^{(pre)}_i \\
}
\end{equation}
\begin{equation}
\footnotesize{
\hat{h}^{(pos)} = \frac{1}{N}\sum^N_i h^{(pos)}_i
}
\end{equation}
\begin{equation}
\footnotesize{
c_i = [\hat{h}^{(pre)}, \hat{h}^{(pos)}]
}
\end{equation}

\textbf{\texttt{CAtt}} is an \ac{LSTM} decoder augmented with an attention mechanism \citep{bahdanau2014neural} over the pre- and pos-context encodings, which is used to compute $c_i$ at each timestep. We compute energies $e^{(pre)}_{ij}$ and $e^{(pos)}_{ij}$ between encoder states $h^{(pre)}_i$ and $h^{(post)}_i$  and decoder state $s_{i-1}$. These scores are normalized through the application of the softmax function to obtain the final attention probability $\alpha^{(pre)}_{ij}$ and $\alpha^{(post)}_{ij}$. Equations \ref{eq:energy} and \ref{eq:align} summarize the process with $k$ ranging over the two encoders ($k \in [pre, pos]$), being the projection matrices $W^{(k)}_a$ and $U^{(k)}_a$ and attention vectors $v^{(k)}_a$ trained parameters.

\begin{equation}
\footnotesize{
e^{(k)}_{ij} = v^{(k)T}_a \text{tanh}(W^{(k)}_a  s_{i-1}  + U^{(k)}_{a}  h^{(k)}_j ) 
}
\label{eq:energy}
\end{equation}

\begin{equation}
\footnotesize{
\alpha^{(k)}_{ij} = \frac{\text{exp}(e^{(k)}_{ij})}{\sum_{n=1}^{N} \text{exp}(e^{(k)}_{in})}
}
\label{eq:align}
\end{equation}

In general, the attention probability $\alpha_{ij}^{(k)}$ determines the amount of contribution of the $j$th token of $k$-context in the generation of the $i$th token of the referring expression. In each decoding step $i$, a final summary-vector for each context $c^{(k)}_i$ is computed by summing the encoder states $h^{(k)}_j$ weighted by the attention probabilities $\alpha^{(k)}_{i}$:

\begin{equation}
\footnotesize{
c^{(k)}_i = \sum_{j=1}^{N} \alpha^{(k)}_{ij} h^{(k)}_j
}
\end{equation}

To combine $c^{(pre)}_i$ and $c^{(pos)}_i$ into a single representation, 
this model simply concatenates the pre- and pos-context summary vectors $c_i = [c^{(pre)}_i, c^{(pos)}_i]$. 

\textbf{\texttt{HierAtt}} implements a second attention mechanism inspired by \citet{libovicky2017attention} in order to generate attention weights for the pre- and pos-context summary-vectors $c^{(pre)}_i$ and $c^{(pos)}_i$ instead of concatenating them. Equations \ref{eq:hier1}, \ref{eq:hier2} and \ref{eq:hier3} depict the process, being the projection matrices $W^{(k)}_b$ and $U^{(k)}_b$ as well as attention vectors $v^{(k)}_b$ trained parameters ($k \in [pre, pos]$).

\begin{equation}
\footnotesize{
e^{(k)}_{i} = v^{(k)T}_b \text{tanh}(W^{(k)}_b  s_{i-1}  + U^{(k)}_b  c^{(k)}_i )
}
\label{eq:hier1}
\end{equation}

\begin{equation}
\footnotesize{
\beta^{(k)}_i = \frac{ \text{exp}(e^{(k)}_{i}) }{\sum_{n}^{} \text{exp}(e^{(n)}_i)}
\label{eq:hier2}
}
\end{equation}

\begin{equation}
\footnotesize{
c_i = \sum_{k} \beta^{(k)}_{i} U^{(k)}_b c^{(k)}_i
}
\label{eq:hier3}
\end{equation}

\textbf{\texttt{Decoding}} Given the summary-vector $c_i$, the embedding of the previous referring expression token $V_{y_{i-1}}$, the previous decoder state $s_{i-1}$ and the entity-embedding $V_{wiki}$, the decoders predict their next state which is used later to compute a probability distribution over the tokens in the output vocabulary for the next timestep as Equations \ref{eq:decoding} and \ref{eq:softmax} show.

\begin{equation}
\footnotesize{
s_i = \Phi_\text{dec}(s_{i-1}, [c_i, V_{y_{i-1}}, V_{wiki}])
}
\label{eq:decoding}
\end{equation}

\begin{align}
  \begin{split}
  p(y_{i}|y_{<i}, X^{(pre)}, & x^{(wiki)},  X^{(pos)}) = \\
  & \text{softmax}(W_c s_i + b)
  \end{split}
\label{eq:softmax}
\end{align}

In Equation \ref{eq:decoding}, $s_0$ and $c_0$ are zero-initialized vectors. In order to find the referring expression $y$ that maximizes the likelihood in Equation \ref{eq:softmax}, we apply a beam search with length normalization with $\alpha = 0.6$ \citep{wu2016}:

\begin{equation}
\footnotesize{
lp(y) = \frac{(5+|y|)^{\alpha}}{(5+1)^{\alpha}}
\label{eq:lengthnorm}
}
\end{equation}

The decoder is trained to minimize the negative log likelihood of the next token
in the target referring expression:
\begin{equation}
\footnotesize{
J(\theta) = - \sum_{i} \text{log p}(y_i|y_{<i}, X^{(pre)}, x^{(wiki)}, X^{(pos)})
}
\end{equation}

\subsection{Automatic Evaluation}

\textbf{Data} - We evaluated our models on the training, development and test referring expression sets of the delexicalized WebNLG.

\textbf{Metrics} - We compared the referring expressions produced by the evaluated models with the gold-standards ones using accuracy and String Edit Distance \citep{levenshtein1966}. Since pronouns are highlighted as the most likely referential form to be used when a referent is salient in the discourse, as argued in the introduction, we also computed pronoun accuracy, precision, recall and F1-score in order to evaluate the performance of the models for capturing discourse salience. Finally, we lexicalized the original templates with the referring expressions produced by the models and compared them with the original texts in the corpus using accuracy and BLEU score \citep{papineni2002} as a measure of fluency. Since our model does not handle referring expressions for constants (dates and numbers), we just copied their source version into the template.

Post-hoc McNemar's and Wilcoxon signed ranked tests adjusted by the Bonferroni method were used to test the statistical significance of the models in terms of accuracy and string edit distance, respectively. To test the statistical significance of the BLEU scores of the models, we used a bootstrap resampling together with an approximate randomization method \citep{clarketal2011}\footnote{\url{https://github.com/jhclark/multeval}}.

\textbf{Settings} - NeuralREG was implemented using Dynet \citep{neubig2017}. Source and target word embeddings were 300D each and trained jointly with the model, whereas hidden units were 512D for each direction, totaling 1024D in the bidirection layers. All non-recurrent matrices were initialized following the method of \citet{glorot2011}. Models were trained using stochastic gradient descent with Adadelta \citep{zeiler2012} and mini-batches of size 40. We ran each model for 60 epochs, applying early stopping for model selection based on accuracy of the development set with patience of 20 epochs. For each decoding version (\texttt{Seq2Seq}, \texttt{CAtt} and \texttt{HierAtt}), we searched for the best combination of drop-out probability of 0.2 or 0.3 in both the encoding and decoding layers, using beam search with a size of 1 or 5 with predictions up to 30 tokens or until 2 ending tokens were predicted (\textit{EOS}). The results described in the next section were obtained on the test set by the NeuralREG version with the highest accuracy on the development set over the epochs.

\subsection{Human Evaluation}

Complementary to the automatic evaluation, we performed an evaluation with human judges, comparing the quality judgments of the original texts to the versions generated by our various models.

\textbf{Material} - We quasi-randomly selected 24 instances from the delexicalized version of the WebNLG corpus related to the test part of the referring expression collection. For each of the selected instances,  we took into account its source triple set and its 6 target texts: one original (randomly chosen) and its versions with the referring expressions generated by each of the five models introduced in this study (two baselines, three neural models). Instances were chosen following two criteria: the number of triples in the source set (ranging from 2 to 7) and the differences between the target texts. 

For each size group, we randomly selected four instances (of varying degrees of variation between the generated texts) giving rise to 144 trials ($=$ 6 triple set sizes $*$ 4 instances $*$ 6 text versions), each consisting of a set of triples and a target text describing it with the lexicalized referring expressions highlighted in yellow.

\textbf{Method} - The experiment had a latin-square design, distributing the 144 trials over 6 different lists such that each participant rated 24 trials, one for each of the 24 corpus instances, making sure that participants saw equal numbers of triple set sizes and generated versions. Once introduced to a trial, the participants were asked to rate the fluency (``does the text flow in a natural, easy to read manner?''), grammaticality (``is the text grammatical (no spelling or grammatical errors)?'') and clarity (``does the text clearly express the data?") of each target text on a 7-Likert scale, focussing on the highlighted referring expressions. The experiment is available on the website of the author\footnote{\url{https://ilk.uvt.nl/~tcastrof/acl2018/evaluation/}}. 

\textbf{Participants} - We recruited 60 participants, 10 per list, via Mechanical Turk. Their average age was 36 years and 27 of them were females. The majority declared themselves native speakers of English (44), while 14 and 2 self-reported as fluent or having a basic proficiency, respectively.
 
\subsection{Results.}

\textit{\textbf{Automatic evaluation}} - Table \ref{table:results} summarizes the results for all models on all metrics on the test set and Table \ref{table:example} depicts a text example lexicalized by each model. The first thing to note in the results of the first table is that the baselines in the top two rows performed quite strong on this task, generating more than half of the referring expressions exactly as in the gold-standard. The method based on \citet{ferreira2016b} performed statistically better than {\it OnlyNames} on all metrics due to its capability, albeit to a limited extent, to predict pronominal references (which {\it OnlyNames}\/ obviously cannot). 

\begin{table*}
\setlength\tabcolsep{2pt}
\footnotesize{
\centering
	\begin{tabular}{l l l | l l l l | l l}
    \midrule
    & \multicolumn{2}{ c }{\textbf{All References}} & \multicolumn{4}{| c |}{\textbf{Pronouns}} & \multicolumn{2}{ c }{\textbf{Text}}  \\
	& Acc. & SED & Acc. & Prec. & Rec. & F-Score & Acc. & BLEU  \\
    \midrule
    \textit{OnlyNames}        & 0.53$^{D}$ & 4.05$^{D}$ & - & - & - & - & 0.15$^{D}$ & 69.03$^{D}$  \\
    \textit{Ferreira}          & 0.61$^{C}$ & 3.18$^{C}$ & 0.43$^{B}$ & 0.57 & 0.54 & 0.55 & 0.19$^{C}$ & 72.78$^{C}$  \\
    \midrule
    NeuralREG+\texttt{Seq2Seq} & 0.74$^{A,B}$ & 2.32$^{A,B}$ & 0.75$^{A}$ & 0.77 & 0.78 & 0.78 & 0.28$^{B}$ & 79.27$^{A,B}$  \\
    NeuralREG+\texttt{CAtt}    & 0.74$^{A}$ & 2.25$^{A}$ & 0.75$^{A}$ & 0.73 & 0.78 & 0.75 & 0.30$^{A}$  & 79.39$^{A}$  \\
    NeuralREG+\texttt{HierAtt} & 0.73$^{B}$ & 2.36$^{B}$ & 0.73$^{A}$ & 0.74 & 0.77 & 0.75 & 0.28$^{A,B}$ & 79.01$^{B}$  \\
    \bottomrule
	\end{tabular}
\caption{(1) Accuracy (Acc.) and String Edit Distance (SED) results in the prediction of all referring expressions; (2) Accuracy (Acc.), Precision (Prec.), Recall (Rec.) and F-Score results in the prediction of pronominal forms; and (3) Accuracy (Acc.) and BLEU score results of the texts with the generated referring expressions. Rankings were determined by statistical significance.}
\label{table:results}
}
\end{table*}

\begin{table*}
\setlength\tabcolsep{2pt}
\fontsize{9.5pt}{9pt}\selectfont
\begin{center}
\begin{tabular}{@{} ll @{}}
\hline
\textbf{Model} & \textbf{Text} \\
\hline
\textit{OnlyNames}      & \textbf{alan shepard} was born in \textbf{new hampshire} on \textbf{1923-11-18} . \\ 
& before \textbf{alan shepard} death in \textbf{california} \textbf{alan shepard} had been awarded \\
& \textbf{distinguished service medal (united states navy)} an award higher than \\ 
& \textbf{department of commerce gold medal} . \\
\hline
\textit{Ferreira}   & \textbf{alan shepard} was born in \textbf{new hampshire} on \textbf{1923-11-18} . \\
& before \textbf{alan shepard} death in \textbf{california} \textbf{him} had been awarded \\
& \textbf{distinguished service medal} an award higher than \textbf{department of commerce gold medal} .
 \\
\hline
\texttt{Seq2Seq}   & \textbf{alan shepard} was born in \textbf{new hampshire} on \textbf{1923-11-18} . \\
& before \textbf{his} death in \textbf{california} \textbf{him} had been awarded \\
& \textbf{the distinguished service medal by the united states navy} an award higher than \\
& \textbf{the department of commerce gold medal} . \\
\hline
\texttt{CAtt}   & \textbf{alan shepard} was born in \textbf{new hampshire} on \textbf{1923-11-18} . before \textbf{his} death in \textbf{california} \\
& \textbf{he} had been awarded \textbf{the distinguished service medal by the us navy} an \\
& award higher than \textbf{the department of commerce gold medal} . \\
\hline
\texttt{HierAtt}   & \textbf{alan shephard} was born in \textbf{new hampshire} on \textbf{1923-11-18} . \\
& before \textbf{his} death in \textbf{california} \textbf{he} had been awarded \\
& \textbf{the distinguished service medal} an award higher than \\
& \textbf{the department of commerce gold medal} . \\
\hline
Original    & \textbf{alan shepard} was born in \textbf{new hampshire} on \textbf{18 november 1923} . \\
& before \textbf{his} death in \textbf{california} \textbf{he} had been awarded \\
& \textbf{the distinguished service medal by the us navy} \\
& an award higher than \textbf{the department of commerce gold medal} . \\
\hline
\end{tabular}
\caption{Example of text with references lexicalized by each model.}
\label{table:example}
\end{center}
\end{table*}

We reported results on the test set for NeuralREG+\texttt{Seq2Seq} and NeuralREG+\texttt{CAtt} using dropout probability 0.3 and beam size 5, and  NeuralREG+\texttt{HierAtt} with dropout probability of 0.3 and beam size of 1 selected based on the highest accuracy on the development set. Importantly, the three NeuralREG variant models statistically outperformed the two baseline systems. They achieved BLEU scores, text and referential accuracies as well as string edit distances in the range of 79.01-79.39, 28\%-30\%, 73\%-74\% and 2.25-2.36, respectively. This means that NeuralREG predicted 3 out of 4 references completely correct, whereas the incorrect ones needed an average of 2 post-edition operations in character level to be equal to the gold-standard. When considering the texts lexicalized with the referring expressions produced by NeuralREG, at least 28\% of them are similar to the original texts. Especially noteworthy was the score on pronoun accuracy, indicating that the model was well capable of predicting when to generate a pronominal reference in our dataset.

The results for the different decoding methods for NeuralREG were similar, with the NeuralREG+\texttt{CAtt} performing slightly better in terms of the BLEU score, text accuracy and String Edit Distance. The more complex NeuralREG+\texttt{HierAtt} yielded the lowest results, even though the differences with the other two models were small and not even statistically significant in many of the cases.

\textit{\textbf{Human evaluation}} - Table \ref{table:human} summarizes the results and reveals a clear pattern: all three neural models scored higher than the baselines on all metrics, with especially NeuralREG+\texttt{CAtt} approaching the ratings for the original sentences, although differences between the neural models were small. Concerning the size of the triple sets, we did not find any clear pattern. To test the statistical significance of the pairwise comparisons, we used the Wilcoxon signed-rank test corrected for multiple comparisons using the Bonferroni method. In contrast to the automatic evaluation, the results of both baselines were not statistically significant for the three metrics. In comparison with the neural models, NeuralREG+\texttt{CAtt} significantly outperformed the baselines in terms of fluency, whereas the other comparisons between baselines and neural models were not statistically significant. The results for the three different decoding methods of NeuralREG did not reveal a significant difference as well. Finally, the original texts were rated significantly higher than both baselines in terms of the three metrics, also than NeuralREG+\texttt{Seq2Seq} and NeuralREG+\texttt{HierAtt} in  terms of fluency, and higher than NeuralREG+\texttt{Seq2Seq} in terms of clarity.

 \begin{table}
\footnotesize{
\centering
	\begin{tabular}{l l l l}
    \midrule
	& Fluency & Grammar & Clarity \\
    \midrule
    \textit{OnlyNames}        & 4.74$^{C}$ & 4.68$^{B}$ & 4.90$^{B}$ \\
    \textit{Ferreira}          & 4.74$^{C}$ & 4.58$^{B}$ & 4.93$^{B}$ \\
    \midrule
    NeuralREG+\texttt{Seq2Seq} & 4.95$^{B,C}$ & 4.82$^{A,B}$ & 4.97$^{B}$ \\
    NeuralREG+\texttt{CAtt}    & 5.23$^{A,B}$ & 4.95$^{A,B}$ & 5.26$^{A,B}$ \\
    NeuralREG+\texttt{HierAtt} & 5.07$^{B,C}$ & 4.90$^{A,B}$ & 5.13$^{A,B}$ \\
    \midrule
    \textit{Original}          & 5.41$^{A}$ & 5.17$^{A}$ & 5.42$^{A}$ \\
    \bottomrule
	\end{tabular}
\caption{Fluency, Grammaticality and Clarity results obtained in the human evaluation. Rankings were determined by statistical significance.}
\label{table:human}
}
\end{table}

\subsection{Reproducibility}

All experimental data, code, and models are publicly available.\footnote{\url{https://github.com/ThiagoCF05/NeuralREG}}.

\subsection{Summary}

The main contributions of this paper can be summarized as follows:
\begin{itemize}
    \item We present the first full \ac{REG} model based on \ac{NN} from \ac{RDF}-\ac{KG}. 
    \item NeuralREG achieves fluency close to the human quality while choosing the referential form of entities in text.
    \item NeuralREG achieves high accuracy in predicting the form of pronouns for entities. 
\end{itemize}

\section{KG-NMT: Utilizing Knowledge Graphs for Neural Machine \-Trans\-la\-tion Augmentation}
\label{sec:KG-NMT}
\begin{figure}[htb]
\centering
    \includegraphics[width=\textwidth]{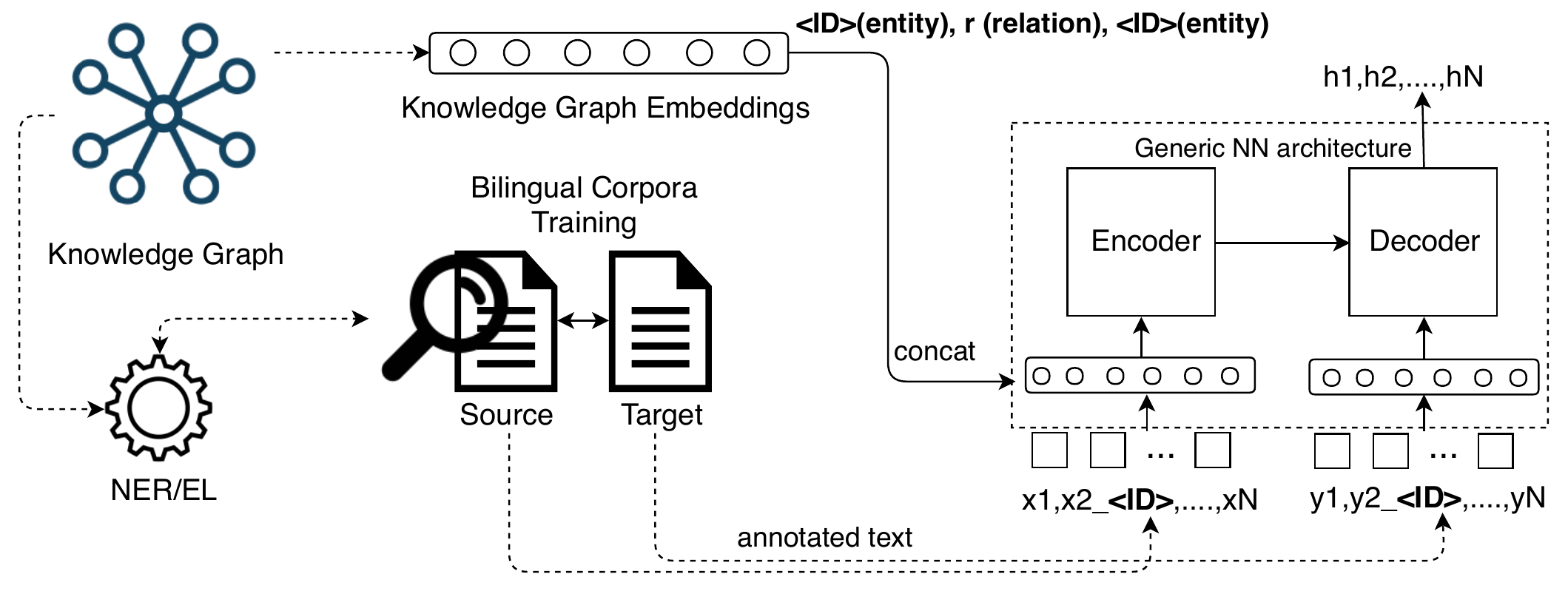}
    \caption{Overview of the KG-NMT methodology.}
    \label{fig:arch}
\end{figure}

{\small\textit{FOR ADDRESSING THE ENTITY TRANSLATION PROBLEM IN TEXT} \ref{challengeNMT}}, we designed KG-NMT, an \ac{NMT} model which is augmented with \ac{KG}. KG-NMT is based on the observation that more than 150 billion facts referring to more than 3 billion entities are available in the form of \ac{KG} on the \ac{LOD} Cloud.\footnote{\url{http://lod-cloud.net/}} Hence, the intuition behind our methodology is as follows: \emph{Given that \ac{KG}s describe real-world entities, we can use a \ac{KG} along with \ac{EL} to optimize values of entities' vector in the embedding space and consequently to achieve a better translation quality of entities in text}. Figure~\ref{fig:arch} depicts the general idea of our methodology. 

We devised two strategies to instantiate our methodology. In the first strategy, we link the \ac{NE}s in the source and target texts to a reference \ac{KB}, i.e., DBpedia, using MAG~\citep{moussallem2017mag}, a multilingual \ac{EL} system. We then incorporate the \ac{URI}s of entities along with the tokens akin to Li et al.~\citeyearpar{li2018named} with the \ac{NE}-tags. For example, the word \textit{cancer} can be annotated with \texttt{cancer|dbr\_Cancer},\footnote{\url{http://dbpedia.org/resource/Cancer}} and its translation represented in the German part of the DBpedia \ac{KB} (\texttt{dbr\_Krebs\_\-(Medizin)}). After incorporating the URIs, we embed the reference \ac{KB}, using the \textit{fastText} \ac{KGE} algorithm~\citep{joulin2017fast}. Once the \ac{KGE}s are created, we concatenate their vectors to the internal vectors of \ac{NMT} embeddings. The concatenation is possible as the annotations are present in the texts and consequently in the vocabulary. We chose to concatenate the vectors instead of leveraging their values because the concatenation preserves the values of \ac{KGE}s while leveraging loses its original values. In case a given entity in the text does not have a vector in the \ac{KGE}s, the concatenation inserts an empty vector, while the performance of \ac{NMT} remains unaffected. For example, suppose the entity ``USA'' appears in the parallel training data, i.e., a 500 dimensional vectors space is learned. Likewise, it also appears in the \ac{KG}, thus a vector space of the same dimension is learned. After concatenation, the \textit{USA} embedding becomes 2x500 dimensions, whereby entities, which are not present in the \ac{KGE}s are concatenated with an empty 500 dimensional vector. 

Although incorporating \ac{EL} as a feature into \ac{NMT} is interesting by itself, the annotation of entities in the training set and the post-editing task can be resource-intensive. Additionally, one limitation of structure-based \ac{KGE}s is that it can only work with word-based models since the application of any segmentation model, such as \ac{BPE}, on entities and relations may force the algorithm to assign wrong vectors to the entities. \ac{BPE} is a form of data compression that iteratively replaces the most frequent pair of bytes in a sequence with a single, unused byte. 
For example, the entities \texttt{dbr:Leipzig} and \texttt{dbr:Leibniz}\footnote{Subword segmentation (BPE) - Leipzig: Le$\blacksquare$ ip$\blacksquare$ zig ; Leibniz: Le$\blacksquare$ ib$\blacksquare$ n$\blacksquare$ iz} can be similar when considering sub-word units (characters), however, the first is a location while the second is a person. Both entities can be connected via the entity \texttt{dbr:University\_of\_Leipzig} but in this case, we are analyzing the sub-word units and not their graph connections. Thus, they should not be regarded as similar in the perspective of entities.

To overcome this limitation, we devise our second strategy, which uses only semantically-enriched \ac{KGE}s and skips the \ac{EL} part. Here, we enrich the structure-based \ac{KGE}s with surface forms of the entities found in the DBpedia \ac{KB}, thus decreasing the annotation effort and allowing the use of segmentation-models, i.e, sub-word information, on the surface forms differently from the first strategy, which considered only entities. Our hypothesis lies in the unsupervised learning capability of \ac{NN}s that can predict the annotation and alignment of the entities by itself. To generate the semantically-enriched \ac{KGE}s, we rely on multinomial logistic regression~\citep{bohning1992multinomial} as a classifier in a supervised training implemented in \textit{fastText} which assigns labels to the vector representations. The classification task creates inverse relations among the resources in order to map the relation of the subjects and objects in the triples and also it assigns a label, in this case, the entity's \ac{URI}, to the surface forms of entities and include them in the same vector space. The goal of the task is to predict the  \ac{URI}s of entities by their surface form. For example, we add to the triple, \texttt{<USA, type, Country>} the following information, \texttt{<USA, surfaceForm, United States of America>}. Thus, the training data looks \texttt{\_\_label\_\_dbr:USA United States America}.\footnote{More than one surface forms can be assigned to the entities.} The classifier creates an additional (hidden) relational vector between every entity (labels) and their associated words. Thus, the model when asked for the label of "United States of America" returns \texttt{dbr:USA}. However, we do not rely on the complete model rather we use the \ac{KGE} generated output with the surface forms attached to it. The semantically-enriched-\ac{KGE}s training jointly embeds entities and words into the same vector space, thus generating a vector for every word, which composes the entity's label. Therefore, while learning the \ac{KGE} along with the \ac{NMT} vocabulary, the \ac{NN} can retrieve from the lookup table the surface forms of entities and use their labels \ac{URI}s to align both source and target entities.
By enriching the \ac{KGE}s with surface forms, it allows using these vectors to initialize the embedding layer's weights of the \ac{NMT} models. This initialization is similar to the one used with pre-trained monolingual embeddings in \ac{NMT}~\citep{neishi2017bag}. But, instead of containing various words, the se\-man\-ti\-ca\-lly-en\-ri\-ched \ac{KGE}s has only the surface forms of the entities, which are also present in the \ac{NMT} vocabulary. The default initialization of the embeddings layer is a function that assigns random values to the weight matrix, whereas, in our second strategy, the values from \ac{KGE}s matrix are used to assign constant values to the matrix using a default concat function. Moreover, the employment of semantically-enriched \ac{KGE}s prevents some errors from the alignment of entities from source and target texts. For example, the entity linker runs in two distinct instances for each language, thus in case a certain entity is annotated only on the source or on the target language side, the \ac{NMT} approach is affected as the translation task requires aligned bilingual parallel texts for training.

\subsection{Evaluation}

Different \ac{NN} architectures are hard to compare as they are susceptible to hyper-pa\-ra\-me\-ters. Therefore, we follow the idea of using a minimal reasonable configuration set to the \ac{NMT} in order to fairly analyze the contributions of the used \ac{KG}. 

\textbf{\textit{\ac{NMT} Framework}} - For our overall experiments, we used a bi-directional \ac{RNN}-\ac{LSTM} 2-layer encoder-decoder model with attention mechanism~\citep{bahdanau2014neural}. The training uses a batch size of 32 and the stochastic gradient descent with an initial learning rate of 0.0002. We set a source and target word embeddings' size of 500, and hidden layers to size 500, dropout = 0.3 (naive). We used a maximum sentence length of 80, a vocabulary of 50,000 words for the word based models and a beam size of 5. All experiments were performed with OpenNMT~\citep{2017opennmt}. In addition, we used a copy mechanism for investigating the \ac{OOV} words issue. Moreover, we encoded words using \ac{BPE} with 32,000 merge operations to achieve an open vocabulary. Therefore, we created three kinds of baseline models (word-based, copyM, BPE32) using all the options mentioned above for evaluating the quality of the translation models. For training the \ac{NMT} models, we attempted to be as generic as possible. Our training set consists of a merge of the initial one-third of JRC-Acquis 3.0~\citep{steinberger2006jrc}, Europarl~\citep{koehn2005europarl}, and OpenSubtitles2013~\citep{TIEDEMANN12.463}, obtaining a parallel training corpus of two million sentences, containing around 38M running words. We performed our experiments on the English-German language pair as it is one of the most evaluated language pairs in the evaluation campaigns and translating into German is challenging in itself, due to its complex morphology and compounding.

\textbf{\textit{\ac{NMT} Augmentation}} - For augmenting the three baseline models with our two \ac{KG}-based strategies, we annotated the parallel bilingual corpora with MAG (first strategy), a multilingual \ac{EL} system~\citep{moussallem2017mag}, which is language and \ac{KB} agnostic. Afterwards, we trained the \ac{KGE}s, with a vector dimension size of 500 and a window size of 50 using hierarchical softmax. For semantically-enriched \ac{KGE}s, we added the surface forms whereby we used the same \ac{BPE} models on it. For the sake of comparison, we dubbed the KG-NMT approach that relies on \ac{EL} and structured-based \ac{KGE}s (first strategy) as \textit{KG-NMT (EL+KGE)} and the version with semantic information (second strategy) as \textit{KG-NMT (SemKGE)}. For overcoming both limitations of \textit{fastText}, which are having a clean KB and information of local graph connectivity, we relied on specific sub-sets of the English and Germany DBpedia \ac{KG} which contain transitive and \ac{CBD} resources along with their surface forms.\footnote{The files are mappingbased\_objects, labels, and interlinking\_languages} The English \ac{KB} contains 4.2 million entities, 661 relations, and 2.1 million surface forms, where the German version has 1 million entities, 249 relations, and 0.5 million surface forms.

\textbf{\textit{Evaluation Metrics}} - We used three automatic \ac{MT} standard metrics, {\sc BLEU}~\citep{papineni2002bleu}, { \sc METEOR}~\citep{banerjee2005meteor} and  {\sc chrF3}~\citep{popovic2017chrf++} to ensure a consistent and clear evaluation on the common evaluation datasets of the WMT evaluation shared tasks, named \textit{newstest}, between 2015 and 2018 as well as on the domain-specific datasets. Moreover, we carried out a manual analysis of outputs for assuring the contribution from \ac{KG}s, DBpedia, and we investigated the use of \ac{KG} in other settings as follows:  


{\textit{\textbf{Monolingual Embeddings vs. \ac{KGE}s}}} - Here, we aim to compare the performance of an \ac{NMT} using pre-trained monolingual embeddings with the semantically-enriched \ac{KGE}s as both can be used to initialize the internal vectors' values of an \ac{NMT} model. Our focus is to analyze if the \ac{KGE}s with fewer vectors can perform better than the monolingual embeddings for addressing the translation of entities and terminologies. Commonly, pre-trained monolingual word embeddings are used when the bilingual training data of a given language pair is scarce. These monolingual embeddings can be used to maximize the vector values of both, source or target languages as they are usually trained on a large monolingual corpus. Thus, we used a pre-trained monolingual embeddings model, which has 9.2 billion words for English and another with 1.3 billion words for German from~\citep{grave2018learning}. We dubbed as \textit{biRNN+MonoE}, the \ac{NMT} model  which is maximized with the pre-trained monolingual embeddings.

\textbf{\textit{Continuous Training on Domain-Specific Parallel Datasets}} - Our goal is to inspect the capability of improving the domain-specific translations using \ac{KG}s since they document domain-specific information. To this end, we relied on the continued training technique~\citep{luong2015stanford} to adapt a generic \ac{NMT} system to the financial, medical and \ac{IT} domains. For the financial domain, we used the \ac{IFRS} ontology and divided the documented labels into training (for continued training), development and test set, containing 1,000 labels each. Similarly, we used the \ac{ICD}-10 ontology, with 1,000 labels in the medical domain for the continued training, development and test set. Finally, for the \ac{IT} domain, we used the IT-WMT16\footnote{\url{http://www.statmt.org/wmt16/it-translation-task.html}} sets. The continued training set contains 50,121, the development 2,000 and test 1,000 sentences. We followed the same methodology to insert the DBpedia \ac{KG} in the domain adapted models. The adapted models are named \textit{KG-NMT(EL+KGE)\_adapt} and \textit{KG-NMT(SemKGE)\_adapt} respectively.

\begin{table*}[]
\caption{Results in BLEU, METEOR, chrF3 on WMT newstest datasets.}
\label{tbl:newstest-birnn}
\fontsize{9pt}{12pt}\selectfont
\setlength\tabcolsep{1.8pt}
\centering
\begin{tabular}{@{}llccc|ccc|ccc|ccc@{}}
\textbf{Models} &  & \multicolumn{3}{c}{\textbf{newstest2015}} & \multicolumn{3}{c}{\textbf{newstest2016}} & \multicolumn{3}{c}{\textbf{newstest2017}} & \multicolumn{3}{c}{\textbf{newstest2018}} \\
\toprule
&&  \rot{{BLEU}} & \rot{{METEOR}} & \rot{{chrF3}} & \rot{{BLEU}} & \rot{{METEOR}} & \rot{{chrF3}} & \rot{{BLEU}} & \rot{{METEOR}} & \rot{{chrF3}} & \rot{{BLEU}} & \rot{{METEOR}} & \rot{{chrF3}} \\
\hline
\multirowcell{3}{\textbf{Word-based}} & biRNN-lstm baseline & 16.77 & 35.20 & 41.11 & 18.55 & 36.62 & 42.54 & 15.10 & 33.75 & 39.52 & 20.53 & 39.02 & 43.92 \\
&KG-NMT(EL+KGE) & 19.86 & 38.25 & 42.92 & 22.38 & 40.40 & 45.18 & 18.04 & 36.94 & 41.55 & 24.87 & 43.49 & 46.88 \\
&KG-NMT(SemKGE) &\textbf{ 21.49} & \textbf{40.19} & \textbf{44.72} &  \textbf{24.01} & \textbf{42.47} & \textbf{46.84} & \textbf{19.66} & \textbf{38.89} & \textbf{43.11} & \textbf{27.02} & \textbf{45.77} & \textbf{48.70} \\
\hline
\multirowcell{3}{\textbf{CopyM}} & biRNN-lstm baseline & 19.63 & 39.20 & 46.38 & 21.37 & 40.90 & 47.85 & 17.88 & 37.89 & 44.85 & 24.22 & 43.96 & 50.15 \\
& KG-NMT(EL+KGE) & 22.46 & 41.67 & 48.28 & 25.05 & 44.23 & 50.66 & 20.77 & 40.58 & 47.04 & 28.44 & 47.86 & 53.25 \\
& KG-NMT(SemKGE) & \textbf{24.08} & \textbf{43.43} & \textbf{49.72} & \textbf{26.70} & \textbf{46.08} & \textbf{52.05} & \textbf{22.30} & \textbf{42.37} & \textbf{48.36} & \textbf{30.55} & \textbf{49.92} & \textbf{54.71} \\
\hline
\multirowcell{3}{\textbf{BPE}} & biRNN-lstm baseline & 15.89 & 36.51 & 45.97 & 21.95 & 42.88 & 52.68 & 16.80 & 39.12 & 49.35 & 23.85 & 45.85 & 54.98 \\
& KG-NMT(EL+KGE) & N/A & N/A & N/A & N/A & N/A & N/A & N/A & N/A & N/A & N/A & N/A & N/A \\
& KG-NMT(SemKGE)  & \textbf{21.74} & \textbf{41.41} & \textbf{50.04} & \textbf{24.86} & \textbf{44.32} & \textbf{53.59} & \textbf{20.45} & \textbf{40.62} & \textbf{49.45} & \textbf{28.02} & \textbf{47.51} & \textbf{55.16} \\
\bottomrule
\end{tabular}
\end{table*}

\subsection{Results}

In this section, we report the results of our experiments and perform a manual analysis of each experimental setting.

\textit{\textbf{Overall Results}} - Table~\ref{tbl:newstest-birnn} shows the results for \emph{KG-NMT} models in comparison to the baselines on the \emph{newstest} dataset between 2015 and 2018. Using \ac{KGE}s leads to a clear improvement over the baseline as it significantly improved the translation quality in terms of  {\sc BLEU} (+3), {\sc METEOR} (+4) and {\sc chrF3} (+3) metrics. \emph{KG-NMT (SemKGE)} outperformed \emph{KG-NMT (EL+KGE)} by around +1.3 in BLEU and chrF3, while we observed a +2 point improvement for METEOR. This difference between the contribution of \ac{KGE} types is directly related to the \ac{EL} performance, which did not manage to annotate all kind of entities present in the text. Consequently, the \ac{RNN} was not able to learn the translations of entities from the DBpedia \ac{KG}. A different \ac{EL} that is able to disambiguate more types of entities can improve the results of \emph{KG-NMT (EL+KGE)}, but still its training time is considerable longer than \emph{KG-NMT (SemKGE)}. The augmented model on \ac{BPE}, \emph{KG-NMT (SemKGE)}, presented consistent improvements showing that the model was capable of learning the segmentation applied on surface forms when translating to morphologically complex languages, such as German. We observed that the copy mechanism can be beneficial for named entities, which were not found in \ac{KG}. These entities were copied from the source language and added consequently as a translation to the target language. For example, \emph{Chad Johnston} was copied from the source into the target languages as a translation, since this name was not found in the \ac{KB}.

A detailed study of our results showed that the number of \ac{OOV} words decreased considerably with the augmentation through \ac{KGE}s. Table~\ref{tab:unk} shows the number of \ac{OOV} words generated by the models across all WMT \textit{newstest} datasets. The statistics cannot ensure that every \ac{OOV} word that became a known word was essentially an entity presented in DBpedia \ac{KG}. Thus, we chose the \textit{newstest2015} for a manual analysis.\footnote{
our full analysis of the entities' translation can be found in \url{https://git.io/KG-NMT-experiments-entities}} 
We observed that many \ac{OOV} words, that became known were in fact entities contained in the \ac{KG}. As an example (newstest2015 line 1265), the acronym  \textit{UK} was not translated by the \textit{biRNN-lstm} baseline even when the copy mechanism (\textit{UK}) or \ac{BPE} (\textit{Britische}) was used. However, it was correctly translated into German as \textit{Gro{\ss}britannien} by both \ac{KGE}s augmented models. Similarly, the entity \textit{Coastguard} (line 1540) was not translated correctly by the baseline models, whereby both \ac{KGE}s models were able to translate it correctly into \textit{K\"{u}stenwache}. Moreover, \textit{KG-NMT (EL+KGE)} was able to translate the word \textit{teacher} (line 438) correctly into \textit{Lehrer} using the knowledge acquired from \ac{KG}.\footnote{http://dbpedia.org/resource/Teacher} This human evaluation confirms our hypothesis by showing that the \ac{KG}-augmented \ac{RNN} models were able to correctly learn the translation of entities through the relations found in \ac{KGE}s and that \textit{KG-NMT (SemKGE)} improved generally the translation quality in comparison to other \ac{NMT} models, which concludes that a supervised annotation of entities with \ac{EL} is not entirely necessary. 

\begin{table}[htb]
\caption{Number of \ac{OOV} words across a baseline NMT model and KG-NMT models on the \textit{newstest} dataset.}
\vspace{-2mm}
\label{tab:unk}
\centering
\fontsize{9pt}{12pt}\selectfont
\begin{tabular}{l|cccc}
\textbf{Models} & \textbf{2015} & \textbf{2016} & \textbf{2017} & \textbf{2018} \\
\toprule
biRNN-lstm baseline & 6,004 & 9,559 & 9,707 & 9,383 \\
\midrule
KG-NMT(EL+KGE) & 4,427 & 6,524 & 6,603 & 6,914 \\
\midrule
KG-NMT(SemKGE) & 4,067 & 5,990 & 6,130 & 6,236 \\
\bottomrule
\end{tabular}
\end{table}


\begin{table*}[htb] 
\caption{Comparison between pre-trained monolingual embeddings and \ac{KGE}s.}
\label{tbl:comparison}
\centering
\fontsize{9pt}{12pt}\selectfont
\setlength{\tabcolsep}{1.8pt}
\begin{tabular}{@{}llccc|ccc|ccc|ccc@{}}
\textbf{Models} & & \multicolumn{3}{c}{\textbf{newtest2015}} & \multicolumn{3}{c}{\textbf{newtest2016}} & \multicolumn{3}{c}{\textbf{newtest2017}} & \multicolumn{3}{c}{\textbf{newtest2018}} \\
\toprule
&&  \rot{{BLEU}} & \rot{{METEOR}} & \rot{{chrF3}} & \rot{{BLEU}} & \rot{{METEOR}} & \rot{{chrF3}} & \rot{{BLEU}} & \rot{{METEOR}} & \rot{{chrF3}} & \rot{{BLEU}} & \rot{{METEOR}} & \rot{{chrF3}} \\
\hline
\multirowcell{2}{\textbf{Word-based}} & biRNN-lstm+MonoE & \textbf{21.59} & \textbf{40.54} & \textbf{45.37} & \textbf{24.12} & \textbf{42.82} & \textbf{47.37} & \textbf{20.05} & \textbf{39.42} & \textbf{43.90} & \textbf{27.15} & \textbf{46.13} & \textbf{49.35} \\
& KG-NMT(SemKGE) & 21.49 & 40.19 & 44.72 & 24.01 & 42.47 & 46.84 & 19.66 & 38.89 & 43.11 & 27.02 & 45.77 & 48.70 \\
\hline
\multirowcell{2}{\textbf{CopyM}} & biRNN-lstm+MonoE & \textbf{24.21} & \textbf{43.81} & \textbf{50.32} & \textbf{26.97} & \textbf{46.52} & \textbf{52.61} & \textbf{22.61} & \textbf{42.87} & \textbf{49.01} & \textbf{30.77} & \textbf{50.39} & \textbf{55.41} \\
& KG-NMT(SemKGE) & 24.08 & 43.43 & 49.72 & 26.70 & 46.08 & 52.05 &22.30 & 42.37 & 48.36 & 30.55 & 49.92 & 54.71 \\
\hline
\multirowcell{2}{\textbf{BPE}} & biRNN-lstm+MonoE & 19.65 & 39.24 & 47.58 & \textbf{25.13} & \textbf{44.66} & \textbf{53.54} & \textbf{20.93} & \textbf{41.41} & \textbf{50.33} &\textbf{ 28.42} & \textbf{48.00} & \textbf{55.98} \\
& KG-NMT(SemKGE) & \textbf{21.74} & \textbf{41.41 }& \textbf{50.04} & 24.86 & 44.32 & 52.59 & 20.45 & 40.62 & 49.45 & 28.02 & 47.51 & 55.16 \\
\bottomrule
\end{tabular}
\end{table*}

\begin{table*}[htb] 
\caption{Results of models in BLEU, METEOR, chrF3 on domain-specific testsets.}
\label{tab:spec}
\centering
\fontsize{9pt}{12pt}\selectfont
\setlength\tabcolsep{2pt}
\begin{tabular}{@{}llccc|ccc|ccc@{}}
\textbf{Models} & & \multicolumn{3}{c|}{ \textbf{ICD-10}} & \multicolumn{3}{c|}{ \textbf{IFRS }} & \multicolumn{3}{c}{ \textbf{IT} } \\
\midrule
& & \rot{{BLEU}} & \rot{{METEOR}} & \rot{{chrF3}} & \rot{{BLEU}} & \rot{{METEOR}} & \rot{{chrF3}} & \rot{{BLEU}} & \rot{{METEOR}} & \rot{{chrF3}} \\
\hline
\multirowcell{3}{\textbf{Word-based models}} & biRNN-lstm baseline\_adapt & 15.31 & 23.27 & 29.63 &\textbf{ 52.59} & \textbf{60.59} & \textbf{62.04} & 11.57 & 28.04 & 30.50 \\
& KG-NMT(EL+KGE)\_adapt & \textbf{21.08} & \textbf{31.07} & 36.93 & 52.38 & 60.55 & 61.86 & 21.78 & 40.29 & 42.75 \\
& KG-NMT(SemKGE)\_adapt & 20.79 & 30.70 & \textbf{37.00} & 51.58 & 59.18 & 60.05 & \textbf{23.41} & \textbf{41.71} & \textbf{44.42} \\
\hline
\multirowcell{3}{\textbf{CopyM models}} & biRNN-lstm baseline\_adapt & 16.59 & 26.49 & 39.12 & \textbf{52.91} & \textbf{61.68} & \textbf{64.34} & 13.87 & 31.61 & 36.10 \\
& KG-NMT(EL+KGE)\_adapt & \textbf{22.59} & \textbf{34.54} & \textbf{46.89} & 52.72 & 61.97 & 64.65 & 25.31 & 44.24 & 48.96 \\
& KG-NMT(SemKGE)\_adapt  & 22.24 & 34.10 & 46.74 & 51.91 & 60.33 & 62.51 & \textbf{26.84} &\textbf{ 45.81} & \textbf{50.38} \\
\hline
\multirowcell{3}{\textbf{BPE Models}} & biRNN-lstm baseline\_adapt & \textbf{41.98} & \textbf{50.81} & \textbf{66.87} & \textbf{66.74} & \textbf{74.83} & \textbf{84.52} & 27.83 & 47.01 & \textbf{55.91} \\
& KG-NMT(EL+KGE)\_adapt & N/A & N/A & N/A & N/A & N/A & N/A & N/A & N/A & N/A \\
& KG-NMT(SemKGE)\_adapt & 41.44 & 50.54 & 66.12 & 66.21 & 74.45 & 84.07 & \textbf{28.04} & \textbf{47.30} & 55.68 \\
\bottomrule
\end{tabular}
\end{table*}

\textit{\textbf{Monolingual Embeddings vs \ac{KGE}s}} - Table \ref{tbl:comparison} reports no significant difference between monolingual embeddings and \ac{KGE}s in terms of {\sc BLEU}, {\sc METEOR} and {\sc chrF3}. 
This finding is interesting since the monolingual embeddings contain billions of words, compared to the DBpedia \ac{KG} with 4.2 million entities. Taking a deeper look, our manual analysis showed that the \ac{OOV} words addressed by the monolingual embeddings were not in fact entities, but common words and the entities remained unknown. As an example, the \textit{RNN+MonoE} model translated incorrectly the entity \textit{Principal} into \textit{Wichtigste}, while the \textit{KG-NMT (SemKGE)} used the knowledge documented in the \ac{KG}s.\footnote{\url{http://dbpedia.org/resource/Principal_(school)} to translate the entity correctly} Moreover, \textit{RNN+MonoE} was unable to translate the entities \textit{UK} and \textit{Coastguard} while the \textit{KG-NMT (SemKGE)} generated the right translations. Therefore, \ac{KGE}s leverage the world knowledge better than pre-trained monolingual word embeddings for translating entities and we envisage that a combination of both is promising and may lead to further translation improvements.

\textbf{\textit{Continuous Training on Domain-Specific Parallel Datasets}} - Table~\ref{tab:spec} shows that the knowledge documented in \ac{KG}s is able to improve significantly the word-based models +5 BLEU, METEOR and chrF3 on the ICD-10 ontology and IT domain. However, no improvement is seen in the IFRS ontology (Financial domain) and all \ac{BPE} models. Investigating the data and results manually, we perceived that although the terminological expressions infrequently appear in the DBpedia \ac{KG}, e.g, 14,051 entities in the medical domain in comparison to 4.2 million in the whole graph, its application in the word-based models improved fairly the translations. The same applies to the IT domain, where the evaluation metric calculates a { \sc BLEU} score of 11.57 for the baseline system and 23.41 for the semantically-enriched \ac{KGE}s. However, the lack of improvement in the IFRS ontology was caused by the in-domain training data used in the continued training (domain adaptation). For example, the IFRS data used for continuous training already contained terminological expressions and therefore the adapted models ignored the values from the \ac{KGE}s. Differently, in the in-domain training of ICD-10 ontology and IT data, the terminological expressions do not appear. Therefore, the IFRS adapted models ignored the values from the \ac{KGE}s. For this reason, no improvement is seen in the word-based models for the financial domain. Regarding the \ac{BPE} models, the explanation lies in the capability of \ac{NN}s for estimating the translation of rarely seen terminological expressions when \ac{BPE} is applied. Basically, the operations applied by \ac{BPE} were not common due to very specific \ac{NE}, for example, names of diseases, for which the \ac{NMT} system was incapable of learning the translations from the \ac{KGE}s. In summary, \ac{KGE}s contribute to the translation of very domain-specific data. However, a further investigation of \ac{BPE} models in combination with KG-NMT methods is required, as they did not show consistent improvements across the targeted domains.  

\subsection{Reproducibility}
All experimental data, code, and model are publicly available.\footnote{\url{https://github.com/dice-group/KG-NMT}}

\subsection{Summary}

The main contributions of this paper can be summarized as follows:
\begin{itemize}
    \item We present the first \ac{KG}-augmented \ac{NMT} model, named KG-NMT.
    \item KG-NMT proposes two strategies for incorporating \ac{KG}s into \ac{NMT} models with consistent improvements over baseline.
    \item KG-NMT shows that \ac{KGE} leverages better real world knowledge, entities, in comparison two large pre-trained word embeddings trained on large corpora.
    \item KG-NMT is also capable of translating domain-specific dataset and ontologies.
\end{itemize}

\section{THOTH: Neural Translation and Enrichment of Knowledge Graphs}
\label{sec:THOTH}
\begin{figure*}[htb]
\centering
    \includegraphics[width=\textwidth]{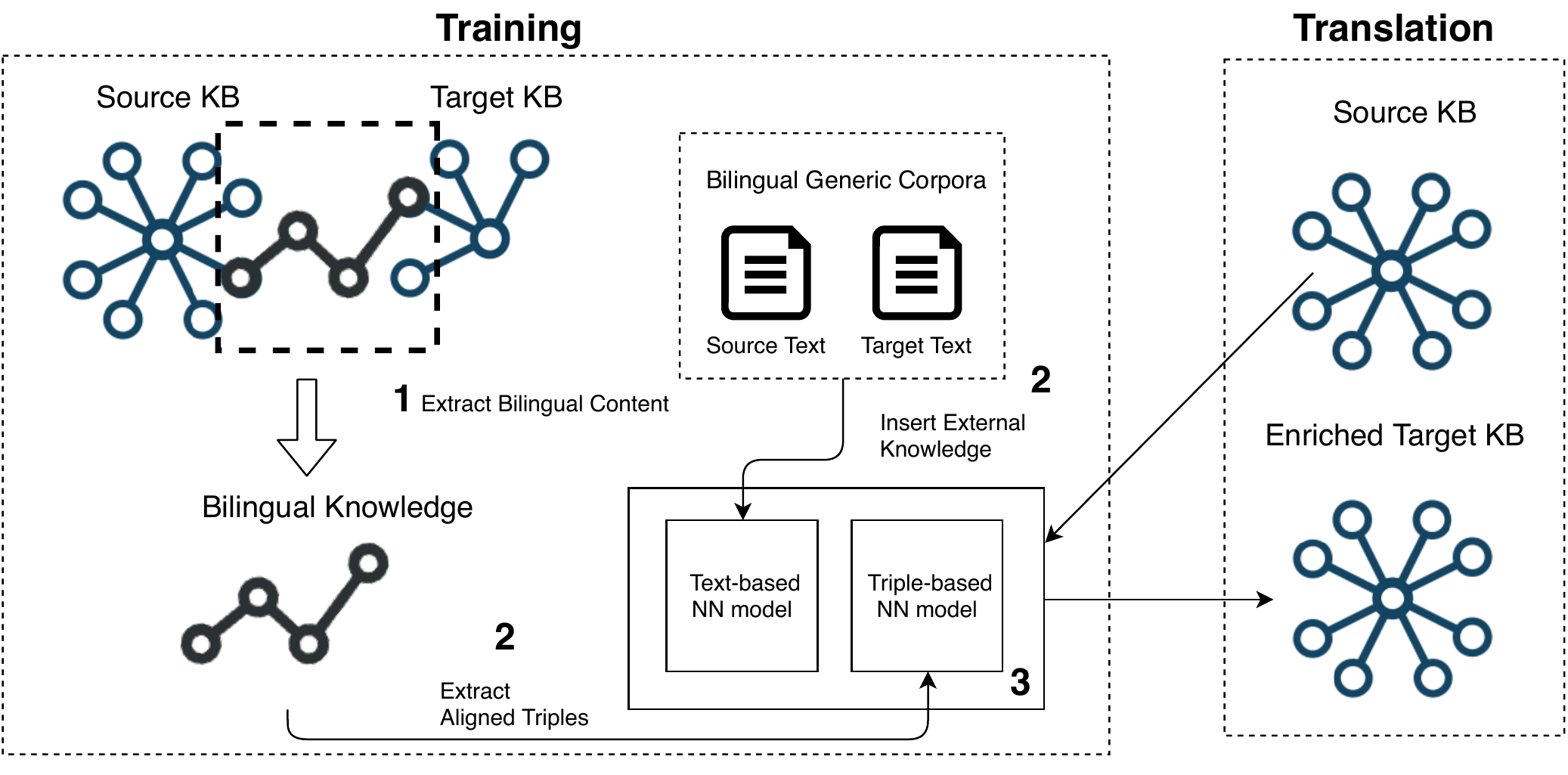}
    \caption{Overview of THOTH.}
    \label{fig:archTHOTH}
\end{figure*}

{\small\textit{FOR ALLEVIATING THE LACK OF MULTILINGUAL \ac{KG}}~\ref{challengeKG}},  we proposed THOTH, an approach for translating and enriching knowledge graphs by relying on neural models. The underlying idea behind our approach, THOTH, is based on the formal description of a translation problem as follows: \emph{Given that \ac{KG}s are composed of facts extracted from text, we can consider the facts (i.e., triples) as sentences, where \ac{URI}s are tokens and train a \ac{NMT} model to translate the facts from one language into another}. The enrichment process implemented by THOTH consists of two phases: the training phase and the translation phase.
The data gathering and preprocessing steps occur in the training phase, while the enrichment per se is carried out during the translation phase and consists of two steps: 1) translation and 2) enrichment. All steps carried out in THOTH are language-agnostic, which allow the use of other language-based \ac{KG}s. An overview can be found in Figure~\ref{fig:archTHOTH}.

\subsection{Training Phase}

While devising our approach, we perceived that one crucial requirement is that all resources and predicates in the source and target \ac{KG}s must have at least one label via a common predicate such as \texttt{rdfs:label}.\footnote{\url{https://www.w3.org/TR/webont-req/section-requirements}} This avoids the generation of inadequate resources. After establishing that, we divide THOTH into two models in order to take into account the challenge of translating datatype property values (i.e., texts) and object property values (i.e., entities). Trying to tackle both kinds of statements with a single model is likely to fail as labels can easily reach a length of 50 characters. Therefore, we divide the data gathering process into two blocks in order to be able to train two models. 

\textbf{Data gathering process} - First, we upload the source and target \ac{KG} into a SPARQL endpoint and query both graphs by looking for resources which have the same ``identity". Identical resources are usually connected via \texttt{owl:sameAs} links. However, aligned triples must not contain \texttt{owl:sameAs} as predicates in themselves. Second, we perform another SPARQL query for gathering only the labels of the aligned resources. Thus, we generate two bilingual training files, one with triples and another with labels (see Listing~\ref{lst:train} for an example). Once both training files are created, we split them into training, development, and test sets.    

\begin{lstlisting}[label=lst:train, float=htb, style=sparql, numbers=left, numberstyle=\tiny, 
caption=Sample of the triple-based training data]
EN: dbr:crocodile_dundee_ii	        dbo:country         dbr:united_states
DE: dbr_de:crocodile_dundee_ii      dbo:country         dbr_de:vereinigte_staaten 
EN: dbr:til_there_was_you           dbo:writer	        dbr:winnie_holzman	
DE: dbr_de:zwei_singles_in_l.a.     dbo:writer          dbr_de:winnie_holzman
\end{lstlisting}

\textit{\textbf{Preprocessing}} - Before we start training the triple- and text-based models, we tokenize both training data files. Subsequently, we apply \ac{BPE} models on them for dealing with \ac{OOV} words~\citep{sennrich2015neural}. \ac{BPE} is a form of data compression that iteratively replaces the most frequent pair of bytes in a sequence with a single, unused byte. Applying \ac{BPE} on the training data allows the translation models to translate words and sub-words and consequently improve their translation performance.

\textit{\textbf{Knowledge Graph Embeddings}} - Based on recent findings~\citep{moussallem2019augmenting}, we generate \ac{KGE}s from the aligned triples along with their labels by using \textit{fastText}. We rely on multinomial logistic regression~\citep{bohning1992multinomial} as a classifier in a supervised training implemented in \textit{fastText}. It assigns the entity's \ac{URI} to its surface forms. This technique enables the \ac{NN} to retrieve from \ac{KGE} the surface form of the entities through their \ac{URI}s.

\begin{figure*}[htb]
\centering
    \includegraphics[width=\textwidth]{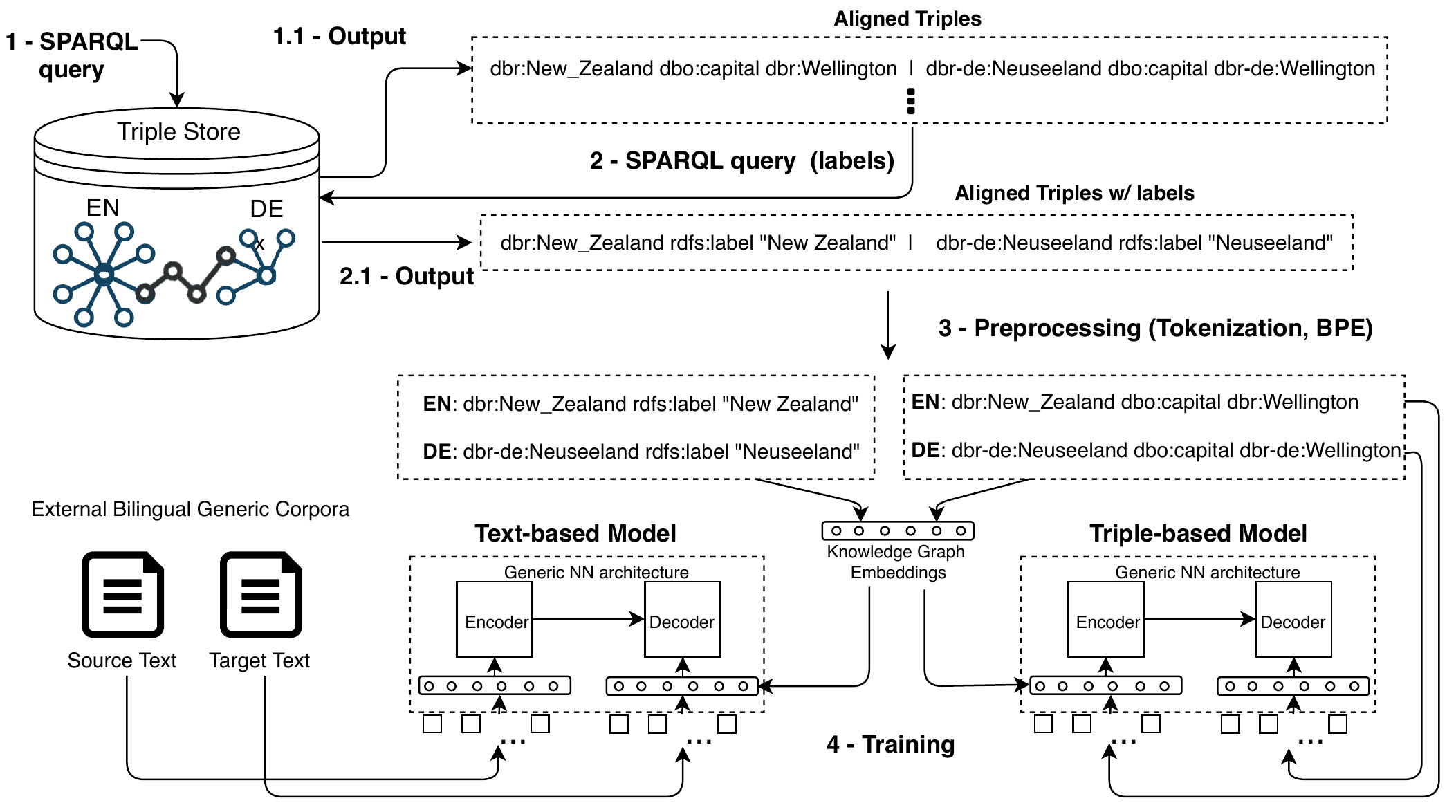}
    \caption{Training phase overview}
    \label{fig:train}
\end{figure*}

\textit{\textbf{Training}} - Both triple- and text-based models rely on a standard \ac{RNN} model. The difference between both models is the training data format. The Triple-based model is trained only with the aligned triples, while the text-based was trained with an external generic bilingual corpora. Additionally, both models are augmented with the same \ac{KGE} model. The idea of using \ac{KGE} is to maximize the vector values of the triple-based and text-based \ac{NMT} embeddings layers while training their models. An overview of the training phase can be found in Figure~\ref{fig:train}.

\subsection{Translation Phase}

\begin{figure*}[htb]
\centering
    \includegraphics[width=0.85\textwidth]{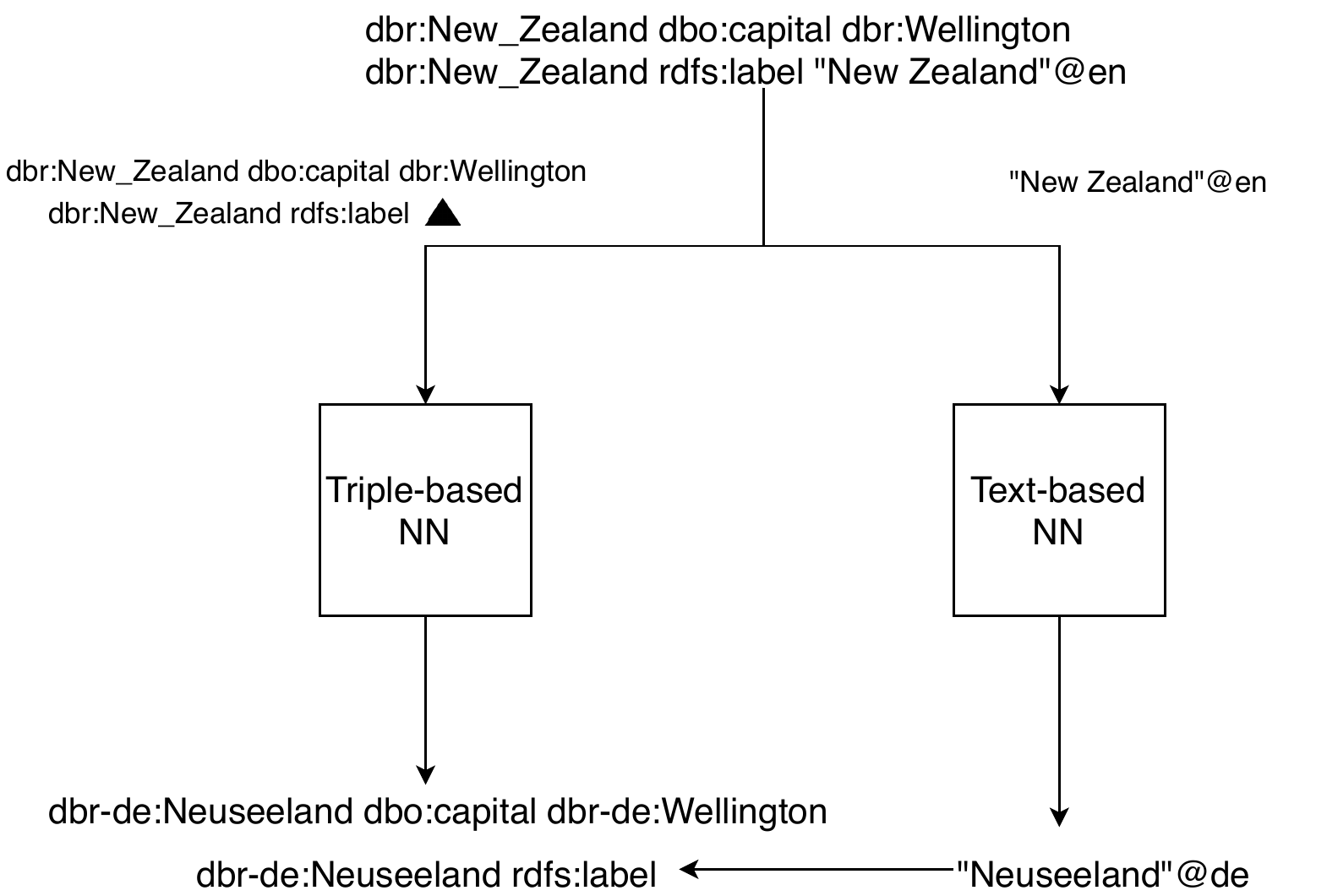}
    \caption{Translation phase overview}
    \label{fig:trans}
\end{figure*}

Here, THOTH expects the entire source \ac{KG} as an input to be translated and enriched into the target language as an output. To this end, THOTH first relies on a script, which is responsible for splitting the \ac{KG} triples that comprises only the resources in one file and the triples that contain literals as objects in a different file. Once the division is done, and two set files are generated, THOTH starts translating the triples only with resources. After that, THOTH has to deal with the triples which have labels, and such triples are handled differently. The subject and predicate of the triples are sent to the Triple-based model along with a special character in the place of its object. This special character simply tells the model to ignore the value and copy it to the target. In turn, the Text-based \ac{NMT} model translates only the object. We argue that the Text-based model can translate the labels correctly since its model was augmented with a \ac{KGE} model representing the \ac{URI}s of both \ac{KG}s, source and target. Afterwards, subject and predicate are attached with their object literal in a triple again. Finally, the two different files are combined into one again resulting in a translated \ac{KG}. An overview of the training phase can be found in Figure~\ref{fig:trans}. 

Once the translation step is complete, THOTH gets the translated \ac{KG}, and the original target (German) \ac{KG} used in the training part and combines both into a single \ac{KG}. The idea here is to enrich the original \ac{KG} with translated triples. When conflicts of values happen, e.g., the triples match partially, and duplicated triples appear between the original \ac{KG} and the translated \ac{KG}, we opt to maintain the triples from the original \ac{KG} as THOTH's aim is not to produce a newly translated \ac{KG} but enrich the original one.

\subsection{Evaluation} 

We designed our evaluation in three-fold set. First, we measured the performance of THOTH using an automatic \ac{MT} evaluation metric, {\sc BLEU}, along with its translation accuracy. Second, we evaluated THOTH extrinsically by comparing the German DBpedia with the German translation of the English DBpedia on two tasks: Fact Validation and Entity Linking. Third, we ran a manual intrinsic evaluation of the translation. We choose German as a target language because of the abundance of benchmarking systems and datasets for this pair.

\textit{\textbf{Settings}} - In our experiments, both the triple-based and the text-based \ac{NMT} models are built upon an \ac{RNN} architecture using a bi-directional 2-layer \ac{LSTM} encoder-decoder model with attention mechanism~\citep{bahdanau2014neural}. The training uses a batch size of 32 and the stochastic gradient descent with an initial learning rate of 0.0002. We set the dimension of the word embeddings to 500 and the internal embeddings of hidden layers to size 500. The dropout is set to 0.3 (naive). We use a maximum sentence length of 50, a vocabulary of 50,000 words and a beam size of 5. All experiments are performed with the OpenNMT framework~\citep{2017opennmt}. In addition, we encode the triples and words using \ac{BPE}~\citep{sennrich2015neural} with 32,000 merge operations. For training the text-based model, our training set consists of a merge of all parallel training data provided by the \ac{WMT} tasks\footnote{\url{http://www.statmt.org/wmt18/translation-task.html}}, obtaining after preprocessing a corpus of five million sentences with 79M running words. In the triple-based model, we use the bilingual alignments from the English, and German versions of DBpedia\footnote{We selected the subsets of mapping-based objects and labels to evaluate the quality of our approach since they are the most used ones for training Linked-Data \ac{NLP} approaches.} for training. This alignment contains 346,373 subjects, 292 relations and 208,079 objects in 1,012,681 triples. We divide this data into 80\% training, 10\% development and 10\% test. Overall, the English \ac{KG} contains 4.2 million entities, 661 relations, and 2.1 million surface forms, while the German version has 1 million entities, 249 relations, and 0.5 million surface forms. Additionally, we train the \ac{KGE} on both DBpedia versions using the \textit{fastText} algorithm with a vector dimension size of 500 and a window size of 50 by using 12 threads with hierarchical softmax.

\textit{\textbf{Translation task}} - The overall enrichment quality of THOTH is measured by working through different steps. Firstly, we evaluate the translations automatically by computing a translation accuracy with {\sc BLEU}~\citep{papineni2002bleu} score. In the subsequent evaluation steps, we investigate THOTH's performance on a full \ac{KG} translation setting. In this case, we use THOTH models for translating and enriching all \ac{CBD} resources of English DBpedia to an enriched-German DBpedia version. The further extrinsic evaluation steps are described below. 

\textit{\textbf{Fact validation task}} - We selected FactBench---a multilingual benchmark dataset for the evaluation of fact validation algorithms~\citep{gerber2015defacto}---for our experiments. FactBench contains positive and negative facts. We only use the 750 positive facts distributed over 10 relations as reference data in our experiment. Our aim is to check the number of true facts which existed in the original \ac{KG} (i.e., in the German version of DBpedia) and how many true triples THOTH was able to add to the \ac{KG} through enrichment. We used 5 of the 10 predicates in our evaluation data set, i.e., \texttt{award}, \texttt{birthplace}, \texttt{deathplace}, \texttt{leader}, \texttt{starring} because the other predicates do not lead to sufficient training data. Overall, our evaluation dataset consists of  a total count of 375 facts. 

\textit{\textbf{NLP task}} - Our idea here is to exploit the graphs connections from the enriched-German DBpedia (THOTH) \ac{KG} to improve a given \ac{EL} system on a disambiguation task. We chose MAG, our multilingual \ac{EL} system introduced by \cite{moussallem2017mag}, which is language- and \ac{KG}-agnostic. MAG does not require any training even though shows competitive results. Also, we selected GERBIL~\citep{gerbil} as a benchmarking platform. As the evaluation is on the German language, we uploaded four German datasets to GERBIL. 



\subsection{Results}

In this section, we report the results of THOTH's enrichment in the German DBpedia on the settings mentioned above.

\textit{\textbf{Translation results}} - We evaluated our translation on the test set of the bilingual data we extracted via SPARQL queries. THOTH achieved a {\sc BLEU} score of 65.47, which is superior to the state-of-the-art translation scores achieved on natural language~\citep{edunov2018understanding}. 

\begin{figure*}[htb]
\centering
    \includegraphics[width=0.85\textwidth]{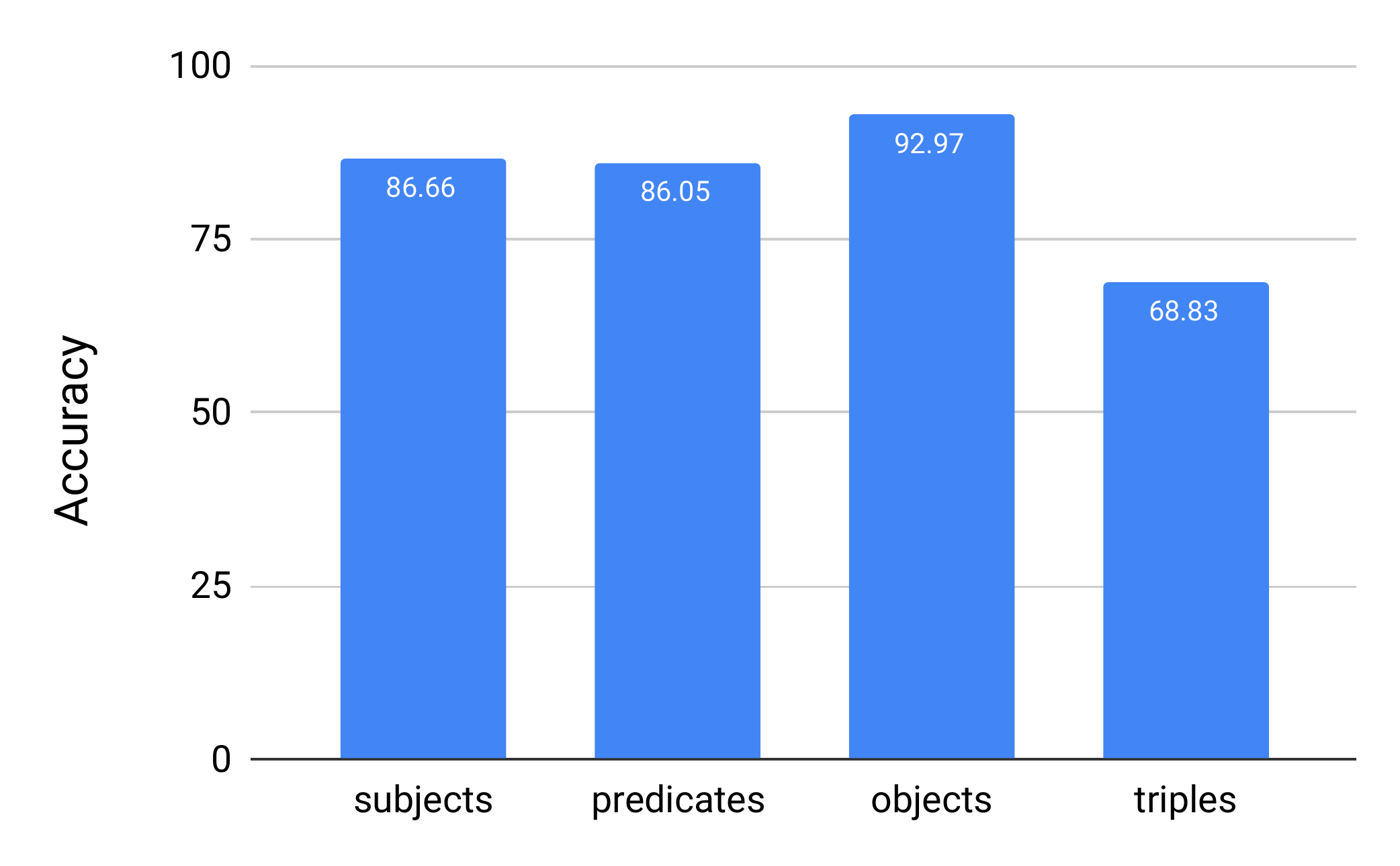}
    \caption{Overall translation accuracy}
    \label{fig:accu}
\end{figure*}

Given that it is not possible to infer the quality of a given translation only relying on one automatic evaluation metric, we created an additional evaluation script that computes the exact string match of subjects, predicates, and objects between an output and a reference translation triple. Additionally, we also computed the overall triple accuracy. Figure~\ref{fig:accu} depicts the accuracy results of THOTH's output in comparison to the German test set. THOTH achieved up to 80\% accuracy for subjects, predicates, and objects. As expected, THOTH's accuracy decreased to 68.83\% when measuring entire triples. We analyzed the results manually to understand this drop in the performance. Our manual analysis suggests that the poorer performance w.r.t. triples is linked to the partially weak disambiguation power of the underlying \ac{KGE} model, which assigned the same vector value for similar predicates. Our results confirm that \ac{NN}s along \ac{KGE} can support a full \ac{KG} translation by considering the consistent quality of THOTH translations.

\begin{figure}[htb]
\centering
    \includegraphics[width=\textwidth]{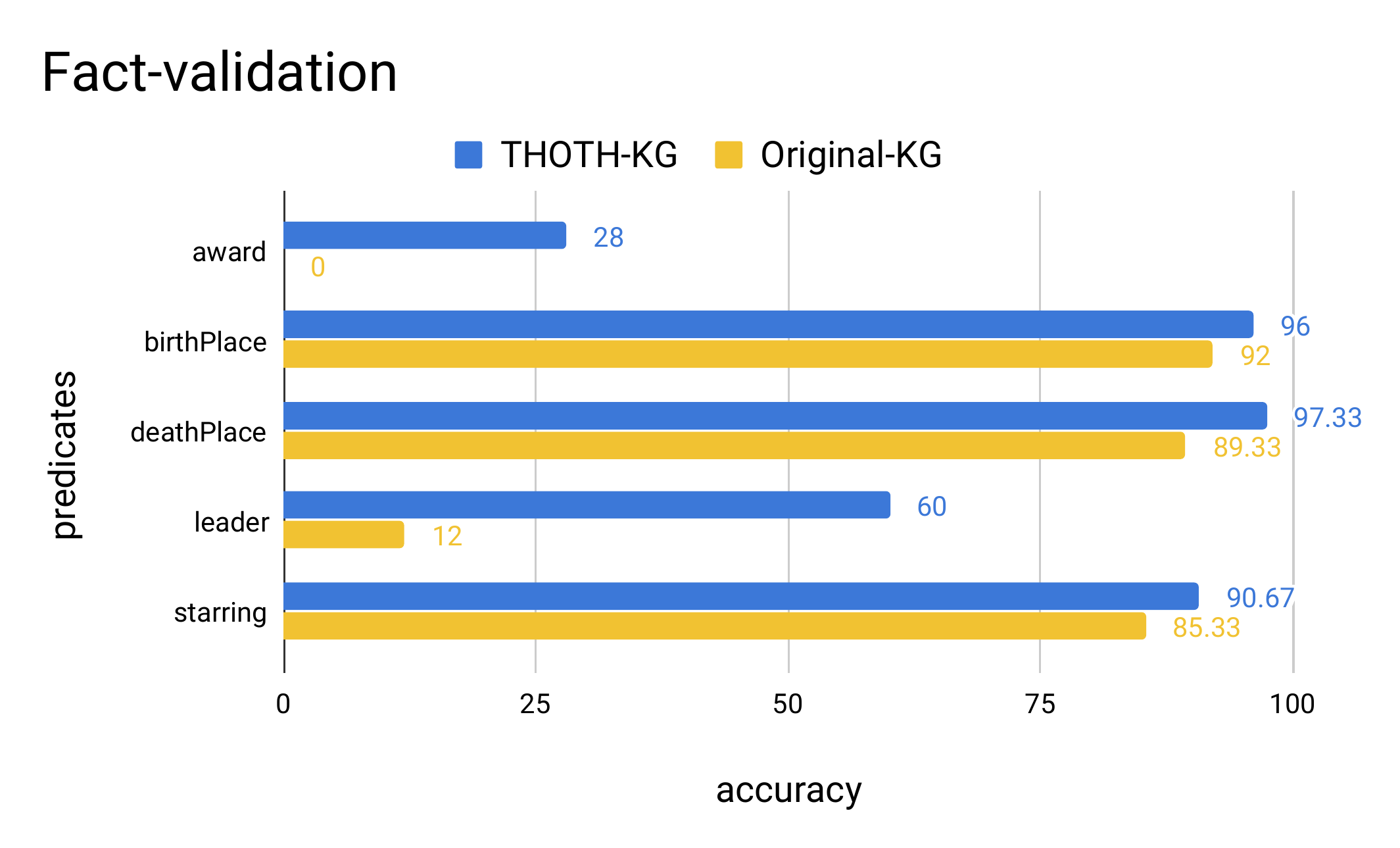}
    \caption{A comparison between the enriched-German DBpedia (THOTH) \ac{KG} with the original German DBpedia on the validation of facts.}
    \label{fig:fact-validation}
\end{figure}

\textit{\textbf{Fact validation results}} - Here, we used THOTH to translate the entire English DBpedia to German. In this case, we do not have a gold standard translation to compare automatically. Therefore, we evaluated the THOTH's enrichment capability in the perspective of a fact-validation task. The main goal here was to check if THOTH could enrich the original German \ac{KG} with new correct facts which were not present in its original version. Figure~\ref{fig:fact-validation} reports an improvement of 18.4\% across all predicates. THOTH led to a significant increase in the number of correct facts in the original \ac{KG}.  

\textit{\textbf{Entity Linking results}} - For this evaluation, we used the optimal parameter configuration for MAG described by~\cite{moussallem2017mag}. Table~\ref{tab:el-thoth} reports the results of MAG in two configuration sets, one with original German DBpedia and another with the \textit{Enriched-German DBpedia (THOTH)} as reference \ac{KG}s. The version of MAG running on the translated \ac{KG} achieves significantly better results than that running on the original \ac{KG}. The average improvement across all datasets is around 19\% in F-measure. The results of the \textit{German abstracts} data set and \textit{N$^3$ news.de} are surprisingly high. We sampled the results manually, and we established that the results were correct. We also investigated the creation of both benchmarking datasets, and concluded that at the time of their creation, the links used in both were based on the English DBpedia as an auxiliary \ac{KG}. Therefore, when THOTH translated the English \ac{KG} to German and enriched the original German DBpedia with English knowledge, MAG was able to get very high HITS scores for many resources. 
\begin{table}[hbt]
\setlength\tabcolsep{2pt}
\centering
\fontsize{9pt}{12pt}\selectfont
\caption{Micro results in a comparison between German DBpedia \ac{KG} with Enriched-German DBpedia (THOTH) \ac{KG} in MAG.}
\label{tab:el-thoth}
\begin{tabular}{@{} lcccccc @{}}
\toprule
\multirow{2}{*}{ \textbf{Datasets} } & \multicolumn{3}{c}{ \textbf{MAG-DBpedia-KG} } & \multicolumn{3}{c}{ \textbf{MAG-THOTH-KG} } \\
& F-measure & Precision & Recall & F-measure & Precision & Recall \\
\midrule
German Abstracts & 0.78 & 0.79 & 0.76 & \textbf{0.97} & \textbf{0.99} & \textbf{0.96} \\
N$^3$ news.de & 0.77 & 0.78 & 0.76 & \textbf{0.98} & \textbf{0.99} & \textbf{0.97} \\
VoxEL-strict & 0.40 & 0.46 & 0.35 & \textbf{0.70} &\textbf{0.81} & \textbf{0.61} \\
VoxEL-relaxed & 0.57 & 0.57 & 0.57 & \textbf{0.64} & \textbf{0.64} & \textbf{0.64} \\
\bottomrule
\end{tabular}
\end{table}

\subsection{Reproducibility}

all experimental data, code, models are publicly available.\footnote{\url{https://github.com/dice-group/THOTH}}

\subsection{Summary}

The main contributions of this paper can be summarized as follows:

\begin{itemize}

    \item We present a novel approach based on \ac{NN}s along with \ac{KGE}s for translating and enriching \ac{KG}s across languages.
    
    \item THOTH achieves a translation accuracy of 88.56\% across all elements of a triple. 
    
    \item THOTH improves the quality of the original German DBpedia significantly in both the fact checking and the \ac{EL} tasks: 18.4\% for fact validation and 19\% for \ac{EL}.
        
\end{itemize}

\chapter{Conclusions and Outlook}
\label{ch:conclusion}

This chapter concludes the thesis by summarizing our results in Section \ref{sec:conclusion} and giving an outlook on future research directions in Section \ref{sec:outlook}.

\section{Conclusions}
\label{sec:conclusion}

Handling entities across distinct \ac{NLP} tasks is a difficult task. With thesis, we showed that real-world \acp{KG} can contribute to improve the results of \ac{EL}, \ac{MT} and \ac{NLG} tasks. Therewith, we answered our main research question posed in Section~\ref{challenges}: 
\begin{itemize}
    \item[RQ.] Can \acp{KG} alleviate the ambiguity problem and be used to improve the quality of automatic text translation and generation? 
\end{itemize} 

In addition, we addressed its respective challenges by (1) devising a multilingual \ac{KG}-based \ac{EL} approach; (2) developing a multilingual \ac{RDF}-to-Text verbalizer; (3) creating the first neural- and \ac{KG}-based \ac{REG} model; (4) creating the first \ac{KG}-augmented \ac{NMT} model; (5) designing the first neural translation approach for enriching low resource \ac{KG}s. 

\subsection{MAG: A Multilingual, Knowledge-base Agnostic and Deterministic Entity Linking Approach}

We asked the following two questions in Section~\ref{challengeNED}:

\begin{itemize}
\item[RQ1.] Can a \ac{KG}-based \ac{EL} approach achieve a similar performance across languages? 

\item[RQ2.] Does a language-based \ac{KG} influence the disambiguation quality?
\end{itemize}

We answered the above mentioned questions by presenting MAG, a \ac{KB}-agnostic and deterministic approach for multilingual \ac{EL}. MAG outperforms state of the art on all non-English data sets. In addition, MAG achieves a performance similar to state of the art on English data sets. An average 0.63 F-measure places MAG 1st out of 13 annotation systems. Furthermore, we analyzed the influence of different indexing and searching methods, as well as the influence of the data set structure in a fine-grained evaluation. We also provided a context search without relying on machine learning, as previously done. Moreover, we showed that current ML-based \ac{EL} approaches are strongly biased due to their learned model. This behavior can be seen in multilingual data sets. We also deployed and analyzed the influence of acronyms and last names. 

\subsection{RDF2PT: Generating Brazilian Portuguese Texts from \ac{RDF} Data}

We asked the following question in Section~\ref{challengeNLG}:

\begin{itemize}
\item[RQ3:] Can \ac{KG}s as input support the generation of multilingual text?
\end{itemize}

We answered the above mentioned question by presenting RDF2PT, the first approach that verbalizes \ac{RDF} data to Brazilian Portuguese texts.  Compared with human texts, RDF2PT generates texts with high fluency and clarity. We identified essential challenges for generating multilingual texts from \ac{RDF} using a rule- and template-based approach. Moreover, we extended RDF2PT to Spanish and English~\citep{bengal, Ngonga2019}, thus demonstrating that \ac{KG}s definitely support the multilingualism in \ac{NLG}.  

\subsection{NeuralREG: An End-to-End Approach to Referring Expression Generation}

We asked the following question in Section~\ref{challengeNLG}:

\begin{itemize}
    \item[RQ4:] Can \ac{KG}s be used for accomplishing the full \ac{REG} task?
\end{itemize}

We answered the above mentioned question by introducing NeuralREG, the first end-to-end approach based on neural networks for the \ac{REG} task. NeuralREG generates referring expressions for discourse entities by simultaneously selecting form and content without any need for feature extraction techniques. NeuralREG showed that the neural model substantially improves over two strong baselines, both in terms of the accuracy of referring expressions and the fluency of lexicalized texts.

\subsection{KG-NMT: Utilizing Knowledge Graphs for Neural Machine Translation Augmentation}

We asked the following question in Section~\ref{challengeNMT}:

\begin{itemize}
\item[RQ5:] Can an \ac{NMT} model enhanced with a bilingual \ac{KG} improve translation quality?
\end{itemize}

We answered the above mentioned question by presenting KG-NMT. KG-NMT is the first augmentation methodology, which relies on the use of \acp{KG} to improve the performance of \ac{NMT} systems for translating domain-specific expressions and named entities in texts. We implemented two strategies for incorporating \ac{KGE}s into \ac{NMT} models that work on word- and sub-word units-based models. Additionally, we carried out an extensive evaluation with a manual analysis, which showed consistent translation improvements provided by incorporating DBpedia \ac{KG} in \ac{NMT}. The overall methodology can be applied to any \ac{NMT} model since it does not modify the main \ac{NMT} model structure and also allows the replacement of different \ac{EL} systems.

\subsection{THOTH: Translating and Enriching Low-Resource 
\ac{KG}}

We asked the following two questions in Section~\ref{challengeKG}:

\begin{itemize}
    \item[RQ6:]  Can \ac{NMT} support a full (triples and labels) translation of \ac{KG}s?
    \item[RQ7:]  Can an artificially-enriched \ac{KG} improve the performance of a system on \ac{NLP} tasks?
\end{itemize}

We answered the above mentioned questions by introducing THOTH, the first neural-based approach for translating and enriching \ac{KG}s from different languages. THOTH is a promising approach that achieves a translation accuracy of 88.56\%. Moreover, its enrichment improves the quality of the German DBpedia significantly, as we report +18.4\% accuracy for fact validation and +19\% F$_1$ for entity linking. THOTH relies on two different \ac{RNN}-based \ac{NMT} models along with \ac{KGE}s for translating triples and texts jointly. We carried out an extensive evaluation set for certifying the quality of our approach.

\section{Outlook}
\label{sec:outlook}

Extensions of our contributions could be performed in multiple directions: exploiting other \ac{KG} features, such as ontologies within \ac{NMT} models, and extending our \ac{NLG} approaches to other languages. Moreover, a further investigation of our \ac{KG} translation and enrichment approach on other \ac{NN} architectures along other \ac{KGE} algorithms has to be carried out.

\subsection{Exploiting \ac{KG} features}

The syntactic disambiguation problem still lacks good solutions. For instance, the English language contains irregular verbs like ``set'' or ``put''. Depending on the structure of a sentence, it is not possible to recognize their verbal tense, e.g., present or past tense. Even statistical approaches trained on huge corpora may fail to find the exact meaning of some words due to the structure of the language. Although this challenge has successfully been dealt with since \ac{NMT} has been used for European languages~\citep{bojar2017findings}, implementations of \ac{NMT} for some non-European languages have not been fully exploited (e.g., Brazilian Portuguese, Latin-America Spanish, Hindi) due to the lack of large bilingual data sets on the Web to be trained on. We suggest using ontology properties via semantic annotations to alleviate the syntactic issue of irregular verbs. For instance, the sentence ``Anna usually put her notebook on the table for studying" may be annotated using a given vocabulary by triples. Thus, the verb ``put", which is represented by a predicate that groups essential information about the verbal tense, may support the generation step of a given \ac{NMT} system. This sentence usually fails when translated to morphologically rich languages, such as Brazilian-Portuguese and Arabic, for which the verb influences the translation of ``usually" to the past tense. In this case, the ontology properties contained in a \ac{KG} may support the problem of finding a specific rule behind relationships between source and target texts in the training phase~\citep{moussallem2015using}. Some researchers, including 
~\cite{harriehausen2012semantic, seo2009syntactic}, have used ontology properties to disambiguate words in \ac{MT} systems. However, the ontologies were not exploited in the context of \ac{NMT}.

To include ontology properties as features in \ac{NMT} models, some steps need to be addressed. Currently, the Ontology-Lexica Community Group\footnote{\url{https://www.w3.org/community/ontolex/}} at W3C has combined efforts to represent lexical entries, with their linguistic information, in ontologies across languages. Modeling different languages using the same model may provide alignment between the languages, where it is possible to infer new rules using the language dependency graph structure and visualize a similarity among languages.~\footnote{This insight is already supported by a recent publication at the Cicling Conference~\url{https://www.cicling.org/2017/posters.html} named ``The Fix-point of Dependency Graph -- A Case Study of Chinese-German Similarity" by Tiansi Dong et al.} 

\subsection{Combining 
the models of \ac{NMT} with \ac{NLG}}

\begin{figure}[htb]
\centering
    \includegraphics[width=\textwidth]{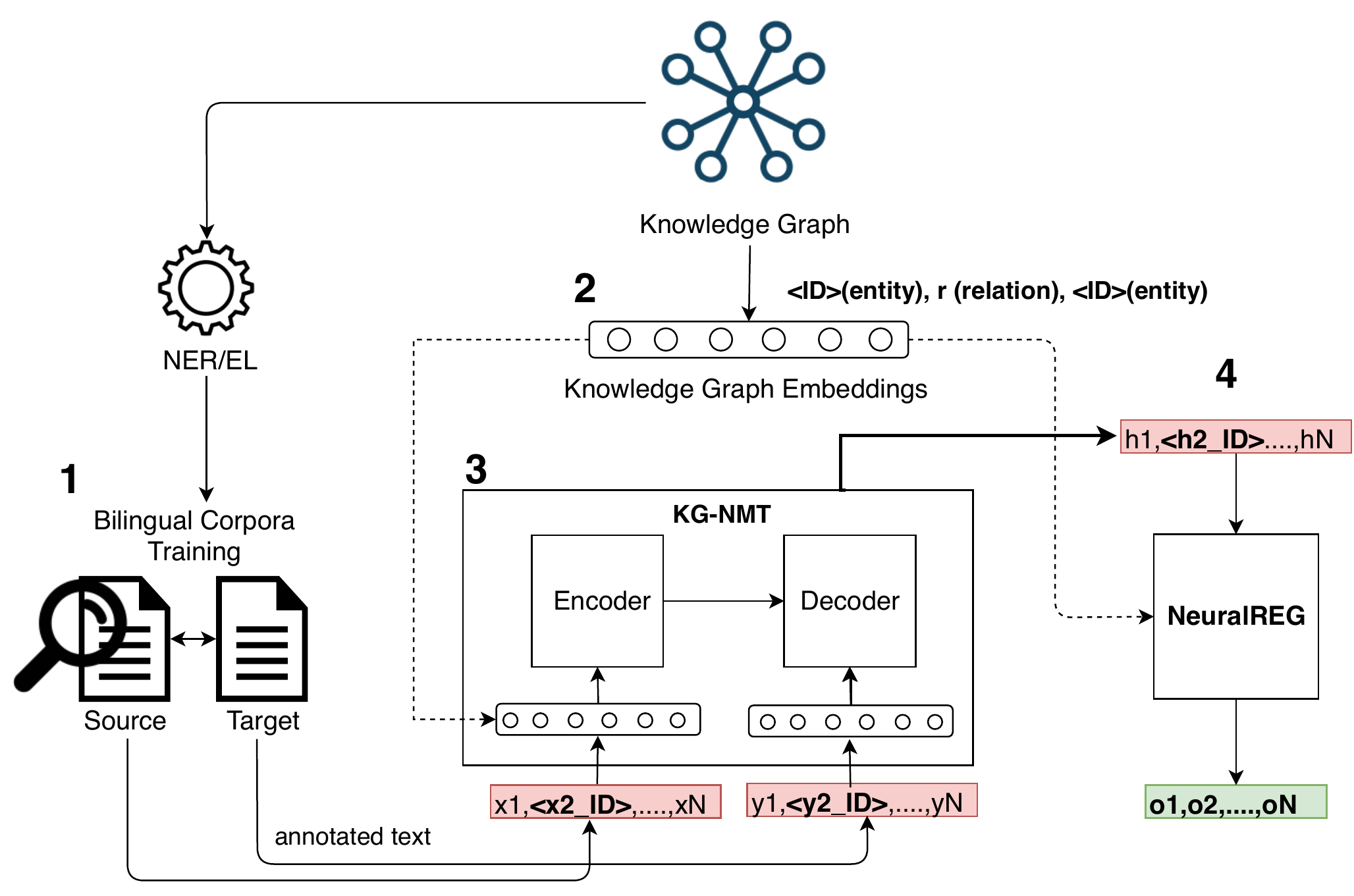}
    \caption{Overview of the combination of KG-NMT + NeuralREG.}
    \label{fig:archKG-NMT+NeuralREG}
\end{figure}

During the course of this thesis, we realized that combining a \ac{KG}-augmented \ac{NMT} model with an \ac{REG} model can be fruitful for improving the fluency of entities in text translation. The idea relies on feeding the NeuralREG model with the output of KG-NMT, but instead of generating the final translation with KG-NMT, it keeps the \ac{URI} of the entities in the \ac{NMT} output. Figure~\ref{fig:archKG-NMT+NeuralREG} depicts the general idea of this promising direction. 

\subsection{Deeper Investigation of Multilingualism in \ac{NLG}}

During the development of RDF2PT, some challenges became clear while handling morphologically rich languages. For example, recognizing gender continues to be a hard task. For example, in ``\texttt{Os Lus\'{i}adas \'e uns obra liter\'{a}ria}", the determiner \texttt{uns} should be feminine and singular, because \texttt{obra} is singular and has a feminine gender. However, it is accorded to the subject \texttt{Os Lus\'{i}adas}. Although our approach, NeuraLREG, was capable of improving the generation of pronouns, it still requires improvements and investigation with regard to other languages. We hence envisage adding gender as a feature for improving the text generation. 

Another observed challenge was the generation of coordinated sentences by RDF2PT and NeuralREG, which helped the users in our experimental setup recognize if the models or humans generated a given text. This behavior arises because humans are likely to write subordinate sentences. For example, while RDF2PT can generate \texttt{Albert Einstein foi um cientista e ele nasceu em Ulm.} (eng: Albert Einstein was a scientist and he was born in Ulm), a human would write this same sentence in the following way, \texttt{Albert Einstein foi um cientista que/cujo nasceu em Ulm} (eng: Albert Einstein was a scientist who was born in Ulm). This difference was crucial in the perspective of our evaluators. Therefore, the generation of subordinate sentences must be investigated. 

\subsection{Translating \ac{KG}s with other \ac{NN}s}

While analyzing THOTH's output, we noticed some mistranslations of similar predicates that were responsible for decreasing the accuracy of the triple translation. For example, the following English source triple \texttt{dbr:zenyatt\`{a}\_\-mon\-dat\-ta dbo:\-art\-ist dbr:the\_police} was translated into \texttt{dbr\_de:ze\-ny\-att\`{a}\_mon\-da\-tta dbo:\-pro\-du\-cer dbr\_de:the\_police}. This example shows that THOTH translated the subject and object correctly. However, the predicate was incorrect and was mistranslated from dbo:artist to dbo:producer. A similar problem occurred while translating the triple, \texttt{dbr:albert\_einstein dbo:citizenship dbr:Switzerland} to \texttt{dbr:albert\_einstein dbo:birthplace dbr:\-der\_Schweiz}. After a manual analysis, we identified that both cases happened because THOTH could not distinguish the predicates that share the same domain and range. In a more in-depth analysis, we perceived that the predicates mentioned above are very close to each other in the vector space, thus complicating the disambiguation process of \ac{NN} models. The performance of THOTH was not affected by these false triples since they were automatically removed in the enrichment step.
After this manual analysis of the results, we believe that addressing the problem of similar predicates (e.g., through novel embedding techniques) can enhance the translation quality of THOTH. The application of sub-graphs~\citep{cao2018link}  and other \ac{NN} architectures, such as Transformer~\citep{vaswani2017attention}, for improving the disambiguation of similar predicates are promising paths.

\cleardoublepage 
\phantomsection
\addcontentsline{toc}{chapter}{References}
\renewcommand{\bibname}{References}
\begin{spacing}{\shcompressedspacing}
  \raggedright
  \bibliography{shthesis-lit}

\begin{thebibliography}{119}
\providecommand{\natexlab}[1]{#1}
\providecommand{\url}[1]{\texttt{#1}}
\expandafter\ifx\csname urlstyle\endcsname\relax
  \providecommand{\doi}[1]{doi: #1}\else
  \providecommand{\doi}{doi: \begingroup \urlstyle{rm}\Url}\fi

\bibitem[Aprosio et~al.(2013)Aprosio, Giuliano, and
  Lavelli]{aprosio2013towards}
Aprosio, A.~P., C.~Giuliano, and A.~Lavelli.
\newblock Towards an automatic creation of localized versions of DBpedia.
\newblock In \emph{International Semantic Web Conference}, pages 494--509.
  Springer, 2013.

\bibitem[Arnold(1994)]{arnold1994machine}
Arnold, D.
\newblock \emph{Machine translation: an introductory guide}.
\newblock Blackwell Pub, 1994.

\bibitem[Auer et~al.(2007)Auer, Bizer, Kobilarov, Lehmann, Cyganiak, and
  Ives]{auer2007dbpedia}
Auer, S., C.~Bizer, G.~Kobilarov, J.~Lehmann, R.~Cyganiak, and Z.~Ives.
\newblock Dbpedia: A nucleus for a web of open data.
\newblock In \emph{The semantic web}, pages 722--735. Springer, 2007.

\bibitem[Bahdanau et~al.(2014)Bahdanau, Cho, and Bengio]{bahdanau2014neural}
Bahdanau, D., K.~Cho, and Y.~Bengio.
\newblock {Neural machine translation by jointly learning to align and
  translate}.
\newblock \emph{arXiv preprint arXiv:1409.0473}, 2014.

\bibitem[Banerjee and Lavie(2005)]{banerjee2005meteor}
Banerjee, S. and A.~Lavie.
\newblock METEOR: An Automatic Metric for MT Evaluation with Improved
  Correlation with Human Judgments.
\newblock In \emph{Proceedings of the ACL Workshop on Intrinsic and Extrinsic
  Evaluation Measures for MT and/or Summarization}, pages 65--72. ACL, 2005.

\bibitem[Bar-Hillel(1960)]{Bar-Hillel1960}
Bar-Hillel, Y.
\newblock {The present status of automatic translation of languages}.
\newblock In \emph{Advances in computers}, volume~1, pages 91--163. Elsevier,
  1960.

\bibitem[Belinkov and Glass(2019)]{belinkov-glass-2019-analysis}
Belinkov, Y. and J.~Glass.
\newblock Analysis Methods in Neural Language Processing: A Survey.
\newblock \emph{Transactions of the Association for Computational Linguistics},
  7:\penalty0 49--72, March 2019.
\newblock \doi{10.1162/tacl_a_00254}.
\newblock URL \url{https://www.aclweb.org/anthology/Q19-1004}.

\bibitem[Berners-Lee et~al.(2001)Berners-Lee, Hendler, and
  Lassila]{Berners-Lee2001}
Berners-Lee, T., J.~Hendler, and O.~Lassila.
\newblock {The Semantic Web}.
\newblock \emph{Scientific american}, 284\penalty0 (5):\penalty0 34--43, 2001.

\bibitem[Bisazza and Federico(2016)]{bisazza2016survey}
Bisazza, A. and M.~Federico.
\newblock {A survey of word reordering in statistical machine translation:
  Computational models and language phenomena}.
\newblock \emph{Computational Linguistics}, 2016.

\bibitem[B{\"o}hning(1992)]{bohning1992multinomial}
B{\"o}hning, D.
\newblock Multinomial logistic regression algorithm.
\newblock \emph{Annals of the institute of Statistical Mathematics},
  1:\penalty0 197--200, 1992.

\bibitem[Bojar et~al.(2017)Bojar, Chatterjee, Federmann, Graham, Haddow, Huang,
  Huck, Koehn, Liu, Logacheva, et~al.]{bojar2017findings}
Bojar, O., R.~Chatterjee, C.~Federmann, Y.~Graham, B.~Haddow, S.~Huang,
  M.~Huck, P.~Koehn, Q.~Liu, V.~Logacheva, et~al.
\newblock {Findings of the 2017 conference on machine translation (WMT17)}.
\newblock In \emph{Proceedings of the Second Conference on Machine
  Translation}, pages 169--214, 2017.

\bibitem[Bonatti et~al.(2019)Bonatti, Decker, Polleres, and
  Presutti]{bonatti2019knowledge}
Bonatti, P.~A., S.~Decker, A.~Polleres, and V.~Presutti.
\newblock Knowledge graphs: new directions for knowledge representation on the
  semantic web (Dagstuhl seminar 18371).
\newblock \emph{Schloss Dagstuhl-Leibniz-Zentrum fuer Informatik}, 2019.

\bibitem[Brown et~al.(1990)Brown, Cocke, Pietra, Pietra, Jelinek, Lafferty,
  Mercer, and Roossin]{brown1990statistical}
Brown, P.~F., J.~Cocke, S.~A.~D. Pietra, V.~J.~D. Pietra, F.~Jelinek, J.~D.
  Lafferty, R.~L. Mercer, and P.~S. Roossin.
\newblock {A statistical approach to machine translation}.
\newblock \emph{Computational linguistics}, 16\penalty0 (2):\penalty0 79--85,
  1990.

\bibitem[Bryl et~al.(2016)Bryl, Bizer, and Paulheim]{bryl2015gathering}
Bryl, V., C.~Bizer, and H.~Paulheim.
\newblock Gathering alternative surface forms for DBpedia entities.
\newblock In \emph{NLP \& DBpedia 2015}, volume 1581, pages 13--24, Aachen,
  2016. RWTH.

\bibitem[Callaway and Lester(2002)]{callaway2002}
Callaway, C.~B. and J.~C. Lester.
\newblock Pronominalization in Generated Discourse and Dialogue.
\newblock In \emph{Proceedings of the 40th Annual Meeting on Association for
  Computational Linguistics}, ACL'02, pages 88--95, Philadelphia, Pennsylvania,
  2002. Association for Computational Linguistics.
\newblock \doi{10.3115/1073083.1073100}.

\bibitem[Cao et~al.(2018)Cao, Wang, and de~Melo]{cao2018link}
Cao, Z., L.~Wang, and G.~de~Melo.
\newblock Link Prediction via Subgraph Embedding-Based Convex Matrix
  Completion.
\newblock In \emph{Proceedings of the 32nd AAAI Conference on Artificial
  Intelligence}, 2018.

\bibitem[Carpuat and Wu(2007)]{carpuat2007improving}
Carpuat, M. and D.~Wu.
\newblock {Improving Statistical Machine Translation Using Word Sense
  Disambiguation.}
\newblock In \emph{EMNLP-CoNLL}, volume~7, pages 61--72, 2007.

\bibitem[Castro~Ferreira et~al.(2016)Castro~Ferreira, Krahmer, and
  Wubben]{ferreira2016b}
Castro~Ferreira, T., E.~Krahmer, and S.~Wubben.
\newblock Towards more variation in text generation: Developing and evaluating
  variation models for choice of referential form.
\newblock In \emph{Proceedings of the 54th Annual Meeting of the Association
  for Computational Linguistics}, ACL'16, pages 568–--577, Berlin, Germany,
  2016. Association for Computational Linguistics.

\bibitem[Castro~Ferreira et~al.(2017)Castro~Ferreira, Krahmer, and
  Wubben]{ferreira2017}
Castro~Ferreira, T., E.~Krahmer, and S.~Wubben.
\newblock Generating flexible proper name references in text: Data, models and
  evaluation.
\newblock In \emph{Proceedings of the 15th Conference of the European Chapter
  of the Association for Computational Linguistics: Volume 1, Long Papers},
  EACL'17, pages 655--664, Valencia, Spain, 2017. Association for Computational
  Linguistics.
\newblock URL \url{http://www.aclweb.org/anthology/E17-1062}.

\bibitem[Chung et~al.(2016)Chung, Cho, and Bengio]{chung2016character}
Chung, J., K.~Cho, and Y.~Bengio.
\newblock A Character-level Decoder without Explicit Segmentation for Neural
  Machine Translation.
\newblock In \emph{Proceedings of the 54th Annual Meeting of the Association
  for Computational Linguistics}, pages 1693--1703. ACL, 2016.

\bibitem[Clark et~al.(2011)Clark, Dyer, Lavie, and Smith]{clarketal2011}
Clark, J.~H., C.~Dyer, A.~Lavie, and N.~A. Smith.
\newblock {Better Hypothesis Testing for Statistical Machine Translation:
  Controlling for Optimizer Instability}.
\newblock In \emph{Proceedings of the 49th Annual Meeting of the Association
  for Computational Linguistics: Human Language Technologies: Short Papers -
  Volume 2}, ACL'11, pages 176--181, Portland, Oregon, 2011.
\newblock ISBN 978-1-932432-88-6.
\newblock URL \url{http://dl.acm.org/citation.cfm?id=2002736.2002774}.

\bibitem[Colin et~al.(2016)Colin, Gardent, Mrabet, Narayan, and
  Perez-Beltrachini]{colin2016webnlg}
Colin, E., C.~Gardent, Y.~Mrabet, S.~Narayan, and L.~Perez-Beltrachini.
\newblock The webnlg challenge: Generating text from dbpedia data.
\newblock In \emph{Proceedings of the 9th International Natural Language
  Generation conference}, pages 163--167, 2016.

\bibitem[Costa-juss{\`a}(2015)]{costa2015much}
Costa-juss{\`a}, M.~R.
\newblock {How much hybridization does machine translation Need?}
\newblock \emph{Journal of the Association for Information Science and
  Technology}, 66\penalty0 (10):\penalty0 2160--2165, 2015.

\bibitem[Costa-Juss{\`a} and Farr{\'u}s(2014)]{costa2014statistical}
Costa-Juss{\`a}, M.~R. and M.~Farr{\'u}s.
\newblock {Statistical machine translation enhancements through linguistic
  levels: A survey}.
\newblock \emph{ACM Computing Surveys (CSUR)}, 46\penalty0 (3):\penalty0 42,
  2014.

\bibitem[Costa-Jussa and Fonollosa(2015)]{costa2015latest}
Costa-Jussa, M.~R. and J.~A. Fonollosa.
\newblock {Latest trends in hybrid machine translation and its applications}.
\newblock \emph{Computer Speech \& Language}, 32\penalty0 (1):\penalty0 3--10,
  2015.

\bibitem[Costa-Jussa et~al.(2012)Costa-Jussa, Farr{\'u}s, Mari{\~n}o, and
  Fonollosa]{costa2012study}
Costa-Jussa, M.~R., M.~Farr{\'u}s, J.~B. Mari{\~n}o, and J.~A. Fonollosa.
\newblock {Study and comparison of rule-based and statistical Catalan-Spanish
  machine translation systems}.
\newblock \emph{Computing and Informatics}, 31\penalty0 (2):\penalty0 245--270,
  2012.

\bibitem[Dale and Haddock(1991)]{dale1991}
Dale, R. and N.~Haddock.
\newblock Generating referring expressions involving relations.
\newblock In \emph{Proceedings of the fifth conference on European chapter of
  the Association for Computational Linguistics}, EACL'91, pages 161--166,
  Berlin, Germany, 1991. Association for Computational Linguistics.
\newblock \doi{10.3115/977180.977208}.

\bibitem[Dale and Reiter(1995)]{dale1995computational}
Dale, R. and E.~Reiter.
\newblock Computational interpretations of the Gricean maxims in the generation
  of referring expressions.
\newblock \emph{Cognitive science}, 19\penalty0 (2):\penalty0 233--263, 1995.

\bibitem[De~Oliveira and Sripada(2014)]{de2014adapting}
De~Oliveira, R. and S.~Sripada.
\newblock Adapting SimpleNLG for Brazilian Portuguese realisation.
\newblock In \emph{INLG}, pages 93--94, 2014.

\bibitem[Devi et~al.(2014)Devi, Gupta, and Dixit]{devi2014comparative}
Devi, P., A.~Gupta, and A.~Dixit.
\newblock Comparative Study of HITS and PageRank Link based Ranking Algorithms.
\newblock \emph{International Journal of Advanced Research in Computer and
  Communication Engineering}, 3\penalty0 (2):\penalty0 5749--5754, 2014.

\bibitem[Edunov et~al.(2018)Edunov, Ott, Auli, and
  Grangier]{edunov2018understanding}
Edunov, S., M.~Ott, M.~Auli, and D.~Grangier.
\newblock Understanding Back-Translation at Scale.
\newblock \emph{arXiv preprint arXiv:1808.09381}, 2018.

\bibitem[Ell et~al.(2011)Ell, Vrandecic, and Simperl]{ell2011}
Ell, B., D.~Vrandecic, and E.~P.~B. Simperl.
\newblock Labels in the Web of Data.
\newblock In \emph{Proceedings of ISWC}, volume 7031, pages 162--176. Springer,
  2011.

\bibitem[Ferreira et~al.(2016)Ferreira, Krahmer, and
  Wubben]{ferreira2016towards}
Ferreira, T.~C., E.~Krahmer, and S.~Wubben.
\newblock Towards more variation in text generation: Developing and evaluating
  variation models for choice of referential form.
\newblock In \emph{ACL (1)}, 2016.

\bibitem[Ferreira et~al.(2017)Ferreira, Krahmer, and
  Wubben]{ferreira2017generating}
Ferreira, T.~C., E.~Krahmer, and S.~Wubben.
\newblock Generating flexible proper name references in text: Data, models and
  evaluation.
\newblock In \emph{Proc. EACL}, volume~17, 2017.

\bibitem[Ferreira et~al.(2018{\natexlab{a}})Ferreira, Moussallem, Krahmer, and
  Wubben]{moussallem2018enriching}
Ferreira, T.~C., D.~Moussallem, E.~Krahmer, and S.~Wubben.
\newblock Enriching the WebNLG corpus.
\newblock In \emph{Proceedings of the 11th International Conference on Natural
  Language Generation}, pages 171--176, 2018{\natexlab{a}}.

\bibitem[Ferreira et~al.(2018{\natexlab{b}})Ferreira, Moussallem, Ákos
  Kádár, Wubben, and Krahmer]{moussallem2018neuralreg}
Ferreira, T.~C., D.~Moussallem, Ákos Kádár, S.~Wubben, and E.~Krahmer.
\newblock {NeuralREG: An end-to-end approach to referring expression
  generation}.
\newblock In \emph{Proceedings of the 55th Annual Meeting of the Association
  for Computational Linguistics (Volume 1: Long Papers)}, 2018{\natexlab{b}}.

\bibitem[Friedman(1937)]{friedman1937use}
Friedman, M.
\newblock The use of ranks to avoid the assumption of normality implicit in the
  analysis of variance.
\newblock \emph{Journal of the american statistical association}, 32\penalty0
  (200):\penalty0 675--701, 1937.

\bibitem[Ganea et~al.(2016)Ganea, Ganea, Lucchi, Eickhoff, and Hofmann]{PBOH}
Ganea, O.-E., M.~Ganea, A.~Lucchi, C.~Eickhoff, and T.~Hofmann.
\newblock Probabilistic Bag-Of-Hyperlinks Model for Entity Linking.
\newblock In \emph{Proceedings of the 25th International Conference on World
  Wide Web}, WWW '16, pages 927--938, Republic and Canton of Geneva,
  Switzerland, 2016. International World Wide Web Conferences Steering
  Committee.
\newblock ISBN 978-1-4503-4143-1.
\newblock \doi{10.1145/2872427.2882988}.
\newblock URL \url{http://dx.doi.org/10.1145/2872427.2882988}.

\bibitem[Gardent et~al.(2017{\natexlab{a}})Gardent, Shimorina, Narayan, and
  Perez-Beltrachini]{claire2017}
Gardent, C., A.~Shimorina, S.~Narayan, and L.~Perez-Beltrachini.
\newblock Creating Training Corpora for {NLG} Micro-Planners.
\newblock In \emph{Proceedings of the 55th Annual Meeting of the Association
  for Computational Linguistics (Volume 1: Long Papers)}, ACL'17, pages
  179--188, Vancouver, Canada, 2017{\natexlab{a}}. Association for
  Computational Linguistics.
\newblock \doi{10.18653/v1/P17-1017}.
\newblock URL \url{http://www.aclweb.org/anthology/P17-1017}.

\bibitem[Gardent et~al.(2017{\natexlab{b}})Gardent, Shimorina, Narayan, and
  Perez-Beltrachini]{claire2017b}
Gardent, C., A.~Shimorina, S.~Narayan, and L.~Perez-Beltrachini.
\newblock The {WebNLG} Challenge: Generating Text from {RDF} Data.
\newblock In \emph{Proceedings of the 10th International Conference on Natural
  Language Generation}, INLG'17, pages 124--133, Santiago de Compostela, Spain,
  2017{\natexlab{b}}. Association for Computational Linguistics.
\newblock URL \url{http://aclweb.org/anthology/W17-3518}.

\bibitem[Gardent et~al.(2017{\natexlab{c}})Gardent, Shimorina, Narayan, and
  Perez-Beltrachini]{gardent2017creating}
Gardent, C., A.~Shimorina, S.~Narayan, and L.~Perez-Beltrachini.
\newblock Creating training corpora for nlg micro-planning.
\newblock In \emph{Proceedings of ACL}, 2017{\natexlab{c}}.

\bibitem[Gatt and Krahmer(2018)]{gatt2017}
Gatt, A. and E.~Krahmer.
\newblock Survey of the State of the Art in Natural Language Generation: Core
  tasks, applications and evaluation.
\newblock \emph{Journal of Artificial Intelligence Research}, 61:\penalty0
  65--170, 2018.

\bibitem[Gerber et~al.(2015)Gerber, Esteves, Lehmann, B{\"u}hmann, Usbeck,
  Ngomo, and Speck]{gerber2015defacto}
Gerber, D., D.~Esteves, J.~Lehmann, L.~B{\"u}hmann, R.~Usbeck, A.-C.~N. Ngomo,
  and R.~Speck.
\newblock Defacto—temporal and multilingual deep fact validation.
\newblock \emph{Web Semantics: Science, Services and Agents on the World Wide
  Web}, 35:\penalty0 85--101, 2015.

\bibitem[Glorot and Bengio(2010)]{glorot2011}
Glorot, X. and Y.~Bengio.
\newblock Understanding the difficulty of training deep feedforward neural
  networks.
\newblock In \emph{Proceedings of the Thirteenth International Conference on
  Artificial Intelligence and Statistics}, volume~9 of \emph{Proceedings of
  Machine Learning Research}, pages 249--256, Chia Laguna Resort, Sardinia,
  Italy, 13--15 May 2010. PMLR.
\newblock URL \url{http://proceedings.mlr.press/v9/glorot10a.html}.

\bibitem[Grave et~al.(2018)Grave, Bojanowski, Gupta, Joulin, and
  Mikolov]{grave2018learning}
Grave, E., P.~Bojanowski, P.~Gupta, A.~Joulin, and T.~Mikolov.
\newblock Learning Word Vectors for 157 Languages.
\newblock In \emph{Proceedings of the International Conference on Language
  Resources and Evaluation (LREC)}, 2018.

\bibitem[Harriehausen-M{\"u}hlbauer and Heuss(2012)]{harriehausen2012semantic}
Harriehausen-M{\"u}hlbauer, B. and T.~Heuss.
\newblock {Semantic web based machine translation}.
\newblock In \emph{Proceedings of the Joint Workshop on Exploiting Synergies
  between Information Retrieval and Machine Translation (ESIRMT) and Hybrid
  Approaches to Machine Translation (HyTra)}, pages 1--9. Association for
  Computational Linguistics, 2012.

\bibitem[Henschel et~al.(2000)Henschel, Cheng, and Poesio]{henschel2000}
Henschel, R., H.~Cheng, and M.~Poesio.
\newblock Pronominalization Revisited.
\newblock In \emph{Proceedings of the 18th Conference on Computational
  Linguistics - Volume 1}, COLING'00, pages 306--312, Saarbr{\"{u}}cken,
  Germany, 2000. Association for Computational Linguistics.
\newblock ISBN 1-55860-717-X.
\newblock \doi{10.3115/990820.990865}.
\newblock URL \url{https://doi.org/10.3115/990820.990865}.

\bibitem[Heuss(2013)]{heuss2013lessons}
Heuss, T.
\newblock {Lessons learned (and questions raised) from an interdisciplinary
  Machine Translation approach}.
\newblock In \emph{Position paper for the W3C Workshop on the Open Data on the
  Web}, pages 23--24, 2013.

\bibitem[Hochreiter and Schmidhuber(1997)]{hochreiter1997long}
Hochreiter, S. and J.~Schmidhuber.
\newblock Long short-term memory.
\newblock \emph{Neural computation}, 9\penalty0 (8):\penalty0 1735--1780, 1997.

\bibitem[Hoffart et~al.(2011)Hoffart, Yosef, Bordino, F{\"u}rstenau, Pinkal,
  Spaniol, Taneva, Thater, and Weikum]{AIDA}
Hoffart, J., M.~A. Yosef, I.~Bordino, H.~F{\"u}rstenau, M.~Pinkal, M.~Spaniol,
  B.~Taneva, S.~Thater, and G.~Weikum.
\newblock {Robust Disambiguation of Named Entities in Text}.
\newblock In \emph{Conference on Empirical Methods in Natural Language
  Processing}, 2011.

\bibitem[Hoffart et~al.(2014)Hoffart, Altun, and
  Weikum]{Hoffart:2014:DEE:2566486.2568003}
Hoffart, J., Y.~Altun, and G.~Weikum.
\newblock Discovering Emerging Entities with Ambiguous Names.
\newblock In \emph{Proceedings of the 23rd International Conference on World
  Wide Web}, WWW '14, pages 385--396, New York, NY, USA, 2014. ACM.

\bibitem[Hutchins and Somers(1992)]{hutchins1992introduction}
Hutchins, W.~J. and H.~L. Somers.
\newblock \emph{{An introduction to machine translation}}, volume 362.
\newblock Academic Press London, 1992.

\bibitem[Joulin et~al.(2017)Joulin, Grave, Bojanowski, Nickel, and
  Mikolov]{joulin2017fast}
Joulin, A., E.~Grave, P.~Bojanowski, M.~Nickel, and T.~Mikolov.
\newblock Fast Linear Model for Knowledge Graph Embeddings.
\newblock \emph{arXiv preprint arXiv:1710.10881}, 2017.

\bibitem[Jurafsky(2000)]{Jurafsky2000}
Jurafsky, D.
\newblock \emph{{Speech and language processing: An introduction to natural
  language processing}}.
\newblock Prentice Hall, 2000.

\bibitem[K~M et~al.(2018)K~M, Basu Roy~Chowdhury, and Dukkipati]{annervaz2018}
K~M, A., S.~Basu Roy~Chowdhury, and A.~Dukkipati.
\newblock Learning beyond Datasets: Knowledge Graph Augmented Neural Networks
  for Natural Language Processing.
\newblock In \emph{Proceedings of the 2018 Conference of the North American
  Chapter of the Association for Computational Linguistics: Human Language
  Technologies, Volume 1 (Long Papers)}, pages 313--322. Association for
  Computational Linguistics, 2018.
\newblock URL \url{http://aclweb.org/anthology/N18-1029}.

\bibitem[Keet and Khumalo(2017)]{keet2017toward}
Keet, C.~M. and L.~Khumalo.
\newblock Toward a knowledge-to-text controlled natural language of isiZulu.
\newblock \emph{Language Resources and Evaluation}, 51\penalty0 (1):\penalty0
  131--157, 2017.

\bibitem[{Klein} et~al.(2017){Klein}, {Kim}, {Deng}, {Senellart}, and
  {Rush}]{2017opennmt}
{Klein}, G., Y.~{Kim}, Y.~{Deng}, J.~{Senellart}, and A.~M. {Rush}.
\newblock {OpenNMT: Open-Source Toolkit for Neural Machine Translation}.
\newblock \emph{ArXiv e-prints}, 2017.

\bibitem[Kleinberg(1999)]{HITS}
Kleinberg, J.~M.
\newblock Authoritative sources in a hyperlinked environment.
\newblock \emph{J. ACM}, 46\penalty0 (5):\penalty0 604--632, 1999.

\bibitem[Koehn(2005)]{koehn2005europarl}
Koehn, P.
\newblock Europarl: A parallel corpus for statistical machine translation.
\newblock In \emph{MT summit}, volume~5, pages 79--86, 2005.

\bibitem[Koehn(2010)]{Koehn2010}
Koehn, P.
\newblock \emph{{Statistical Machine Translation}}.
\newblock Cambridge University Press, 2010.

\bibitem[Koehn and Knowles(2017)]{koehn2017six}
Koehn, P. and R.~Knowles.
\newblock Six Challenges for Neural Machine Translation.
\newblock In \emph{Proceedings of the First Workshop on Neural Machine
  Translation}, pages 28--39, 2017.

\bibitem[Koehn et~al.(2007)Koehn, Hoang, Birch, Callison-Burch, Federico,
  Bertoldi, Cowan, Shen, Moran, Zens, et~al.]{koehn2007moses}
Koehn, P., H.~Hoang, A.~Birch, C.~Callison-Burch, M.~Federico, N.~Bertoldi,
  B.~Cowan, W.~Shen, C.~Moran, R.~Zens, et~al.
\newblock {Moses: Open source toolkit for statistical machine translation}.
\newblock In \emph{Proceedings of the 45th annual meeting of the ACL}, pages
  177--180. Association for Computational Linguistics, 2007.

\bibitem[Krahmer and Van~Deemter(2012{\natexlab{a}})]{krahmer2012}
Krahmer, E. and K.~Van~Deemter.
\newblock Computational generation of referring expressions: A survey.
\newblock \emph{Computational Linguistics}, 38\penalty0 (1):\penalty0 173--218,
  2012{\natexlab{a}}.

\bibitem[Krahmer and Van~Deemter(2012{\natexlab{b}})]{krahmer2012computational}
Krahmer, E. and K.~Van~Deemter.
\newblock Computational generation of referring expressions: A survey.
\newblock \emph{Computational Linguistics}, 38\penalty0 (1):\penalty0 173--218,
  2012{\natexlab{b}}.

\bibitem[Lakshen et~al.(2018)Lakshen, Janev, and
  Vrane{\v{s}}]{lakshen2018challenges}
Lakshen, G.~A., V.~Janev, and S.~Vrane{\v{s}}.
\newblock Challenges in Quality Assessment of Arabic DBpedia.
\newblock In \emph{Proceedings of the 8th International Conference on Web
  Intelligence, Mining and Semantics}, page~15. ACM, 2018.

\bibitem[{Levenshtein}(1966)]{levenshtein1966}
{Levenshtein}, V.~I.
\newblock {Binary Codes Capable of Correcting Deletions, Insertions and
  Reversals}.
\newblock \emph{Soviet Physics Doklady}, 10:\penalty0 707, February 1966.

\bibitem[Li et~al.(2018)Li, Wang, Aw, Chng, and Li]{li2018named}
Li, Z., X.~Wang, A.~Aw, E.~S. Chng, and H.~Li.
\newblock Named-Entity Tagging and Domain adaptation for Better Customized
  Translation.
\newblock In \emph{Proceedings of the Seventh Named Entities Workshop}, pages
  41--46. ACL, 2018.

\bibitem[Libovick{\'y} and Helcl(2017)]{libovicky2017attention}
Libovick{\'y}, J. and J.~Helcl.
\newblock Attention Strategies for Multi-Source Sequence-to-Sequence Learning.
\newblock In \emph{Proceedings of the 55th Annual Meeting of the Association
  for Computational Linguistics (Volume 2: Short Papers)}, ACL'17, pages
  196--202, Vancouver, Canada, 2017. Association for Computational Linguistics.
\newblock \doi{10.18653/v1/P17-2031}.
\newblock URL \url{http://www.aclweb.org/anthology/P17-2031}.

\bibitem[Lopez and Post(2013)]{lopez2013beyond}
Lopez, A. and M.~Post.
\newblock {Beyond bitext: Five open problems in machine translation}.
\newblock In \emph{Proceedings of the EMNLP Workshop on Twenty Years of
  Bitext}, pages 1--3, 2013.

\bibitem[Luong and Manning(2015)]{luong2015stanford}
Luong, M.-T. and C.~D. Manning.
\newblock Stanford neural machine translation systems for spoken language
  domains.
\newblock In \emph{Proceedings of the International Workshop on Spoken Language
  Translation}, pages 76--79, 2015.

\bibitem[Luong and Manning(2016)]{luong2016achieving}
Luong, M.-T. and C.~D. Manning.
\newblock Achieving Open Vocabulary Neural Machine Translation with Hybrid
  Word-Character Models.
\newblock In \emph{Proceedings of the 54th Annual Meeting of the Association
  for Computational Linguistics}, pages 1054--1063. ACL, 2016.

\bibitem[Moussallem and Choren(2015)]{moussallem2015using}
Moussallem, D. and R.~Choren.
\newblock {Using Ontology-Based Context in the Portuguese-English Translation
  of Homographs in Textual Dialogues}.
\newblock \emph{Artificial Intelligence and Applications}, 1510, 2015.

\bibitem[Moussallem et~al.(2017)Moussallem, Usbeck, R{\"o}eder, and
  Ngomo]{moussallem2017mag}
Moussallem, D., R.~Usbeck, M.~R{\"o}eder, and A.-C.~N. Ngomo.
\newblock {MAG: A Multilingual, Knowledge-base Agnostic and Deterministic
  Entity Linking Approach}.
\newblock In \emph{Proceedings of the Knowledge Capture Conference}, page~9.
  ACM, 2017.

\bibitem[Moussallem et~al.(2018{\natexlab{a}})Moussallem, Ferreira, Zampieri,
  Cavalcanti, Xexéo, Neves, and Ngomo]{rdf2pt_lrec_2018}
Moussallem, D., T.~C. Ferreira, M.~Zampieri, M.~C. Cavalcanti, G.~Xexéo,
  M.~Neves, and A.-C.~N. Ngomo.
\newblock {RDF2PT: Generating Brazilian Portuguese Texts from RDF Data}.
\newblock In \emph{The 11th edition of the Language Resources and Evaluation
  Conference, 7-12 May 2018, Miyazaki (Japan)}, 2018{\natexlab{a}}.
\newblock URL \url{https://arxiv.org/abs/1802.08150}.

\bibitem[Moussallem et~al.(2018{\natexlab{b}})Moussallem, Sherif, Esteves,
  Zampieri, and Ngomo]{moussallemlrec2018}
Moussallem, D., M.~A. Sherif, D.~Esteves, M.~Zampieri, and A.-C.~N. Ngomo.
\newblock {LI}dioms: A {M}ultilingual {L}inked {I}dioms {D}ata {S}et.
\newblock In \emph{LREC 2018}, page~7, 2018{\natexlab{b}}.

\bibitem[Moussallem et~al.(2018{\natexlab{c}})Moussallem, Usbeck, Röder, and
  Ngomo]{moussallem2018entity}
Moussallem, D., R.~Usbeck, M.~Röder, and A.-C.~N. Ngomo.
\newblock {Entity Linking in 40 Languages using MAG}.
\newblock In \emph{The Semantic Web, ESWC 2018, Lecture Notes in Computer
  Science}, 2018{\natexlab{c}}.

\bibitem[Moussallem et~al.(2018{\natexlab{d}})Moussallem, Wauer, and
  Ngomo]{moussallem2018machine}
Moussallem, D., M.~Wauer, and A.-C.~N. Ngomo.
\newblock {Machine Translation Using Semantic Web Technologies: A Survey}.
\newblock \emph{Journal of Web Semantics}, 51:\penalty0 1--19,
  2018{\natexlab{d}}.

\bibitem[Moussallem et~al.(2019{\natexlab{a}})Moussallem, Ngomo, Buitelaar, and
  Arcan]{moussallem2019augmenting}
Moussallem, D., A.-C.~N. Ngomo, P.~Buitelaar, and M.~Arcan.
\newblock {Utilizing Knowledge Graphs for Neural Machine Translation
  Augmentation}.
\newblock In \emph{Proceedings of the 10th International Conference on
  Knowledge Capture}, pages 139--146. ACM, 2019{\natexlab{a}}.

\bibitem[Moussallem et~al.(2019{\natexlab{b}})Moussallem, Soru, and
  Ngomo]{moussallem2019thoth}
Moussallem, D., T.~Soru, and A.-C.~N. Ngomo.
\newblock {THOTH: Neural Translation and Enrichment of Knowledge Graphs}.
\newblock In \emph{The Semantic Web ISWC 2019}, pages 1--17. Springer,
  2019{\natexlab{b}}.

\bibitem[Navigli(2009)]{navigli2009word}
Navigli, R.
\newblock {Word sense disambiguation: A survey}.
\newblock \emph{ACM Computing Surveys (CSUR)}, 41\penalty0 (2):\penalty0 10,
  2009.

\bibitem[Neishi et~al.(2017)Neishi, Sakuma, Tohda, Ishiwatari, Yoshinaga, and
  Toyoda]{neishi2017bag}
Neishi, M., J.~Sakuma, S.~Tohda, S.~Ishiwatari, N.~Yoshinaga, and M.~Toyoda.
\newblock A bag of useful tricks for practical neural machine translation:
  Embedding layer initialization and large batch size.
\newblock In \emph{Proceedings of the 4th Workshop on Asian Translation}, pages
  99--109, 2017.

\bibitem[{Neubig} et~al.(2017){Neubig}, {Dyer}, {Goldberg}, {Matthews},
  {Ammar}, {Anastasopoulos}, {Ballesteros}, {Chiang}, {Clothiaux}, {Cohn},
  {Duh}, {Faruqui}, {Gan}, {Garrette}, {Ji}, {Kong}, {Kuncoro}, {Kumar},
  {Malaviya}, {Michel}, {Oda}, {Richardson}, {Saphra}, {Swayamdipta}, and
  {Yin}]{neubig2017}
{Neubig}, G., C.~{Dyer}, Y.~{Goldberg}, A.~{Matthews}, W.~{Ammar},
  A.~{Anastasopoulos}, M.~{Ballesteros}, D.~{Chiang}, D.~{Clothiaux},
  T.~{Cohn}, K.~{Duh}, M.~{Faruqui}, C.~{Gan}, D.~{Garrette}, Y.~{Ji},
  L.~{Kong}, A.~{Kuncoro}, G.~{Kumar}, C.~{Malaviya}, P.~{Michel}, Y.~{Oda},
  M.~{Richardson}, N.~{Saphra}, S.~{Swayamdipta}, and P.~{Yin}.
\newblock {DyNet: The Dynamic Neural Network Toolkit}.
\newblock \emph{ArXiv e-prints}, January 2017.

\bibitem[Ngonga~Ngomo et~al.(2013)Ngonga~Ngomo, B{\"u}hmann, Unger, Lehmann,
  and Gerber]{ngonga2013sorry}
Ngonga~Ngomo, A.-C., L.~B{\"u}hmann, C.~Unger, J.~Lehmann, and D.~Gerber.
\newblock Sorry, i don't speak SPARQL: translating SPARQL queries into natural
  language.
\newblock In \emph{Proceedings of the 22nd international conference on World
  Wide Web}, pages 977--988. ACM, 2013.

\bibitem[Ngonga~Ngomo et~al.(2018)Ngonga~Ngomo, R{\"o}der, Moussallem, Usbeck,
  and Speck]{bengal}
Ngonga~Ngomo, A.-C., M.~R{\"o}der, D.~Moussallem, R.~Usbeck, and R.~Speck.
\newblock BENGAL: An Automatic Benchmark Generator for Entity Recognition and
  Linking.
\newblock In \emph{Proceedings of the 11th International Conference on Natural
  Language Generation}, pages 339--349, 2018.

\bibitem[{Ngonga Ngomo} et~al.(2019){Ngonga Ngomo}, Moussallem, and
  Bühman]{Ngonga2019}
{Ngonga Ngomo}, A.-C., D.~Moussallem, and L.~Bühman.
\newblock {A Holistic Natural Language Generation Framework for the Semantic
  Web}.
\newblock In \emph{Proceedings of the International Conference Recent Advances
  in Natural Language Processing}, page~8. ACL (Association for Computational
  Linguistics), 2019.

\bibitem[Orita et~al.(2015)Orita, Vornov, Feldman, and
  Daum{\'e}~III]{orita2015}
Orita, N., E.~Vornov, N.~Feldman, and H.~Daum{\'e}~III.
\newblock Why discourse affects speakers' choice of referring expressions.
\newblock In \emph{Proceedings of the 53rd Annual Meeting of the Association
  for Computational Linguistics and the 7th International Joint Conference on
  Natural Language Processing (Volume 1: Long Papers)}, ACL'15, pages
  1639--1649, Beijing, China, 2015. Association for Computational Linguistics.
\newblock \doi{10.3115/v1/P15-1158}.
\newblock URL \url{http://www.aclweb.org/anthology/P15-1158}.

\bibitem[Page et~al.(1999)Page, Brin, Motwani, and Winograd]{page1999pagerank}
Page, L., S.~Brin, R.~Motwani, and T.~Winograd.
\newblock The PageRank citation ranking: Bringing order to the web.
\newblock Technical report, Stanford InfoLab, 1999.

\bibitem[Papineni et~al.(2002{\natexlab{a}})Papineni, Roukos, Ward, and
  Zhu]{papineni2002}
Papineni, K., S.~Roukos, T.~Ward, and W.-J. Zhu.
\newblock Bleu: a Method for Automatic Evaluation of Machine Translation.
\newblock In \emph{Proceedings of 40th Annual Meeting of the Association for
  Computational Linguistics}, ACL'02, pages 311--318, Philadelphia,
  Pennsylvania, USA, July 2002{\natexlab{a}}. Association for Computational
  Linguistics.
\newblock \doi{10.3115/1073083.1073135}.
\newblock URL \url{http://www.aclweb.org/anthology/P02-1040}.

\bibitem[Papineni et~al.(2002{\natexlab{b}})Papineni, Roukos, Ward, and
  Zhu]{papineni2002bleu}
Papineni, K., S.~Roukos, T.~Ward, and W.-J. Zhu.
\newblock {BLEU: a method for automatic evaluation of machine translation}.
\newblock In \emph{Proceedings of the 40th annual meeting on association for
  computational linguistics}, pages 311--318. Association for Computational
  Linguistics, 2002{\natexlab{b}}.

\bibitem[Popovi{\'c}(2017)]{popovic2017chrf++}
Popovi{\'c}, M.
\newblock {chrF++: words helping character n-grams}.
\newblock In \emph{Proceedings of the Second Conference on Machine
  Translation}, pages 612--618, 2017.

\bibitem[Ramos et~al.(2003)]{ramos2003using}
Ramos, J. et~al.
\newblock Using tf-idf to determine word relevance in document queries.
\newblock In \emph{Proceedings of the first instructional conference on machine
  learning}, 2003.

\bibitem[Reiter and Dale(2000)]{reiter2000}
Reiter, E. and R.~Dale.
\newblock \emph{Building natural language generation systems}.
\newblock Cambridge University Press, New York, NY, USA, 2000.
\newblock ISBN 0-521-62036-8.

\bibitem[R{\"o}der et~al.(2018)R{\"o}der, Usbeck, and Ngomo]{roder2017gerbil}
R{\"o}der, M., R.~Usbeck, and A.-C.~N. Ngomo.
\newblock {GERBIL--Benchmarking Named Entity Recognition and Linking
  Consistently}.
\newblock \emph{Semantic Web Journal}, 2018.
\newblock URL
  \url{http://www.semantic-web-journal.net/system/files/swj1671.pdf}.

\bibitem[Schuster and Paliwal(1997)]{schuster1997bidirectional}
Schuster, M. and K.~K. Paliwal.
\newblock Bidirectional recurrent neural networks.
\newblock \emph{IEEE Transactions on Signal Processing}, 45\penalty0
  (11):\penalty0 2673--2681, 1997.

\bibitem[Sennrich et~al.(2016{\natexlab{a}})Sennrich, Haddow, and
  Birch]{sennrich2015neural}
Sennrich, R., B.~Haddow, and A.~Birch.
\newblock Neural Machine Translation of Rare Words with Subword Units.
\newblock In \emph{Proceedings of the 54th Annual Meeting of the Association
  for Computational Linguistics}, pages 1715--1725. ACL, 2016{\natexlab{a}}.

\bibitem[Sennrich et~al.(2016{\natexlab{b}})Sennrich, Haddow, and
  Birch]{sennrich2016improving}
Sennrich, R., B.~Haddow, and A.~Birch.
\newblock Improving Neural Machine Translation Models with Monolingual Data.
\newblock In \emph{Proceedings of the 54th Annual Meeting of the Association
  for Computational Linguistics (Volume 1: Long Papers)}, volume~1, pages
  86--96, 2016{\natexlab{b}}.

\bibitem[Seo et~al.(2009)Seo, Song, Kim, and Choi]{seo2009syntactic}
Seo, E., I.-S. Song, S.-K. Kim, and H.-J. Choi.
\newblock {Syntactic and semantic English-Korean machine translation using
  ontology}.
\newblock In \emph{Advanced Communication Technology, 2009. ICACT 2009. 11th
  International Conference on}, volume~3, pages 2129--2132. IEEE, 2009.

\bibitem[Siddharthan et~al.(2011)Siddharthan, Nenkova, and
  McKeown]{siddharthan2011}
Siddharthan, A., A.~Nenkova, and K.~McKeown.
\newblock Information Status Distinctions and Referring Expressions: An
  Empirical Study of References to People in News Summaries.
\newblock \emph{Computational Linguistics}, 37\penalty0 (4):\penalty0 811--842,
  2011.
\newblock \doi{10.1162/COLI_a_00077}.
\newblock URL \url{http://dx.doi.org/10.1162/COLI_a_00077}.

\bibitem[Slocum(1985)]{slocum1985survey}
Slocum, J.
\newblock {A survey of machine translation: its history, current status, and
  future prospects}.
\newblock \emph{Computational linguistics}, 11\penalty0 (1):\penalty0 1--17,
  1985.

\bibitem[Sorokin and Gurevych(2018)]{Sorokincoling2018}
Sorokin, D. and I.~Gurevych.
\newblock Modeling Semantics with Gated Graph Neural Networks for Knowledge
  Base Question Answering.
\newblock In \emph{Proceedings of the 27th International Conference on
  Computational Linguistics}, pages 3306--3317. ACL, 2018.

\bibitem[Stahlberg(2019)]{stahlberg2019neural}
Stahlberg, F.
\newblock Neural Machine Translation: A Review, 2019.

\bibitem[Steinberger et~al.(2006)Steinberger, Pouliquen, Widiger, Ignat,
  Erjavec, Tufis, and Varga]{steinberger2006jrc}
Steinberger, R., B.~Pouliquen, A.~Widiger, C.~Ignat, T.~Erjavec, D.~Tufis, and
  D.~Varga.
\newblock The JRC-Acquis: A multilingual aligned parallel corpus with 20+
  languages.
\newblock \emph{arXiv preprint cs/0609058}, 2006.

\bibitem[Sutskever et~al.(2014)Sutskever, Vinyals, and
  Le]{sutskever2014sequence}
Sutskever, I., O.~Vinyals, and Q.~V. Le.
\newblock {Sequence to sequence learning with neural networks}.
\newblock In \emph{Advances in neural information processing systems}, pages
  3104--3112, 2014.

\bibitem[Thurmair(2004)]{thurmair2004comparing}
Thurmair, G.
\newblock {Comparing rule-based and statistical MT output}.
\newblock In \emph{The Workshop Programme}, page~5, 2004.

\bibitem[Thurmair(2009)]{Thumair}
Thurmair, G.
\newblock {Comparing different architectures of hybrid Machine Translation
  systems}.
\newblock \emph{MT Summit XII: proceedings of the twelfth Machine Translation
  Summit}, pages 340--347, 2009.

\bibitem[Tiedemann(2012)]{TIEDEMANN12.463}
Tiedemann, J.
\newblock Parallel Data, Tools and Interfaces in OPUS.
\newblock In Chair), N. C.~C., K.~Choukri, T.~Declerck, M.~U. Dogan,
  B.~Maegaard, J.~Mariani, J.~Odijk, and S.~Piperidis, editors,
  \emph{Proceedings of the Eight International Conference on Language Resources
  and Evaluation (LREC)}, Istanbul, Turkey, may 2012. European Language
  Resources Association (ELRA).
\newblock ISBN 978-2-9517408-7-7.

\bibitem[Toutanova and Manning(2000)]{toutanova2000enriching}
Toutanova, K. and C.~D. Manning.
\newblock Enriching the knowledge sources used in a maximum entropy
  part-of-speech tagger.
\newblock In \emph{Proceedings of the 2000 Joint SIGDAT conference on Empirical
  methods in natural language processing and very large corpora: held in
  conjunction with the 38th Annual Meeting of the Association for Computational
  Linguistics-Volume 13}, pages 63--70. Association for Computational
  Linguistics, 2000.

\bibitem[Ugawa et~al.(2018)Ugawa, Tamura, Ninomiya, Takamura, and
  Okumura]{ugawa2018neural}
Ugawa, A., A.~Tamura, T.~Ninomiya, H.~Takamura, and M.~Okumura.
\newblock Neural Machine Translation Incorporating Named Entity.
\newblock In \emph{Proceedings of the 27th International Conference on
  Computational Linguistics}, pages 3240--3250, 2018.

\bibitem[Usbeck et~al.(2014)Usbeck, Ngomo, R{\"{o}}der, Gerber, Coelho, Auer,
  and Both]{AGDISTIS_ISWC}
Usbeck, R., A.~N. Ngomo, M.~R{\"{o}}der, D.~Gerber, S.~A. Coelho, S.~Auer, and
  A.~Both.
\newblock {AGDISTIS} - Graph-Based Disambiguation of Named Entities Using
  Linked Data.
\newblock In \emph{The Semantic Web - {ISWC} 2014 - 13th International Semantic
  Web Conference, October 19-23, 2014. Proceedings, Part {I}}, pages 457--471,
  Riva del Garda, Italy, 2014.

\bibitem[Usbeck et~al.(2015)Usbeck, R{\"{o}}der, Ngomo, Baron, Both,
  Br{\"{u}}mmer, Ceccarelli, Cornolti, Cherix, Eickmann, Ferragina, Lemke,
  Moro, Navigli, Piccinno, Rizzo, Sack, Speck, Troncy, Waitelonis, and
  Wesemann]{gerbil}
Usbeck, R., M.~R{\"{o}}der, A.~N. Ngomo, C.~Baron, A.~Both, M.~Br{\"{u}}mmer,
  D.~Ceccarelli, M.~Cornolti, D.~Cherix, B.~Eickmann, P.~Ferragina, C.~Lemke,
  A.~Moro, R.~Navigli, F.~Piccinno, G.~Rizzo, H.~Sack, R.~Speck, R.~Troncy,
  J.~Waitelonis, and L.~Wesemann.
\newblock {GERBIL:} General Entity Annotator Benchmarking Framework.
\newblock In \emph{Proceedings of the 24th International Conference on World
  Wide Web, {WWW}, May 18-22}, pages 1133--1143, Florence, Italy, 2015.

\bibitem[van Deemter(2016)]{deemter2016}
van Deemter, K.
\newblock Designing Algorithms for Referring with Proper Names.
\newblock In \emph{Proceedings of the 9th International Natural Language
  Generation conference}, INLG'16, pages 31--35, Edinburgh, UK, 2016.
  Association for Computational Linguistics.
\newblock \doi{10.18653/v1/W16-6605}.
\newblock URL \url{http://www.aclweb.org/anthology/W16-6605}.

\bibitem[Vaswani et~al.(2017)Vaswani, Shazeer, Parmar, Uszkoreit, Jones, Gomez,
  Kaiser, and Polosukhin]{vaswani2017attention}
Vaswani, A., N.~Shazeer, N.~Parmar, J.~Uszkoreit, L.~Jones, A.~N. Gomez,
  {\L}.~Kaiser, and I.~Polosukhin.
\newblock Attention is all you need.
\newblock In \emph{Advances in Neural Information Processing Systems}, pages
  5998--6008, 2017.

\bibitem[Vrande{\v{c}}i{\'c} and Kr{\"o}tzsch(2014)]{vrandevcic2014wikidata}
Vrande{\v{c}}i{\'c}, D. and M.~Kr{\"o}tzsch.
\newblock Wikidata: a free collaborative knowledgebase.
\newblock \emph{Communications of the ACM}, 57\penalty0 (10):\penalty0 78--85,
  2014.

\bibitem[Waitelonis et~al.(2016)Waitelonis, J\"{u}rges, and
  Sack]{waitelonis2016don}
Waitelonis, J., H.~J\"{u}rges, and H.~Sack.
\newblock Don'T Compare Apples to Oranges: Extending GERBIL for a Fine Grained
  NEL Evaluation.
\newblock In \emph{Proceedings of the 12th International Conference on Semantic
  Systems}, SEMANTiCS 2016, pages 65--72, New York, NY, USA, 2016. ACM.

\bibitem[Wu et~al.(2016)Wu, Schuster, Chen, Le, Norouzi, Macherey, Krikun, Cao,
  Gao, Macherey, et~al.]{wu2016}
Wu, Y., M.~Schuster, Z.~Chen, Q.~V. Le, M.~Norouzi, W.~Macherey, M.~Krikun,
  Y.~Cao, Q.~Gao, K.~Macherey, et~al.
\newblock {Google's Neural Machine Translation System: Bridging the Gap between
  Human and Machine Translation}.
\newblock \emph{arXiv preprint arXiv:1609.08144}, 2016.

\bibitem[Yang and Mitchell(2017)]{yang2017leveraging}
Yang, B. and T.~Mitchell.
\newblock Leveraging knowledge bases in lstms for improving machine reading.
\newblock In \emph{Proceedings of the 55th Annual Meeting of the Association
  for Computational Linguistics}, volume~1, pages 1436--1446, 2017.

\bibitem[Young et~al.(2018)Young, Hazarika, Poria, and
  Cambria]{young2018recent}
Young, T., D.~Hazarika, S.~Poria, and E.~Cambria.
\newblock Recent trends in deep learning based natural language processing.
\newblock \emph{ieee Computational intelligenCe magazine}, 13\penalty0
  (3):\penalty0 55--75, 2018.

\bibitem[Zeiler(2012)]{zeiler2012}
Zeiler, M.~D.
\newblock {ADADELTA}: An Adaptive Learning Rate Method.
\newblock \emph{CoRR}, abs/1212.5701, 2012.
\newblock URL \url{http://arxiv.org/abs/1212.5701}.

\bibitem[Zwicklbauer et~al.(2016)Zwicklbauer, Seifert, and Granitzer]{doser}
Zwicklbauer, S., C.~Seifert, and M.~Granitzer.
\newblock DoSeR - A Knowledge-Base-Agnostic Framework for Entity Disambiguation
  Using Semantic Embeddings.
\newblock In \emph{The Semantic Web. Latest Advances and New Domains: 13th
  International Conference, ESWC 2016, Heraklion, Crete, Greece, May 29 -- June
  2, 2016, Proceedings}, pages 182--198, Cham, 2016. Springer International
  Publishing.
\newblock ISBN 978-3-319-34129-3.

\end{thebibliography}
\end{spacing}
\end{document}